\documentclass[twoside,11pt]{article}

\usepackage{jmlr2e}
\usepackage{amsfonts}
\usepackage{amssymb}
\usepackage{amsmath}
	\usepackage[ruled]{algorithm2e}
	\usepackage{algorithmic}
\usepackage{graphicx,psfrag}
\usepackage{color}

\definecolor{dkgreen}{rgb}{0,0.6,0}
\definecolor{orange}{rgb}{0.8,0.4,0}
\definecolor{blue}{rgb}{0.2,0.2,0.9}
\definecolor{pink}{rgb}{1.0,0.5,0.5}

\newcommand{\bmat}{\begin{pmatrix}}
\newcommand{\emat}{\end{pmatrix}}

\newcommand{\Z}{\mathbb{Z}}
\newcommand{\R}{\mathbb{R}}
\newcommand{\Order}{\mathcal{O}}
\newcommand{\diag}{\operatorname{diag}}
\newcommand{\Symm}{\mathcal{S}}
\newcommand{\PD}{\mathcal{P}}
\newcommand{\Normal}{\mathcal{N}}
\newcommand{\Expectation}{\mathbb{E}}
\newcommand{\trace}{\operatorname{tr}}

\newcommand{\Mean}{\boldsymbol{\mu}}
\newcommand{\Cov}{\mathbf{\Sigma}}
\newcommand{\Tail}{\boldsymbol{\tau}}
\newcommand{\sqCov}{\mathbf{A}}
\newcommand{\sqCovB}{\mathbf{B}}
\newcommand{\nsqCov}{\mathbf{M}}
\newcommand{\nCenter}{\boldsymbol{\delta}}
\newcommand{\covs}{\boldsymbol{\sigma}}
\newcommand{\Sample}{\mathbf{z}}
\newcommand{\SampleTwo}{\mathbf{z'}}
\newcommand{\nSample}{\mathbf{s}}
\newcommand{\transp}{^{\top}}
\newcommand{\popsize}{\lambda}
\newcommand{\prevTheta}{\theta^{\prime}}
\newcommand{\fisher}{\mathbf{F}}
\newcommand{\natG}{\widetilde{\nabla}_{\theta} J}
\newcommand{\idM}{\mathbb{I}}

\ShortHeadings{Natural Evolution Strategies}{Wierstra, Schaul, Glasmachers, Sun and Schmidhuber}
\firstpageno{1}

\begin{document}

\title{Natural Evolution Strategies}

\author{\name Daan Wierstra \email daan.wierstra@epfl.ch \\
       \addr Laboratory of Computational Neuroscience, Brain Mind Institute\\
       \'Ecole Polytechnique F\'ed\'erale de Lausanne (EPFL)\\
       Lausanne, Switzerland
       \AND
       \name Tom Schaul \email tom@idsia.ch \\
       \name Tobias Glasmachers \email tobias@idsia.ch \\
       \name Yi Sun \email yi@idsia.ch \\
       \name J\"{u}rgen Schmidhuber \email juergen@idsia.ch \\
       \addr Istituto Dalle Molle di Studi sull'Intelligenza Artificiale (IDSIA)\\
       University of Lugano (USI)/SUPSI\\
       Galleria 2, 6928, Lugano, Switzerland}

\editor{unknown}

\maketitle

\begin{abstract}%
This paper presents Natural Evolution Strategies (NES), a recent family
	of algorithms that constitute a more principled approach to black-box
	optimization than established evolutionary algorithms.
	NES maintains a parameterized distribution on the set of solution candidates, and
	the natural gradient is used to update the distribution's
	parameters in the direction of higher expected fitness. 
	We introduce a collection of techniques that address issues of
	convergence, robustness, sample complexity, computational complexity and sensitivity to
	hyperparameters.
	This paper explores a number of implementations of the NES family,
	ranging from general-purpose multi-variate normal distributions
	to heavy-tailed and separable distributions tailored towards
	global optimization and search in high dimensional spaces,
	respectively.
	Experimental results show best published performance on various standard benchmarks,
	as well as competitive performance on others.
\end{abstract}


\section{Introduction}

\label{sec:introduction}
Many real world optimization problems are too difficult or complex to model directly.
Therefore, they might best be solved in a `black-box' manner, requiring no
additional information on the objective function (i.e., the `fitness' or `cost')
to be optimized besides
fitness evaluations at certain points in parameter space. 
Problems that fall within this category are numerous, ranging from applications in
health and science~\citep{health,quantum,chromatography} to aeronautic design~\citep{aeronautic,nozzle} and control~\citep{control}. 

Numerous algorithms in this vein have been developed and applied in the past fifty years (see section 2 for an overview),
in many cases providing good and even near-optimal solutions to hard tasks,
which otherwise would have required domain experts to hand-craft solutions at substantial
cost and often with worse results.
The near-infeasibility of finding globally optimal solutions requires
a fair amount of heuristics in black-box optimization algorithms, leading to a proliferation
of sometimes highly performant yet unprincipled methods.

In this paper, we introduce Natural Evolution Strategies (NES), 
a novel black-box optimization framework which
is derived from first principles 
and simultaneously provides state-of-the-art performance.
The core idea is to maintain and iteratively update a search distribution
from which search points are drawn and their fitness evaluated.
The search distribution is then updated in the direction of higher expected fitness,
using ascent along the natural gradient.

\subsection{Continuous Black-Box Optimization}
\label{sec:lit}
The problem of black-box optimization has spawned a wide variety of approaches. 
A first class of methods was inspired by classic optimization methods, including simplex methods such as Nelder-Mead~\citep{neldermead}, as well as members of the quasi-Newton family of algorithms.
Simulated annealing~\citep{simulatedannealing}, a popular method introduced in 1983, was inspired by thermodynamics, and is in fact an adaptation
of the Metropolis-Hastings algorithm. 
More heuristic methods, such as those inspired by evolution, have been developed from the early 1950s on.
These include the broad class of genetic algorithms~\citep{Holland,Goldberg}, differential evolution~\citep{differentialevolution}, 
estimation of distribution algorithms~\citep{EDA}, 
particle swarm optimization~\citep{PSO}, 
and the cross-entropy method~\citep{CEM}.

Evolution strategies (ES), introduced by Ingo Rechenberg and Hans-Paul Schwefel in the 1960s and 1970s~\citep{RechenbergES,schwefelES},
were designed to cope with high-dimensional continuous-valued domains and have remained an active field of research for more than four decades~\citep{beyerESintroduction}. 
ESs involve evaluating the fitness of real-valued genotypes in batch (`generation'),
after which the best genotypes are kept, while the others are discarded. 
Survivors then procreate (by slightly mutating all of their genes) 
in order to produce the next batch of offspring. This process, after several generations, 
was shown to lead to reasonable to excellent results for many difficult 
optimization problems. 
The algorithm framework has been developed extensively over the years to include self-adaptation of the search parameters, and the representation of correlated mutations by the use of a full covariance matrix. This allowed the framework to capture interrelated dependencies by exploiting the covariances while `mutating' individuals for the next generation. The culminating algorithm, covariance matrix adaptation evolution strategy (CMA-ES;~\citealp{hansen:2001}), has proven successful in numerous studies (e.g., \citealp{svmIgel,chemotaxis,crystal}).

While evolution strategies prove to be an effective framework for black-box optimization, 
their ad hoc procedures remain heuristic in nature.
Thoroughly analyzing the actual dynamics of the procedure turns out to be difficult, the considerable efforts of various researchers notwithstanding~\citep{EStheory,jah:2010a,Auger2005proof}.
In other words, ESs (including CMA-ES), while powerful, still lack a clear derivation from first principles.

\subsection{The NES Family}
Natural Evolution Strategies (NES) are well-grounded, 
evolution-strategy inspired black-box optimization algorithms, 
which instead of maintaining a population of search points,
iteratively update a search \emph{distribution}.
Like CMA-ES, they can be cast into the framework of evolution strategies.

The general procedure is as follows:
the parameterized search distribution 
is used to produce a batch of search points,
and the fitness function is evaluated at each such point.
The distribution's parameters (which include strategy parameters)
allow the algorithm to adaptively
capture the (local) structure of the fitness function.
For example, in the case of a Gaussian distribution, this comprises the mean and the covariance matrix.
From the samples, NES estimates a search gradient on the parameters
towards higher expected fitness.
NES then performs a gradient ascent step along the \emph{natural gradient},
a second-order method which, unlike the plain gradient,
renormalizes the update w.r.t.~uncertainty. This step is crucial, since
it prevents oscillations, premature convergence, and undesired effects
stemming from a given parameterization.
The entire process reiterates until a stopping criterion is met.

All members of the `NES family' operate based on the same principles.
They differ in the type of distribution and the gradient approximation method used.
Different search spaces require different search distributions;
for example, in low dimensionality it can be highly beneficial to model the full
covariance matrix.
In high dimensions, on the other hand, a more scalable alternative is
to limit the covariance to the diagonal only.
In addition, highly multi-modal search spaces may benefit
from more heavy-tailed distributions (such as Cauchy, as opposed to the Gaussian).
A last distinction arises between 
distributions where we can analytically compute the natural
gradient, and more general distributions where we need to estimate it from samples.

\subsection{Paper Outline}
This paper builds upon and extends our previous
work on Natural Evolution Strategies~\citep{wierstra:2008,sun:2009a,sun:2009b,Glasmachers2010,Glasmachers2010a,Schaul2011snes},
and introduces novel performance- and robustness-enhancing techniques (in sections~\ref{sec:as} and \ref{sec:restarts}), as well as an extension to rotationally symmetric distributions (section~\ref{sec:rotsym})
and a plethora of new experimental results.

The paper is structured as follows. 
Section~\ref{sec:sg} presents the general idea of search gradients as
introduced by \cite{wierstra:2008}, outlining how to perform stochastic
search using parameterized distributions while doing gradient ascent
towards higher expected fitness.
The limitations of the plain gradient are exposed in section~\ref{sec:vanillalimitations},
and subsequently addressed by the introduction of the natural gradient (section~\ref{sec:ng}),
resulting in the canonical NES algorithm.

Section~\ref{sec:robust-techs} then regroups a collection of techniques
that enhance NES's performance and robustness. This
includes fitness shaping (designed to render the algorithm invariant
w.r.t.\ order-preserving fitness transformations (\citealp{wierstra:2008}), section~\ref{sec:fs}),
importance mixing (designed to recycle samples so as to
reduce the number of required fitness evaluations (\citealp{sun:2009b}), section~\ref{sec:im}),
adaptation sampling which is a novel technique for adjusting learning rates online (section~\ref{sec:as}),
and finally restart strategies, designed to improve success rates
on multi-modal problems (section~\ref{sec:restarts}).

In section~\ref{sec:multinormal}, we look in more depth at multivariate
Gaussian search distributions, constituting the most common case.
We show how to constrain the covariances to positive-definite matrices 
using the exponential map (section~\ref{sec:expmap}), and how to
use a change of coordinates to reduce the computational complexity
from $\Order(d^6)$ to $\Order(d^3)$, with $d$ being the search space
dimension, resulting in the xNES algorithm (\citealp{Glasmachers2010a}, section~\ref{sec:xnes}).

Next, in section~\ref{sec:other-dists}, we develop the breadth
of the framework, motivating its usefulness and deriving 
a number of NES variants with different search distributions.
First, we show how a restriction to diagonal parameterization
permits the approach to scale to very high dimensions due to its 
linear complexity (\citealp{Schaul2011snes}; section~\ref{sec:snes}).
Second, we provide a novel formulation of NES for the whole class of multi-variate
versions of distributions with rotation symmetries (section~\ref{sec:rotsym}),
including heavy-tailed distributions with infinite variance, such as the
Cauchy distribution (\citealp{Schaul2011snes}, section~\ref{sec:cauchy-nes}).

The ensuing experimental investigations show the competitiveness of the approach
on a broad range of benchmarks (section~\ref{sec:experiments}). 
The paper ends with a discussion on 
the effectiveness of the different techniques and types of distributions
and an outlook towards future developments (section~\ref{sec:discussion}).

\section{Search Gradients}
\label{sec:sg}

The core idea of Natural Evolution Strategies is to use \emph{search gradients}
to update the parameters of the search distribution.
We define the search gradient as the
sampled  gradient of expected fitness. 
The search distribution
can be taken to be a multinormal distribution, but could in principle be
any distribution for which we can find derivatives of its log-density
w.r.t.\ its parameters. For example, useful distributions include
Gaussian mixture models and the Cauchy distribution with its heavy tail.

If we use $\theta$ to denote the parameters of distribution $\pi(\Sample \,|\, \theta)$ and $f(x)$ to denote the fitness function for samples $\Sample$, we can write
the expected fitness under the search distribution as
\begin{align}
	J(\theta) = \Expectation_\theta[f(\Sample)] = \int f(\Sample) \; \pi(\Sample \,|\, \theta) \; d\Sample
	.
	\label{eq:expected-fitness}	
\end{align}
The so-called `log-likelihood trick' enables us to write
\begin{align*}
 \nabla_{\theta} J(\theta) = & \nabla_{\theta} \int f(\Sample) \; \pi(\Sample \,|\, \theta) \; d\Sample \\
   = & \int f(\Sample) \; \nabla_{\theta} \pi(\Sample \,|\, \theta) \; d\Sample \\
   = & \int f(\Sample) \; \nabla_{\theta} \pi(\Sample \,|\, \theta) \; \frac{\pi(\Sample \,|\, \theta)}{\pi(\Sample \,|\, \theta)} \; d\Sample \\
   = & \int \Big[ f(\Sample) \; \nabla_{\theta} \log\pi(\Sample \,|\, \theta) \Big] \; \pi(\Sample \,|\, \theta) \; d\Sample\\ 
   = & \Expectation_\theta \left[ f(\Sample) \; \nabla_{\theta} \log\pi(\Sample \,|\, \theta)\right]
.
\end{align*}
From this form we obtain the Monte Carlo estimate of the search gradient from samples $\Sample_1\ldots \Sample_\popsize$ as
\begin{align}
\label{eq:mcgradient}
 \nabla_{\theta} J(\theta) \approx 
   \frac{1}{\popsize} 
\sum_{k=1}^{\popsize} f(\Sample_k) \; \nabla_{\theta} 
\log\pi(\Sample_k \,|\, \theta)
	,
\end{align}
where $\popsize$ is the population size.
This gradient on expected fitness provides a search direction in the space of search
distributions. A straightforward gradient ascent scheme can thus
iteratively update the search distribution 
\begin{align*}
 \theta \leftarrow \theta + \eta \nabla_{\theta} J(\theta) 
 ,
\end{align*}
where $\eta$ is a learning rate parameter.
Algorithm~\ref{alg:basicNES} provides the pseudocode for this very general approach to black-box optimization by using a search gradient on search distributions.

\begin{algorithm}
\caption{Canonical Search Gradient algorithm}\label{alg:basicNES}
\DontPrintSemicolon
\SetKwInOut{Input}{input}
 \Input{$f$, $\theta_{init}$}
 \Repeat{stopping criterion is met}{
 \For{$k=1\ldots\popsize$}{
  draw sample $\Sample_k \sim \pi(\cdot | \theta)$\\
  evaluate the fitness $f(\Sample_k)$\\
  calculate log-derivatives $\nabla_\theta \log\pi(\Sample_k | \theta)$\\
 }
   $\displaystyle\nabla_\theta J \leftarrow \frac{1}{\popsize}
\sum_{k=1}^{\popsize} \nabla_\theta\log\pi(\Sample_k | \theta)\cdot f(\Sample_k)$\\
  \vspace{.2cm}
  $\displaystyle\theta \leftarrow \theta + \eta \cdot \nabla_\theta J$ \\
 }
\end{algorithm}

Utilizing the search gradient in this framework is similar to evolution strategies 
 in that it iteratively generates the fitnesses of batches of vector-valued samples --
 the ES's so-called candidate solutions.  It is different however, in that it represents this
 `population' as a parameterized distribution,
 and in the fact that it uses a search gradient to update the parameters of this distribution, 
 which is computed using the fitnesses.

\subsection{Search Gradient for Gaussian Distributions}
In the case of the `default' $d$-dimensional multi-variate normal distribution,
we collect the parameters of the Gaussian, the mean~$\Mean \in \R^{d}$ (candidate solution center) 
and the covariance matrix~$\Cov\in \R^{d \times d}$ (mutation matrix), 
in a single concatenated vector $\theta = \langle\Mean, \Cov\rangle$. However, to
sample efficiently from this distribution we need a square root of the
covariance matrix (a matrix $\sqCov \in \R^{d \times d}$ fulfilling
$\sqCov\transp \sqCov = \Cov$).%
\footnote{For any matrix $\mathbf{Q}$,
  $\mathbf{Q}\transp$ denotes its transpose.}
Then $\Sample = \Mean + \sqCov\transp \nSample$ transforms a standard normal vector
$\nSample \sim \Normal(0, \idM)$ into a sample $\Sample \sim \Normal(\Mean, \Cov)$.
Here, $\idM = \diag(1, \dots, 1) \in \R^{d \times d}$ denotes the identity matrix.
Let
\begin{align*}
	\pi(\Sample \,|\, \theta)
	\,=\, & \frac{1}{(\sqrt{2\pi})^{d} \det(\sqCov)} \cdot \exp \left( -\frac{1}{2} 
		\Big\| \sqCov^{-1} \cdot (\Sample - \Mean) \Big\|^2 \right) \\
	\,=\, & \frac{1}{\sqrt{(2\pi)^d \det(\Cov)}} \cdot \exp \left( -\frac{1}{2}
		(\Sample-\Mean)\transp \Cov^{-1} (\Sample-\Mean) \right)
\end{align*}
denote the density of the multinormal search distribution
$\Normal(\Mean, \Cov)$.

In order to calculate the derivatives of the log-likelihood with respect
to individual elements of $\theta$ for this 
multinormal distribution, first note that
\begin{eqnarray*}
  \log\pi\left(\Sample|\theta\right) = -\frac{d}{2}\log(2\pi) 
  - \frac{1}{2}\log\det\Cov
  - \frac{1}{2}\left(  \Sample-\Mean\right)\transp\Cov
  ^{-1}\left(\Sample-\Mean\right)  
  .
\end{eqnarray*}
We will need its derivatives, that is, $\nabla_{\Mean}\log\pi\left(  \Sample|\theta\right)  $ and $\nabla_{\Cov}\log\pi\left(  \Sample|\theta\right)  $.\ The first is trivially
\begin{equation}
  \nabla_{\Mean}\log\pi\left(  \Sample|\theta\right)  
  =\Cov^{-1}\left(  \Sample-\Mean\right)  
  , \label{eq:vanilla-mean}
\end{equation}
while the latter is 
\begin{equation}
  \nabla_{\Cov}\log \pi\left(  \Sample|\theta\right)
  =  \frac{1}{2}\Cov^{-1}\left(  \Sample-\Mean\right)  \left(  \Sample-\Mean
  \right)  \transp\Cov^{-1}-\frac{1}{2}\Cov^{-1}
  . \label{eq:vanilla-cov}
\end{equation}
In order to preserve symmetry, to ensure
non-negative 
variances and to keep the mutation matrix $\Cov$ positive
definite, $\Cov$ needs to be constrained.
One way to accomplish that is by representing $\Cov$ as a product 
$\Cov = \sqCov\transp\sqCov$ (for a more sophisticated solution to this issue, see section~\ref{sec:expmap}). Instead of using the
log-derivatives on $\nabla_{\Cov}\log\pi\left(  \Sample\right)$ directly, we then compute 
the derivatives with respect to $\sqCov$ as
\begin{equation*}
	\nabla_{\sqCov} \log\pi\left(\Sample\right) = \sqCov\left[
	\nabla_{\Cov}\log\pi\left(\Sample\right) + 
	\nabla_{\Cov}\log\pi\left(\Sample\right)\transp
	\right]
	.
\end{equation*}

\noindent Using these derivatives to calculate  $\nabla_\theta J$, we can then
update parameters $\theta = \langle\Mean, \Cov = \sqCov\transp\sqCov\rangle$ 
as $\theta \leftarrow \theta + \eta \nabla_\theta J$ using learning rate $\eta$. This
produces a new center $\Mean$ for the search distribution, and simultaneously self-adapts its 
associated covariance matrix $\Cov
$. 
To summarize, we provide the pseudocode for following the search gradient in the case of a multinormal search distribution in Algorithm~\ref{alg:multinormalNES}.

\begin{algorithm}
\caption{Search Gradient algorithm: Multinormal distribution}
\label{alg:multinormalNES}
\DontPrintSemicolon
\SetKwInOut{Input}{input}
 \Input{$f$, $\Mean_{init}$,$\Cov_{init}$}
 \Repeat{stopping criterion is met}{
 \For{$k=1\ldots\popsize$}{
  draw sample $\Sample_k \sim \Normal(\Mean, \Cov)$\\
  evaluate the fitness $f(\Sample_k)$\\
$\begin{array}{l}
 \hspace{-0.2cm}\mbox{calculate log-derivatives:} \\ 
 \nabla_{\Mean} \log\pi\left(\Sample_k |\theta\right)
=\Cov^{-1}\left(\Sample_k-\Mean\right)\\
 \nabla_{\Cov} \log \pi\left(  \Sample_k |\theta\right)
  =  -\frac{1}{2}\Cov^{-1}
+\frac{1}{2}\Cov^{^{-1}}\left(  \Sample_k -\Mean\right) 
 \left(  \Sample_k -\Mean\right)  \transp\Cov^{-1}
\end{array}$
}
$\nabla_{\Mean} J \leftarrow \frac{1}{\popsize}
\sum_{k=1}^{\popsize} 
\nabla_{\Mean}\log\pi(\Sample_k | \theta)\cdot f(\Sample_k)$\\
  \vspace{.1cm}
$\nabla_{\Cov} J \leftarrow \frac{1}{\popsize}
\sum_{k=1}^{\popsize} 
\nabla_{\Cov}\log\pi(\Sample_k | \theta)\cdot f(\Sample_k)$\\
  \vspace{.1cm}
$\Mean \leftarrow \Mean + \eta \cdot \nabla_{\Mean} J$\\
$\Cov  \leftarrow \Cov  + \eta \cdot \nabla_{\Cov } J$
 }
\end{algorithm}

\subsection{Limitations of Plain Search Gradients}
\label{sec:vanillalimitations}

\begin{figure}[ht]
	\centerline{	
		\includegraphics[height=0.39\columnwidth, clip=true, trim=0cm 0cm 0cm 0.2cm]{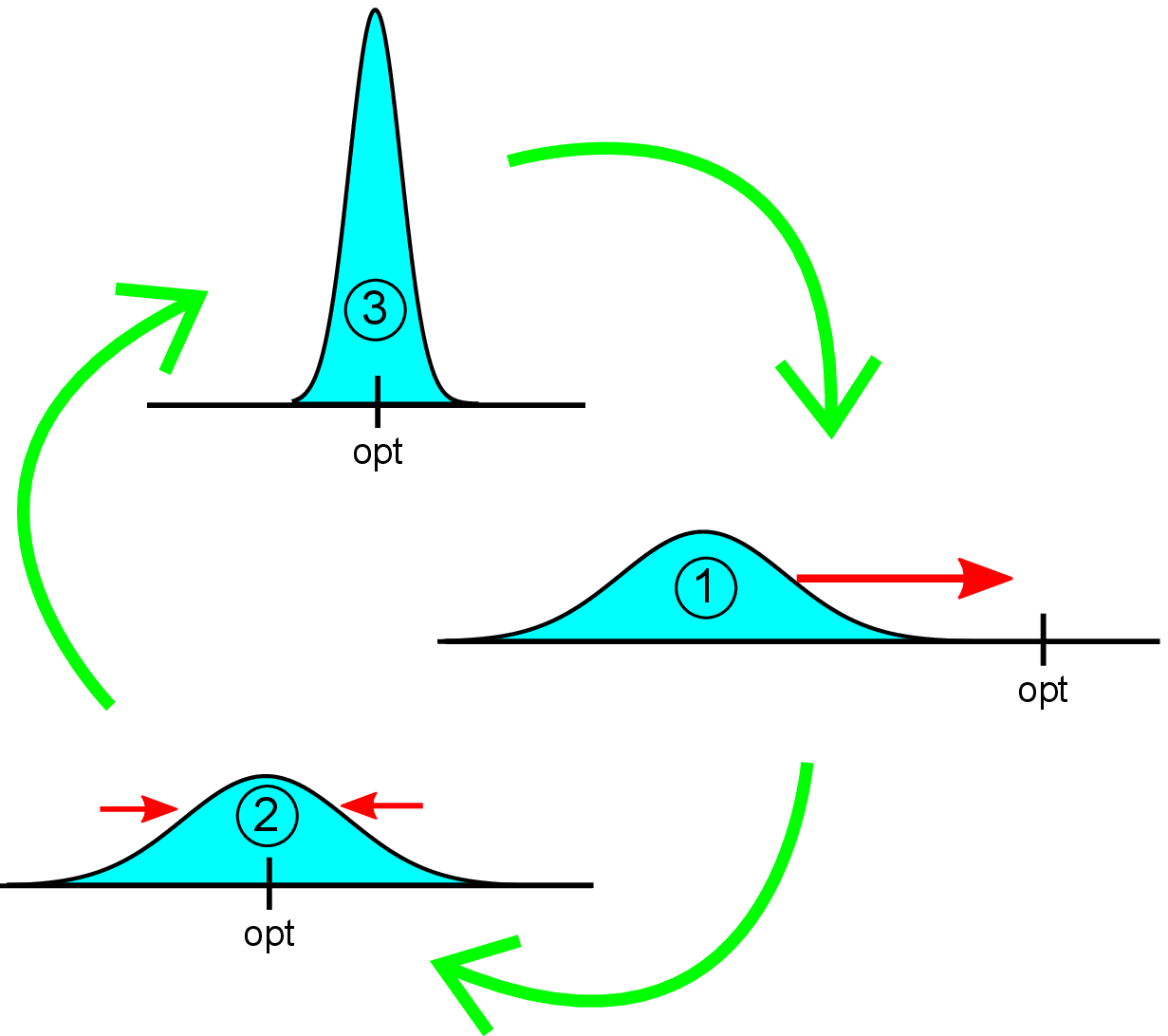}
		\hspace{0.05\columnwidth}
		\includegraphics[height=0.41\columnwidth, clip=true, trim=0.3cm 0.6cm 0.7cm 0cm]{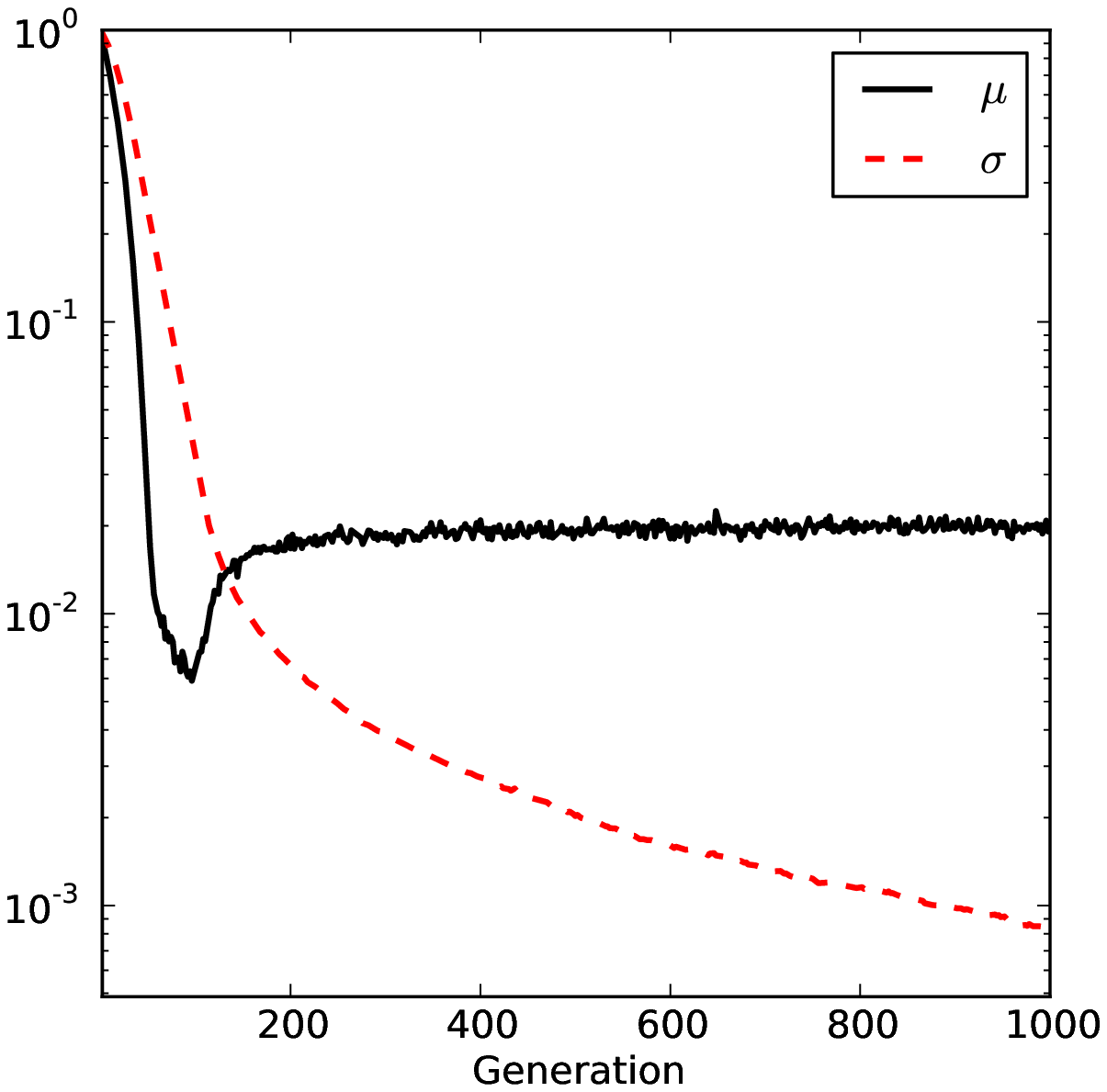}
	}
	\caption[Lack of scale invariance]{\textbf{Left:} 
	Schematic illustration of how the search distribution adapts in the one-dimensional case: 
	from (1) to (2), $\mu$ is adjusted to make the distribution cover the optimum.
	From (2) to (3), $\sigma$ is reduced to allow for a precise localization of the optimum. 
	The step from (3) to (1) then is the problematic case, 
	where a small $\sigma$ induces a largely overshooting update, making the search start over again. 
	\textbf{Right:} Progression of $\mu$ (black) and $\sigma$ (red, dashed) 
	when following the search gradient towards minimizing $f(\Sample) = \Sample^2$, 
	executing Algorithm~\ref{alg:multinormalNES}. 
  Plotted are median values over 1000 runs, with a small learning rate $\eta = 0.01$ and 
  $\popsize= 10$, both of which mitigate the instability somewhat, but still show the failure to 
  precisely locate the optimum (for which both $\mu$ and $\sigma$ need to approach 0).
  }
	\label{fig:mu-update-illust}
\end{figure}

As the attentive reader will have realized, there exists at least one major issue with 
applying the search gradient as-is in practice: 
It is impossible to \emph{precisely locate} a (quadratic) optimum, even in the
one-dimensional case. 
Let $d=1$, $\theta = \langle\mu, \sigma\rangle$, and samples $z \sim \Normal(\mu, \sigma)$. 
Equations~\eqref{eq:vanilla-mean} and~\eqref{eq:vanilla-cov}, the gradients on $\mu$ and $\sigma$, become
\begin{eqnarray*}
\nabla_{\mu} J &=& \frac{z-\mu}{\sigma^2}
\\
\nabla_{\sigma} J &=& \frac{(z-\mu)^2 - \sigma^2}{\sigma^3}
\end{eqnarray*}
%
%
and the updates, assuming simple hill-climbing (i.e. a population size $\popsize=1$) read:
\begin{eqnarray*}
\mu &\leftarrow& \mu + \eta  \frac{z-\mu}{\sigma^2} \\
  \sigma &\leftarrow& \sigma + \eta \frac{(z-\mu)^2 - \sigma^2}{\sigma^3} 
  .
\end{eqnarray*}
For any objective function $f$ that requires locating an (approximately) quadratic optimum 
with some degree of precision (e.g.~$f(\Sample) = \Sample^2$), 
$\sigma$ must decrease, which in turn increases the variance of
the updates, as $\Delta\mu \propto \frac{1}{\sigma}$ and
$\Delta\sigma \propto \frac{1}{\sigma}$ for a typical sample $z$. 
In fact, the updates become increasingly unstable, the smaller $\sigma$ becomes,
an effect which a reduced learning rate or an
increased population size can only delay but not avoid.
Figure~\ref{fig:mu-update-illust} illustrates this effect.
Conversely, whenever $\sigma \gg 1$ is large, the magnitude of a typical update is 
severely reduced. 

Clearly, this update is not at all \emph{scale-invariant}:
Starting with $\sigma \gg 1$ makes all updates minuscule, 
whereas starting with $\sigma \ll 1$ makes the first update huge 
and therefore unstable.

We conjecture that this limitation constitutes one of the main reasons why 
search gradients have not been developed before: in isolation, the plain search gradient's performance
is both unstable and unsatisfying, and it is only the application of natural gradients 
(introduced in section~\ref{sec:ng}) which
tackles these issues and renders search gradients into a viable optimization method.

\subsection{Using the Natural Gradient}
\label{sec:ng}

Instead of using the plain stochastic gradient for updates, NES
follows the \emph{natural gradient}.
The natural gradient was first introduced by Amari in 1998, and has 
been shown to 
possess numerous advantages over the plain gradient~\citep{amari98natural,whynaturalamari}. 
In terms of mitigating the slow convergence of 
plain gradient ascent in optimization landscapes
with ridges and plateaus, natural gradients are a more principled 
(and hyper-parameter-free) approach
than, for example, the commonly used momentum heuristic.

\begin{figure}
\centerline{
\psfrag{mu}[l][l]{$\Mean$}
\psfrag{sigma}[c][c]{$\sigma$}
\includegraphics[width=\columnwidth]{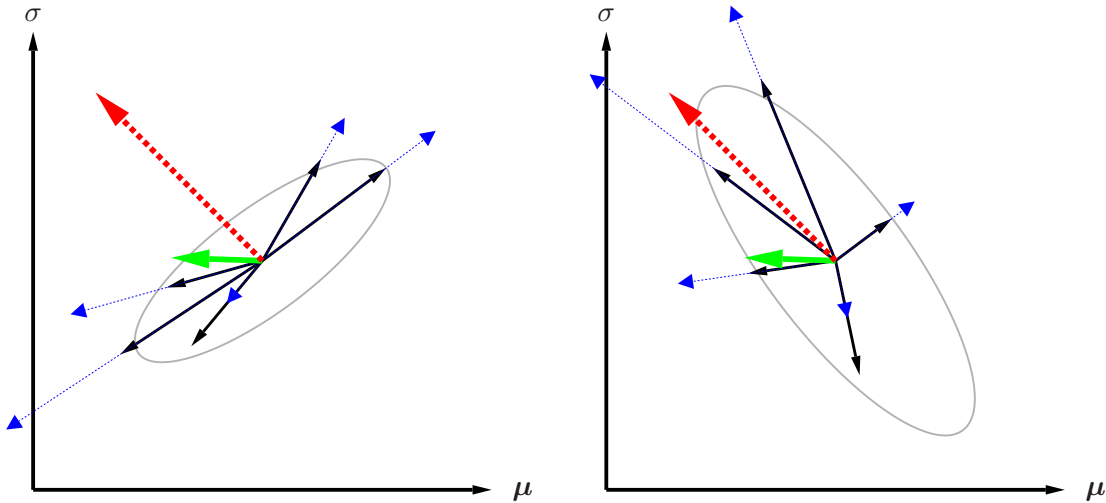}
}
\caption[Illustration of plain vs. natural gradient]{
Illustration of plain versus natural gradient in parameter space.
Consider two parameters, e.g. $\theta = (\mu, \sigma)$, of the search
distribution. In the plot on the left, the solid (black) arrows indicate
the gradient samples $\nabla_{\theta} \log \pi (\Sample\,|\,\theta) $,
while the dotted (blue) arrows correspond to
$f(\Sample) \cdot \nabla_{\theta} \log \pi(\Sample\,|\,\theta) $, that
is, the same gradient estimates, but scaled with fitness. Combining
these, the bold (green) arrow indicates the (sampled) fitness gradient
$\nabla_\theta J$, while the bold dashed (red) arrow indicates the
corresponding natural gradient $\tilde \nabla_\theta J$.

Being random variables with expectation zero, the distribution of the
black arrows is governed by their covariance, indicated by the gray
ellipse. Notice that this covariance is a quantity in \emph{parameter
space} (where the $\theta$ reside), which is not to be confused with the covariance of the 
distribution in the \emph{search space} (where the samples $\Sample$ reside).

In contrast, solid (black) arrows on the right represent
$\tilde \nabla_{\theta} \log \pi(\Sample\,|\,\theta) $, and
dotted (blue) arrows indicate the \emph{natural} gradient samples
$f(\Sample) \cdot \tilde \nabla_{\theta} \log \pi(\Sample\,|\,\theta)$,
resulting in the natural gradient (dashed red).

The covariance of the solid arrows on the right hand side turns out to
be the inverse of the covariance of the solid arrows on the left. This
has the effect that when computing the natural gradient, directions with
high variance (uncertainty) are penalized and thus shrunken, while
components with low variance (high certainty) are boosted, since these
components of the gradient samples deserve more trust. This makes the
(dashed red) natural gradient a much more trustworthy update direction
than the (green) plain gradient.

%
}
\label{fig:natgrad-illust}
\end{figure}

The plain gradient $\nabla J$ simply follows the steepest ascent in the
space of the actual parameters $\theta$ of the distribution. This means that for a
given small step-size $\varepsilon$, 
following it will yield a new distribution with parameters 
chosen from the hypersphere of radius $\epsilon$ and center $\theta$ that maximizes $J$.
In other words, the Euclidean distance in parameter space between subsequent distributions is fixed.
Clearly, this makes the update dependent on the particular parameterization
of the distribution, therefore a change of parameterization leads to different 
gradients and different updates.
See also Figure~\ref{fig:natgrad-illust} for an illustration of how this effectively
renormalizes updates w.r.t.~uncertainty.

The key idea of the natural gradient is to remove this dependence on the 
parameterization by relying on a more `natural' measure of 
distance $D(\prevTheta||\theta)$ between probability distributions
$\pi\left(  \Sample | \theta \right)  $ and $\pi\left(\Sample | \prevTheta\right)$.
One such natural distance measure between two probability distributions is the Kullback-Leibler
divergence~\citep{Kullback1951}. 
The natural gradient 
can then be formalized as the solution to the constrained optimization problem
\begin{align*}
  \max_{\delta\theta} J\left(\theta+\delta\theta\right)  
  & \approx J\left(\theta\right) + \delta\theta\transp\nabla_\theta J,\\
  s.t. ~D\left(  \theta+\delta\theta||\theta\right) & =\varepsilon
  ,
\end{align*}
where $J\left(\theta\right)$ is the expected fitness of equation~\eqref{eq:expected-fitness},
and $\varepsilon$ is a small increment size.
Now, we have for $\delta\theta\rightarrow0$,
\begin{equation*}
  D\left(  \theta+\delta\theta||\theta\right)  
  = \frac{1}{2}\delta\theta\transp\fisher\left(\theta\right)  \delta\theta
  ,
\end{equation*}
where
\begin{align*}  
\fisher& =\int \pi\left(  \Sample|\theta\right)  \nabla_\theta\log \pi\left(
  \Sample|\theta\right)  \nabla_\theta\log \pi\left(  \Sample|\theta\right)  
\transp d\Sample,\\  
  & =\Expectation\left[  \nabla_\theta\log \pi\left(  \Sample|\theta\right)  
\nabla_\theta\log \pi\left(  \Sample|\theta\right)
  \transp\right]
\end{align*}
is the \emph{Fisher information matrix} of the given parametric family of search
distributions. The solution to this can be found 
using a Lagrangian multiplier~\citep{petersthesis}, yielding the necessary condition
\begin{equation}
	\fisher\delta\theta=\beta\nabla_\theta J
  ,
\label{eq:natgradient} 
\end{equation}
for some constant $\beta > 0$. The direction of the natural gradient $\natG$ is given by $\delta\theta$ thus defined.
If $\fisher$ is invertible\footnote{Care has to be taken because the Fisher matrix estimate may
not be (numerically) invertible even if the exact Fisher matrix is.}, the natural gradient amounts to 
\begin{equation*}
	\natG = \fisher^{-1} \nabla_{\theta} J(\theta)
	.
\end{equation*}
The Fisher matrix can be estimated from samples, reusing the log-derivatives $\nabla_\theta \log\pi(\Sample|\theta)$ that we already computed for the gradient $\nabla_\theta J$. 
Then, updating the parameters following the natural gradient instead of the steepest gradient leads
us to the general formulation of NES, as shown in Algorithm~\ref{alg:general-nes}.

\begin{algorithm}
\caption{Canonical Natural Evolution Strategies}
\label{alg:general-nes}
\DontPrintSemicolon
\SetKwInOut{Input}{input}
 \Input{$f$, $\theta_{init}$}
 \Repeat{stopping criterion is met}{
 \For{$k=1\ldots\popsize$}{
  draw sample $\Sample_k \sim \pi(\cdot | \theta)$\\
  evaluate the fitness $f(\Sample_k)$\\
  calculate log-derivatives $\nabla_\theta \log\pi(\Sample_k | \theta)$\\
 }
   $
\nabla_\theta J \leftarrow \frac{1}{\popsize}
\sum_{k=1}^{\popsize} \nabla_\theta\log\pi(\Sample_k | \theta)\cdot f(\Sample_k)$\\
  \vspace{.1cm}
$\displaystyle\fisher \leftarrow \frac{1}{\popsize}\sum_{k=1}^{\popsize} 
\nabla_\theta \log \pi\left(\Sample_k | \theta\right)  
\nabla_\theta \log \pi\left(\Sample_k | \theta\right)\transp$\\
  \vspace{.1cm}
  $\displaystyle\theta \leftarrow \theta + \eta \cdot \fisher^{-1} \nabla_\theta J$ \\
 }
\end{algorithm}

\section{Performance and Robustness Techniques}
\label{sec:robust-techs}
In the following we will present and introduce crucial techniques to improves NES's performance and robustness.
Fitness shaping \citep{wierstra:2008} is designed to make the algorithm invariant
w.r.t.\ arbitrary yet order-preserving fitness transformations (section~\ref{sec:fs}).
Importance mixing \citep{sun:2009b} is designed to recycle samples so as to
reduce the number of required fitness evaluations, and is subsequently presented in section~\ref{sec:im}.
Adaptation sampling, a novel technique for adjusting learning rates online, is introduced in section~\ref{sec:as},
and finally restart strategies, designed to improve success rates
on multimodal problems, is presented in section~\ref{sec:restarts}.

\subsection{Fitness Shaping}
\label{sec:fs}
NES utilizes rank-based fitness shaping in order to render the
algorithm \emph{invariant} under monotonically
increasing
(i.e., rank preserving)
transformations of the fitness function. 
For this purpose, the fitness
of the population is transformed into a set of utility values
$u_1 \geq \dots \geq u_\popsize$. Let $\Sample_i$ denote the $i^{th}$ best individual
(the $i^{th}$ individual in the population, sorted by fitness, such that
$\Sample_1$ is the best and $\Sample_\popsize$ the worst individual). Replacing fitness
with utility, the gradient estimate of equation~\eqref{eq:mcgradient} becomes,
with slight abuse of notation, 
\begin{align}
	\nabla_{\theta} J (\theta) = \sum_{k=1}^\popsize u_k \; \nabla_{\theta} \log\pi(\Sample_k \,|\, \theta)
	.
	\label{eq:gradient}
\end{align}
To avoid entangling the utility-weightings with the learning rate, 
we require that $\sum_{k=1}^{\popsize} |u_k|  = 1$.

The choice of utility function can in fact be seen as a free parameter
of the algorithm. Throughout this paper we will use the following
\[
u_k=\displaystyle \frac{\max\left(0, \log(\frac{\popsize}{2}+1) 
- \log(i)\right)}{\sum_{j=1}^{\popsize} 
\max\left(0, \log(\frac{\popsize}{2}+1) - \log(j)\right)} - \frac{1}{\popsize}
,
\]
which is directly related to the one employed by CMA-ES~\citep{hansen:2001}, for ease of comparison. 
In our experience, however, this choice has not been crucial to performance,
as long as it is monotonous and based on ranks instead of raw fitness (e.g., a function which simply increases linearly with rank).

In addition to robustness, these utility values provide us with an elegant
formalism to describe the (1+1) hill-climber version of the algorithm
within the same framework, by using different utility values, depending on 
success (see section~\ref{sec:elitism} later in this paper).

\subsection{Importance Mixing}
\label{sec:im}
In each batch, we evaluate $\popsize$ new samples generated from search
distribution $\pi\left( \Sample|\theta \right) $. However, since small
updates ensure that the KL divergence between consecutive search
distributions is generally small, most new samples will fall in the high
density area of the previous search distribution $\pi\left( \Sample
|\prevTheta\right) $. This leads to redundant fitness evaluations in
that same area.
We improve the efficiency with 
a new procedure called \emph{importance mixing},
which aims at \emph{reusing} fitness evaluations from the previous batch,
while ensuring the updated batch conforms to the new search
distribution.

\begin{figure}
	\centerline{
		\includegraphics[height=0.4\columnwidth, clip=true, trim=0cm 0cm 0cm 0cm]{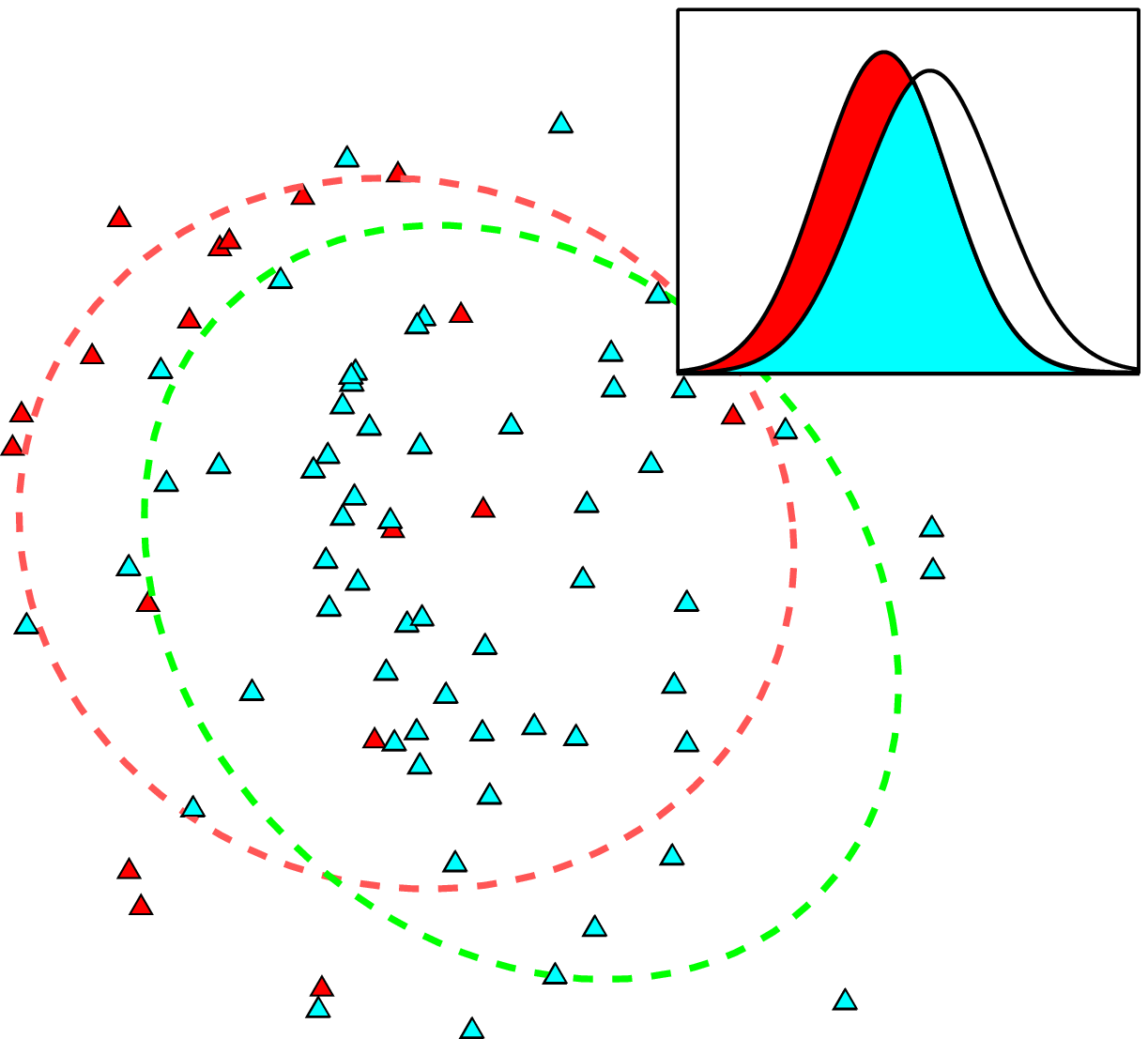}
		\hspace{0.1\columnwidth}
		\includegraphics[height=0.4\columnwidth, clip=true, trim=0cm 0cm 0cm 0cm]{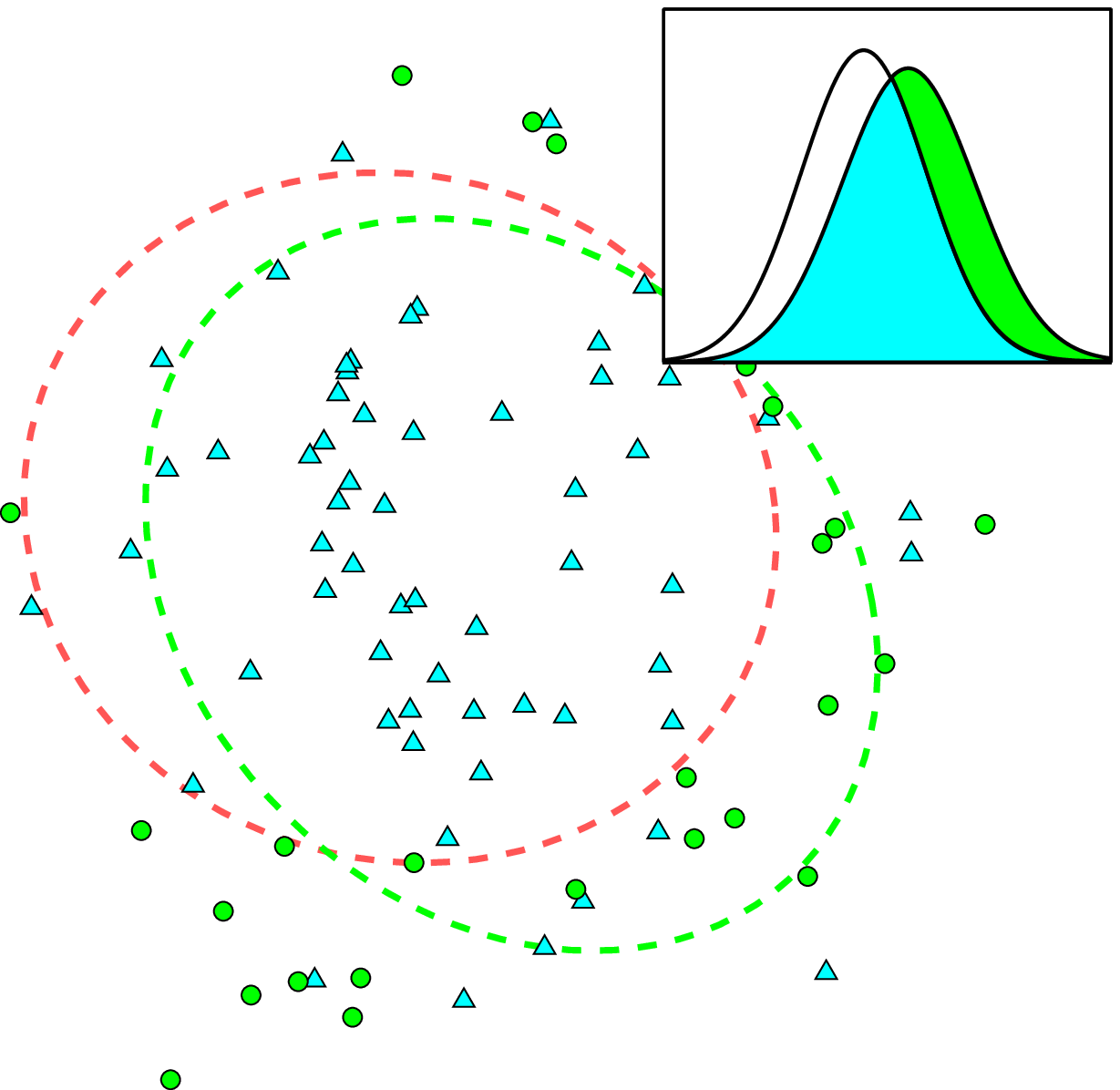}
	}
	\caption[Illustration of importance mixing]{\textbf{}Illustration of importance mixing. 
	\textbf{Left:} In the first step, old samples are eliminated (red triangles) 
	according to~\eqref{eq:im-accept}, and the remaining samples (blue triangles) are reused. 
	\textbf{Right:} In the second step, new samples (green circles) are generated 
	according to~\eqref{eq:im-generate}. 
	}
	\label{fig:im-illust}
\end{figure}

Importance mixing works in two steps: In the first step, rejection sampling
is performed on the previous batch, such that sample $\Sample$
is accepted with probability
\begin{equation}
  \min \left\{ 1,\left( 1-\alpha \right) \frac{\pi\left( \Sample|\theta
  \right) }{\pi\left( \Sample|\prevTheta\right) }\right\} 
  .
  \label{eq:im-accept}
\end{equation}
Here $\alpha \in \left[ 0,1\right] $ is an algorithm hyperparameter called the \emph{minimal refresh rate}. Let 
$\popsize_{a}$ be the number of samples accepted in the first step. In the
second step, reverse rejection sampling is performed as follows: Generate
samples from $\pi\left( \Sample|\theta \right) $ and accept $\Sample$
with probability
\begin{equation}
  \max \left\{ \alpha ,1-\frac{\pi\left( \Sample|\prevTheta\right) }{%
  \pi\left( \Sample|\theta \right) }\right\}   
  \label{eq:im-generate}
\end{equation}
until $\popsize-\popsize_{a}$ new samples are accepted. The $\popsize_{a}$ samples from
the old batch and $\popsize-\popsize_{a}$ newly accepted samples together
constitute the new batch. Figure~\ref{fig:im-illust} illustrates the procedure. 
Note that only the fitnesses of the newly
accepted samples need to be evaluated. The advantage of using importance
mixing is twofold: On the one hand, we reduce the number of fitness
evaluations required in each batch, on the other hand, if we fix the
number of newly evaluated fitnesses, then many more fitness
evaluations can potentially be used to yield more reliable and accurate gradients.

The minimal refresh rate parameter $\alpha$ has two uses.
First, it avoids too low an acceptance probability at the second step 
when $\pi\left( \Sample|\prevTheta\right) 
/\pi\left( \Sample|\theta \right) \simeq 1$.
And second, it permits specifying a lower bound on 
the expected proportion of newly evaluated samples 
$\rho =\mathbb{E}\left[ \frac{\popsize-\popsize_{a}}{\popsize}\right]$, 
namely $\rho \geq \alpha $, with the equality holding if and only if 
$\theta = \prevTheta$. In particular, if $\alpha =1$, all samples from the
previous batch will be discarded, and if $\alpha =0$, $\rho $ depends
only on the distance between $\pi\left( \Sample|\theta \right) $ and 
$\pi\left( \Sample|\prevTheta\right) $. Normally we set $\alpha $ to
be a small positive number, e.g., in this paper we use $\alpha = 0.1$ throughout.

It can be proven that the updated batch conforms to the search
distribution $\pi\left( \Sample|\theta \right) $. In the region where 
\begin{equation*}
  \left( 1-\alpha \right) \frac{\pi\left( \Sample|\theta \right)}
  {\pi\left( \Sample |\prevTheta\right)} \leq 1
  ,
\end{equation*} 
the probability that a sample from
previous batches appears in the new batch is
\begin{equation*}
  \pi\left( \Sample|\prevTheta\right) \cdot \left( 1-\alpha \right)
  \frac{\pi\left( \Sample|\theta \right)}{\pi\left( \Sample|\prevTheta\right)}
  =\left( 1-\alpha \right) \pi\left( \Sample|\theta \right)
  .
\end{equation*}
The probability that a sample generated from the second step entering
the batch is $\alpha \pi\left( \Sample|\theta \right) $, since 
\begin{equation*}
  \max \left\{ \alpha ,1-\frac{\pi\left( \Sample|\prevTheta\right)}
  {\pi\left( \Sample|\theta \right) } \right\}   =\alpha
  .
\end{equation*}
So the probability of a sample entering the batch is just 
$p\left( \Sample|\theta\right) $ in that region. 
The same result holds also for the region where
\begin{equation*}
  \left( 1-\alpha \right) \frac{\pi\left( \Sample|\theta \right)}
  {\pi\left( \Sample|\prevTheta\right)} >1
  .
\end{equation*}

When using importance mixing in the context of NES, 
this reduces the sensitivity to the hyperparameters,
particularly the population size $\popsize$,
as importance mixing implicitly
adapts it to the situation by reusing some or many
previously evaluated sample points.

\subsection{Adaptation Sampling}
\label{sec:as}

To reduce the burden on determining appropriate
hyper-parameters such as the learning rate, we develop
a new self-adaptation or meta-learning technique~\citep{Schaul2010metalearning},
called \emph{adaptation sampling}, that can automatically adapt
the settings in a principled and economical way.

We model this situation as follows:
Let $\pi_{\theta }$ be a distribution with hyper-parameter $\theta$
and $\psi(\Sample)$ a quality measure for each sample $\Sample\thicksim \pi_{\theta}$.
Our goal is to adapt $\theta $ such as to maximize the quality $\psi$.
A straightforward method to achieve this, henceforth dubbed \emph{adaptation sampling},
is to evaluate the quality of the samples $\Sample'$ drawn from $\pi_{\prevTheta}$,
where  $\prevTheta \neq \theta $ is a slight variation of $\theta$,
and then perform hill-climbing: Continue with the new $\prevTheta$
if the quality of its samples is significantly better
(according, e.g., to a Mann-Whitney U-test), and revert to $\theta$ otherwise.
Note that this proceeding is similar to the NES algorithm itself,
but applied at a meta-level to algorithm parameters instead of the
search distribution. The goal of this adaptation is to maximize the
\emph{pace} of progress over time, which is slightly different from
maximizing the fitness function itself.

\emph{Virtual} adaptation sampling
is a lightweight alternative to adaptation sampling
that is particularly useful whenever evaluating $\psi$ is expensive :
\begin{itemize}
\item
	do importance sampling on the existing samples $\Sample_{i}$, according to 
	$\pi_{\prevTheta}$:
	\[
	w_{i}'=\frac{\pi(\Sample|\prevTheta)}{\pi(\Sample|\theta )}
	\]
	(this is always well-defined, because 
	$\Sample\thicksim \pi_{\theta }\Rightarrow \pi(\Sample|\theta)>0$).
\item
	compare $\{\psi (\Sample_{i})\}$ with weights $\{w_{i}=1,\forall i\}$ and 
	$\{\psi'=\psi (\Sample_{i}),\forall i\}$ with weights 
	$\{w'_{i}\}$, 
	using a weighted version of the Mann-Whitney test,
	as introduced in Appendix~A.
\end{itemize}

Beyond determining whether $\theta $ or $\prevTheta$ is better,
choosing a non-trivial confidence level $\rho$ allows us to avoid parameter
drift, as $\theta$ is only updated if the improvement is significant
enough.

There is one caveat, however: the rate of parameter change
needs to be adjusted such that
the two resulting distributions are not too similar (otherwise the
difference won't be statistically significant), but also not too different,
(otherwise the weights $w'$ will be too small and again the test will be inconclusive).

If, however, we explicitly desire large adaptation steps on $\theta$,
we have the possibility of interpolating between adaptation sampling
and virtual adaptation sampling by drawing a few new samples from the
distribution $\pi_{\prevTheta}$ (each assigned weight~1), where
it 
is overlapping least with $\pi_{\theta}$. The best way of achieving
this is importance mixing, as introduced in Section~\ref{sec:im},
uses jointly with the reweighted existing samples.

For NES algorithms, the most
important 
parameter to be adapted by
adaptation sampling is the learning rate $\eta$,
starting with a conservative guess.
This is because half-way into the search, after a local attractor
has been singled out, it may well pay off to increase the learning rate
in order to more quickly converge to it.

In order to produce variations $\eta'$ which can be judged using
the above-mentioned U-test, we propose a procedure similar in spirit to
Rprop-updates~\citep{riedmiller2002direct,rprop}, where the learning
rates are either increased or decreased by a multiplicative constant
whenever there is evidence that such a change will lead to better samples.

More concretely, when using adaptation sampling for NES we test for
an improvement with the hypothetical distribution $\prevTheta$
generated with $\eta' = 1.5 \eta$.
Each time the statistical test is successful with a confidence of at least
$\rho = \frac{1}{2}-\frac{1}{3(d+1)}$ (this value was determined empirically)
we increase the learning rate by a factor of $c^+=1.1$, up to at most $\eta=1$.
Otherwise we bring it closer to its initial value: $\eta \leftarrow 0.9\eta + 0.1\eta_{init}$.

\begin{figure}[ht]
	\centerline{
		\includegraphics[width=0.55\columnwidth, clip=true, trim=0.3cm 1.9cm 0.9cm 0.4cm]{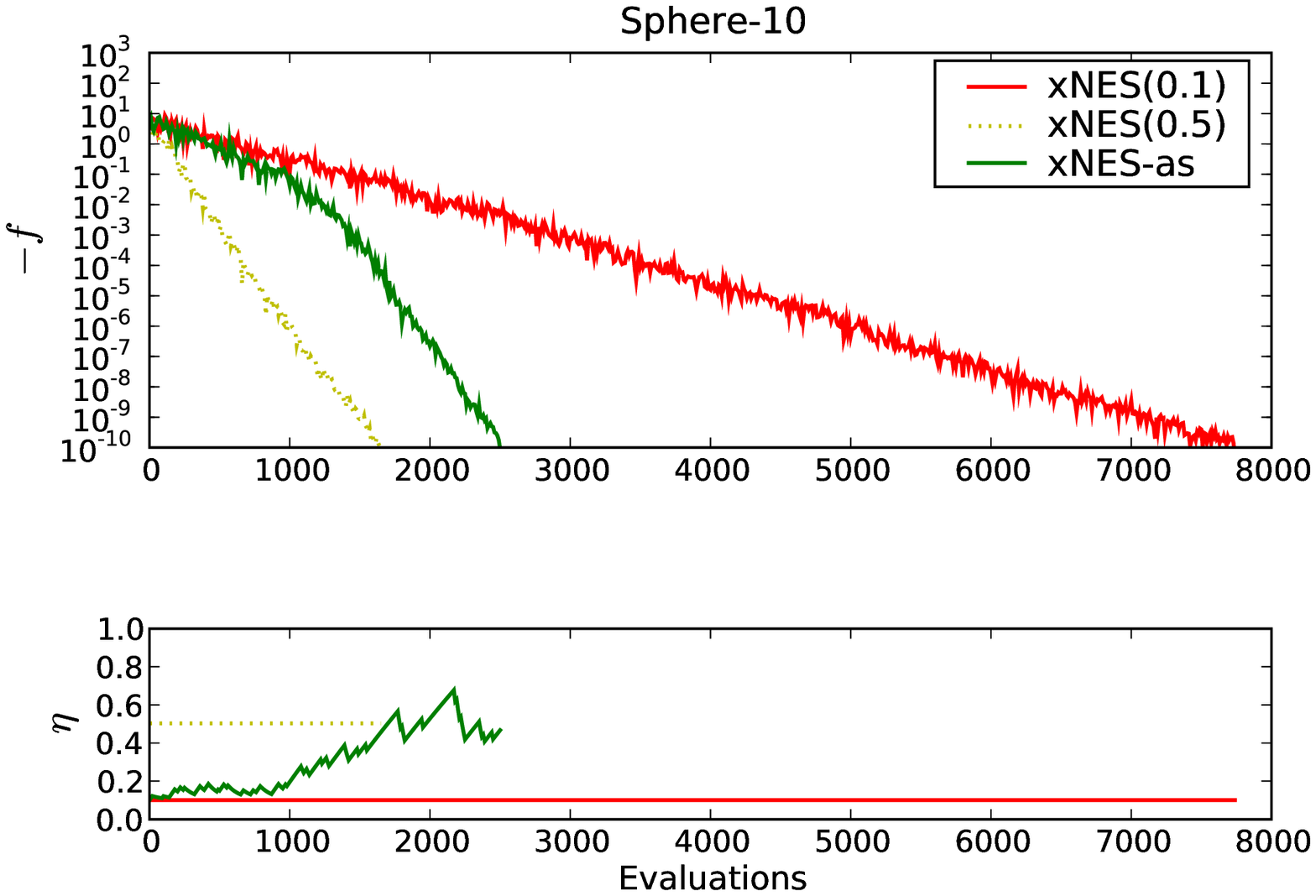}
		\includegraphics[width=0.55\columnwidth, clip=true, trim=0.3cm 1.9cm 0.8cm 0.4cm]{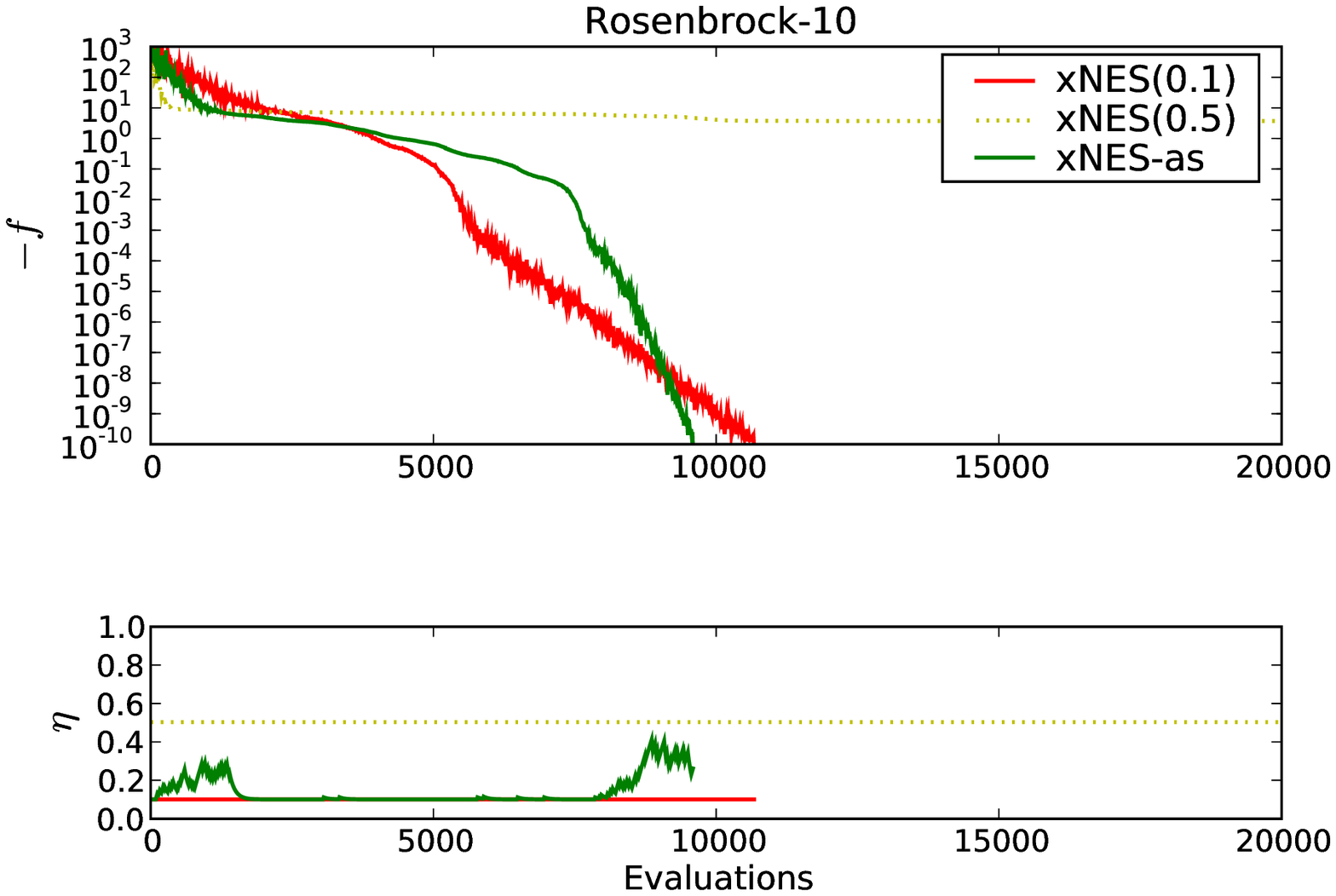}
	}
	\caption[Illustration of adaptation sampling]{\textbf{}Illustration of the effect of adaptation sampling. 
	We show the increase in fitness during a NES run (above) and the corresponding learning rates (below) 
	on two setups: 10-dimensional sphere function (left), and 10-dimensional Rosenbrock function (right).
	Plotted are three variants of xNES (algorithm~\ref{alg:xnes}): fixed default learning rate of $\eta=0.1$ (dashed, red)
	fixed large learning rate of $\eta=0.5$ (dotted, yellow), and an adaptive learning rate starting at $\eta=0.1$ (green).
	We see that for the (simple) Sphere function, it is advantageous to use a large learning rate, and adaptation sampling 
	automatically finds that one. However, using the overly greedy updates of a
	large learning rate fails on harder problems (right). Here adaptation sampling really shines: it boosts the 
	learning rate in the initial phase (entering the Rosenbrock valley), then quickly reduces it while the 
	search needs to carefully navigate the bottom of the valley, and boosts it again at the end when 
	it has located the optimum and merely needs to zoom in precisely.	
	}
	\label{fig:as-effect}
\end{figure}

Figure~\ref{fig:as-effect} illustrates the effect of the virtual
adaptation sampling strategy on two different 10-dimensional unimodal
benchmark functions, the Sphere function $f_1$ and the Rosenbrock
function $f_8$ (see section~\ref{sec:bbob} for details). We find that,
indeed, adaptation sampling boosts the learning rates to the appropriate
high values when quick progress can be made (in the presence of an
approximately quadratic optimum), but keeps them at carefully low values
otherwise.

\subsection{Restart Strategies}
\label{sec:restarts}

A simple but widespread method for mitigating the risk of finding only local optima 
in a strongly multi-modal scenario is to \emph{restart} the optimization algorithm a 
number of times with different initializations, or just with a different random seed.
This is even more useful in the presence of parallel processing resources, in which case
multiple runs are executed simultaneously.

In practice, where the parallel capacity is limited, we need 
to decide when to stop or suspend an ongoing 
optimization run and start or resume another one. 
In this section we provide one such restart strategy that 
takes those decisions. Inspired by recent work on practical universal search~\citep{schaul2010puns}, this
results in an effective use of resources independently of the problem.

\begin{figure}
	\centerline{
		\includegraphics[width=0.55\textwidth, clip=true, trim=0.3cm 0.3cm 0.4cm 0.6cm]{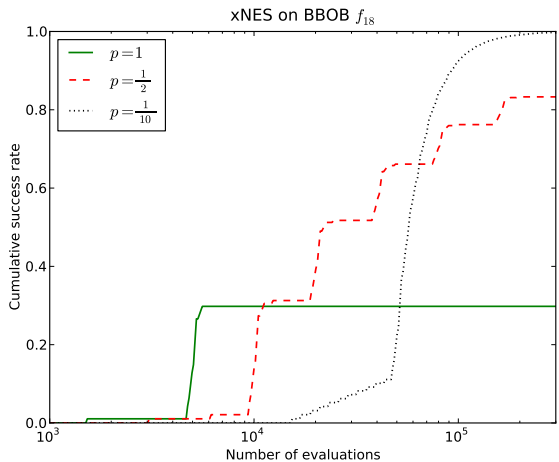}
	}
	\caption[Illustration of restart strategies]{\textbf{}Illustrating the effect of different restart strategies. 
	Plotted is the cumulative empirical 
	success probability, as a function of the total number of evaluations used. 
	Using no restarts, corresponding to $p=1$ (green) is initially faster but unreliable, 
	whereas $p=\frac{1}{10}$ (dotted black) reliably finds the optimum within 300000 evaluations, 
	but at the cost of slowing down success for all runs by a factor 10.
	In-between these extremes, $p=\frac{1}{2}$ (broken line, red) trades off slowdown and reliability.
	}
	\label{fig:repeat-illust}
\end{figure}

The strategy consists in reserving a fixed fraction $p$ of the total time for the first run, 
and then subdividing the remaining time $1-p$ in the same way, recursively 
(i.e. $p(1-p)^{i-1}$ for the $i^{th}$ run). The recursive decomposition of the time budget stops
when the assigned time-slice becomes smaller than the overhead of swapping out 
different runs. In this way, the number of runs with different seeds remains finite, 
but increases over time, as needed. 
Figure~\ref{fig:repeat-illust} illustrates the effect of the restart strategy, for different 
values of $p$, on the example of a multi-modal benchmark function
$f_{18}$ (see section~\ref{sec:bbob} for details), where most runs get caught in local optima.
Whenever used in the rest of the paper, 
the fraction is $p=\frac{1}{5}$.

Let $s(t)$ be the success probability of the underlying search algorithm
at time~$t$. Here, time is measured by the number of generations or
fitness evaluations. Accordingly, let $S_p(t)$ be the boosted success
probability of the restart scheme with parameter~$p$. Approximately,
that is, assuming continuous time, the probabilities are connected by
the formula
\begin{align*}
	S_p(t) = 1 - \prod_{i=1}^{\infty} \Big[ 1 - s \big( p (1-p)^{i-1} t \big) \Big] .
\end{align*}
Two qualitative properties of the restart strategy can be deduced from
this formula, even if in discrete time the sum is finite
($i \leq \log_2(t)$), and the times $p (1-p)^{i-1} t$ need to be
rounded:
\begin{itemize}
\item
	If there exists $t_0$ such that $s(t_0) > 0$ then 
	$\lim\limits_{t \to \infty} S_p(t) = 1$ for all $p \in (0, 1)$.
\item
	For sufficiently small~$t$, we have 
	$S_p(t) \leq s_p(t)$.
\end{itemize}
This captures the expected effect that the restart strategy results in
an initial slowdown, but eventually solves the problem reliably.

\section{Techniques for Multinormal Distributions}
\label{sec:multinormal}

In this section we will describe two crucial techniques to enhance performance of the NES algorithm as applied to 
multinormal distributions. First, the method of exponential parameterization is introduced, guaranteeing that  the covariance matrix
stays positive-definite. Second, a novel method for changing the coordinate system into a ``natural'' one is laid out, 
making the algorithm computationally efficient.

\subsection{Using Exponential Parameterization}
\label{sec:expmap}

Gradient steps on the covariance matrix $\Cov$ result in a number
of technical problems.
When updating $\Cov$ directly with the gradient step $\nabla_\Cov J$,
we have to ensure that $\Cov + \eta \nabla_\Cov J$ remains a valid, positive
definite covariance matrix. This is not guaranteed
\textit{a priori}, because the (natural) gradient $\nabla_\Cov J$ may
be any
symmetric matrix. If we instead update a factor $\sqCov$ of $\Cov$, it is at
least ensured that $\sqCov\transp \sqCov$ is symmetric and positive semi-definite. But
when shrinking an eigenvalue
of $\sqCov$ it may happen that the gradient step swaps the sign of the
eigenvalue, resulting in undesired oscillations.

An elegant way to fix these problems is to represent the covariance
matrix using the exponential map for symmetric matrices
(see e.g., \citealp{glasmachers:2005} for a related approach). Let
\begin{align*}
	\Symm_d := \Big\{\nsqCov \in \R^{d \times d} \,\Big|\, \nsqCov\transp = \nsqCov \Big\}
\end{align*}
and
\begin{align*}
	\PD_d := \Big\{\nsqCov \in \Symm_d \,\Big|\, \mathbf{v}\transp \nsqCov \mathbf{v} > 0 \text{ for all } \mathbf{v} \in \R^d \setminus \{0\} \Big\}
\end{align*}
denote the vector space of symmetric and the (cone) manifold of
symmetric positive definite matrices, respectively. Then the exponential
map
\begin{align}
	\exp : \Symm_d \to \PD_d \; , \qquad \nsqCov \mapsto \sum_{n=0}^{\infty} \frac{\nsqCov^n}{n!}
	\label{eq:exp}
\end{align}
is a diffeomorphism: The map is bijective, and both $\exp$ as well as
its inverse map $\log : \PD_d \to \Symm_d$ are smooth. The mapping can
be computed in cubic time, for example by decomposing the matrix
$\nsqCov = \mathbf{U}\mathbf{D}\mathbf{U}\transp$ into orthogonal $\mathbf{U}$
(eigen-vectors) and diagonal $\mathbf{D}$ (eigen-values), taking the
exponential of $\mathbf{D}$ (which amounts to taking the element-wise
exponentials of the diagonal entries), and composing everything back%
\footnote{The same computation works for the logarithm,
  and thus also for powers $\PD_d \to \PD_d$,
  $\nsqCov \mapsto \nsqCov^c = \exp(c \cdot \log(\nsqCov))$ for all
  $c \in \R$, for example for the (unique) square root ($c = 1/2$).}
as $\exp(\nsqCov) = \mathbf{U} \exp(\mathbf{D}) \mathbf{U}\transp$.

\begin{figure}
	\psfrag{exp}[c][c]{$\exp$}
	\psfrag{Sd}[c][c]{$\Symm_d$}
	\psfrag{Pd}[c][c]{$\PD_d$}
	\psfrag{M}[c][c]{$\nsqCov$}
	\psfrag{Sigma}[c][c]{$\Cov$}	
	\centerline{
		\includegraphics[height=0.31\columnwidth, clip=true, trim=4in 0 0 0]{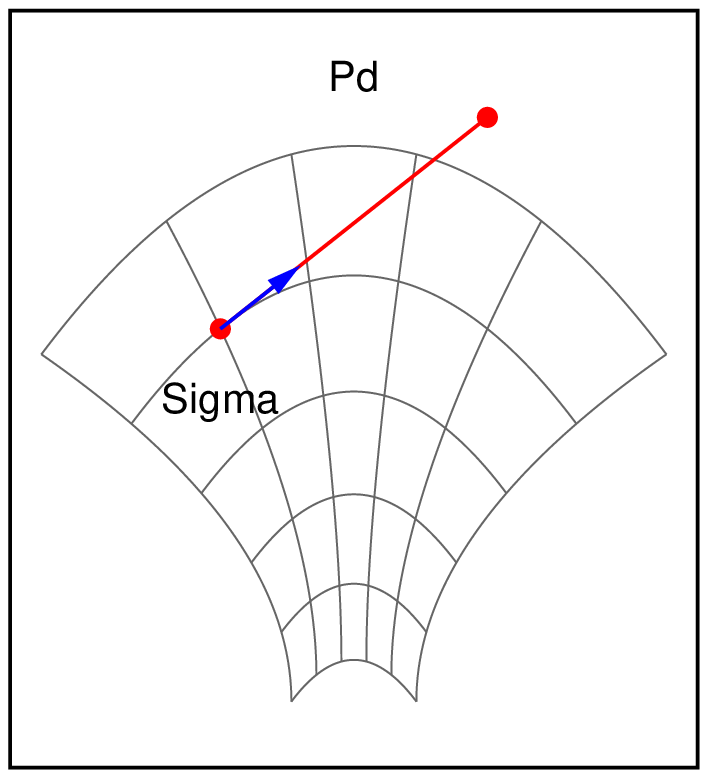}
		\includegraphics[height=0.31\columnwidth, clip=true, trim=0cm 0cm 0cm 0cm]{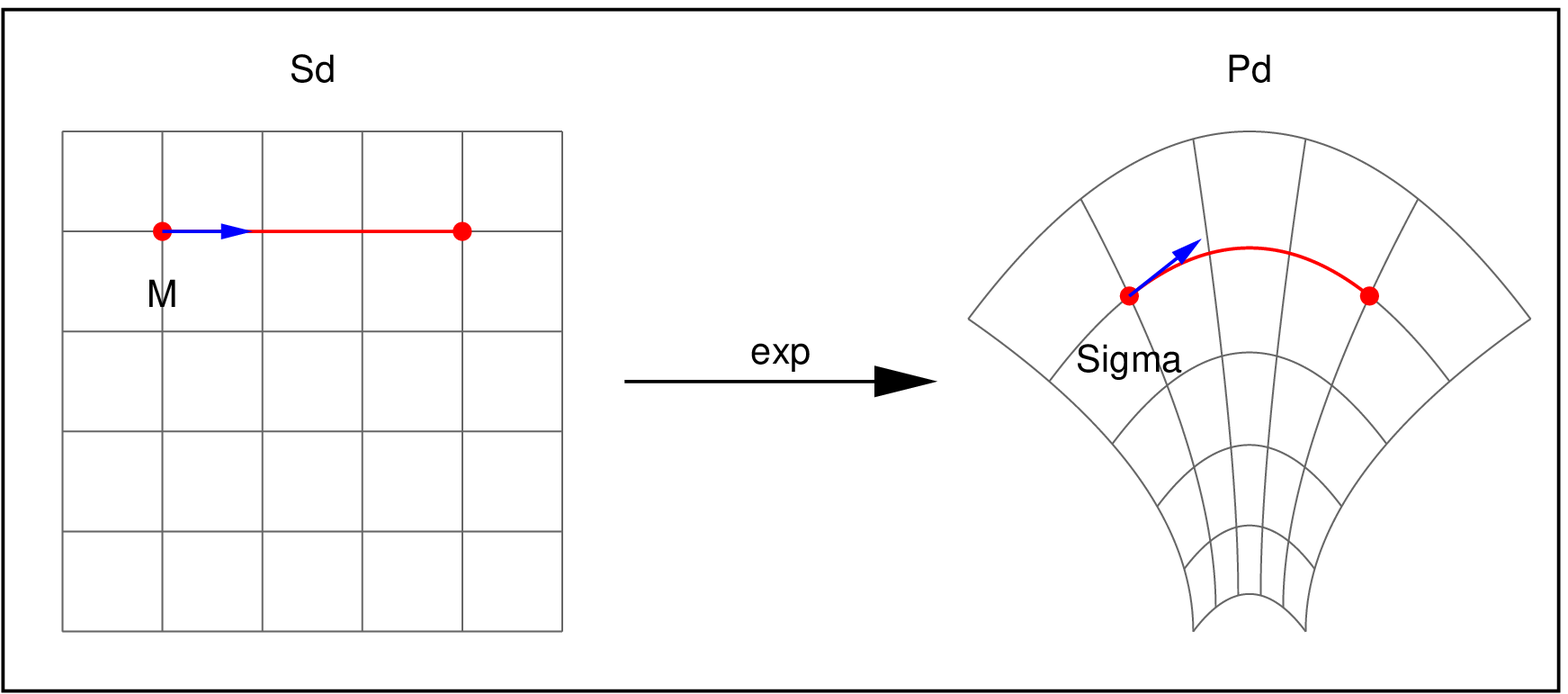}
	}
	\caption[Illustration of exponential map]{\textbf{Left:} 
	updating a covariance matrix $\Cov$ directly can end up outside the manifold 
	of symmetric positive-definite matrices $\PD_d$. 
	\textbf{Right:} first performing the update on 
	$\nsqCov = \frac{1}{2} \log(\Cov)$ in $\Symm_d$ and then 
	mapping back the result into the original space $\PD_d$ 
	using the exponential map is both \emph{safe} 
	(guaranteed to stay in the manifold) and \emph{straight} (the update follows a geodesic).
	}
	\label{fig:expmap}
\end{figure}

Thus, we can represent the covariance matrix $\Cov \in \PD_d$ as
$\exp(\nsqCov)$ with $\nsqCov \in \Symm_d$. The resulting gradient update
for $\nsqCov$ has two important properties: First, because
$\Symm_d$ is a \emph{vector space}, any update automatically corresponds
to a valid covariance matrix.%
\footnote{The tangent bundle $T \PD_d$ of the manifold $\PD_d$ is
  isomorphic to $\PD_d \times \Symm_d$ and globally trivial. Thus,
  arbitrarily large gradient steps are meaningful in this representation.}
Second, the update of $\nsqCov$ makes the gradient invariant
w.r.t.\ linear transformations of the search space $\R^d$. This follows
from an information geometric perspective, viewing $\PD_d$ as the
Riemannian parameter manifold equipped with the Fisher information
metric. The invariance property is a direct consequence of the
Cartan-Hadamard theorem~\citep{cartan:1928}. See also
Figure~\ref{fig:expmap} for an illustration.

However, the exponential parameterization considerably complicates the
computation of the Fisher information matrix~$\fisher$, which now involves
partial derivatives of the matrix exponential~\eqref{eq:exp}.
This can be done in cubic time per partial derivative according
to~\citep{najfeld:1994}, resulting in an unacceptable complexity of
$\Order(d^7)$ for the computation of the Fisher matrix.

\subsection{Using Natural Coordinates}
\label{sec:xnes}
\label{sec:gauss-natcoord}

In this section we describe a technique that allows us to 
avoid the computation of the Fisher information matrix altogether,
for some specific but common classes of distributions.
The idea is to iteratively change the coordinate system in such a way 
that it becomes possible to follow the natural gradient without any costly inverses of the 
Fisher matrix (actually, without even constructing it explicitly).
We introduce the technique for the simplest case of multinormal 
search distributions, and in section~\ref{sec:rotsym}, 
we generalize it to the whole class of distributions
that they are applicable to (namely, rotationally-symmetric distributions).

Instead of using the `global' coordinates $\Cov = \exp(\nsqCov)$ for
the covariance matrix, we linearly transform the coordinate system in
each iteration to a coordinate system in which the current search
distribution is the standard normal distribution with zero mean and unit
covariance. Let the current search distribution be given by
$(\Mean, \sqCov) \in \R^d \times \PD_d$ with $\sqCov\transp \sqCov = \Cov$. We use the tangent
space $T_{(\Mean,\sqCov)} (\R^d \times \PD_d)$ of the parameter manifold
$\R^d \times \PD_d$, which is isomorphic to the vector space
$\R^d \times \Symm_d$, to represent the updated search distribution as
\begin{align}
 	\left( \nCenter ,\, \nsqCov \right) \mapsto & \left( \Mean_{\text{new}} ,\, \sqCov_{\text{new}} \right)
	= \left( \Mean + \sqCov\transp \nCenter ,\, \sqCov \exp \left(\frac{1}{2} \nsqCov \right) \right) \label{eq:update}
 	.
\end{align}
This coordinate system is \emph{natural} in the sense that the Fisher
matrix w.r.t.\ an orthonormal basis of $(\nCenter, \nsqCov)$ is the identity
matrix. The current search distribution $\Normal(\Mean, \sqCov\transp \sqCov)$
is encoded as 
\begin{align*}
  \pi(\Sample|\nCenter, \nsqCov) 
  = \Normal \left(\Mean+\sqCov\transp\nCenter, \;
  \sqCov\transp\exp(\nsqCov)\sqCov
  \right)
  ,
\end{align*}
where at each step we change the coordinates such that $(\nCenter, \nsqCov) = (0, 0)$.
In this case, it is guaranteed that for the variables $(\nCenter, \nsqCov)$ 
the plain gradient and the natural gradient coincide ($\fisher=\idM$). Consequently the
computation of the natural gradient costs $\Order(d^3)$ operations. 
 
In the new coordinate system we produce standard normal samples $\nSample \sim \Normal(0, \idM)$
which are then mapped back into the original coordinates 
$\Sample = \Mean + \sqCov\transp \cdot  \nSample$. 
The log-density becomes 
\begin{align*}  
  \log \pi(\Sample \,|\, \nCenter, \nsqCov)=  
  &  -\frac{d}{2} \log(2 \pi) 
  - \log\big(\det(\sqCov)\big)
  -\frac{1}{2} \trace(\nsqCov) \\
  &-\frac{1}{2} \Big\| \exp\left(-\frac{1}{2} \nsqCov \right) 
  \sqCov^{-\top} \cdot (\Sample - (\Mean + \sqCov\transp \nCenter)) \Big\|^2 
  ,
\end{align*}
and thus the log-derivatives (at $\nCenter =0$, $\nsqCov = 0$) 
take the following, surprisingly simple forms:
\begin{align}  
 \nabla_{_{\nCenter}}|_{_{\nCenter=0}} \log\pi(\Sample \,|\, \nsqCov = 0, \nCenter)
   &= \nabla_{_{\nCenter}}|_{_{\nCenter=0}} \left[ -\frac{1}{2} \Big\| \sqCov^{-\top} 
\cdot (\Sample - (\Mean + \sqCov\transp\nCenter)) \Big\|^2 \right]\notag \\
   &= -\frac{1}{2} \left[- 2 \cdot \sqCov^{-\top} \sqCov\transp 
\cdot \sqCov^{-\top} \cdot (\Sample - (\Mean + \sqCov\transp\nCenter)) 
\right] \Big|_{_{\nCenter=0}} \notag\\
   &= \sqCov^{-\top} (\Sample - \Mean)\notag\\
   &= \nSample \label{eq:logderivdelta}\\
\nabla_{_{\nsqCov}}|_{_{\nsqCov=0}} \log\pi(\Sample \,|\, \nCenter = 0, \nsqCov)
   &=  -\frac{1}{2} \nabla_{_{\nsqCov}}|_{_{\nsqCov=0}} \Bigg[\trace(\nsqCov)
 +\Big\| \exp \left( -\frac{1}{2} \nsqCov \right) \sqCov^{-\top} 
(\Sample - \Mean) \Big\|^2 \Bigg] \notag\\
   &=  -\frac{1}{2} \left[ \idM+ 2 \cdot \left( -\frac{1}{2} \right) 
\cdot[\sqCov^{-\top} (\Sample - \Mean)] \cdot \exp \left( -\frac{1}{2} \nsqCov \right) 
\cdot [\sqCov^{-\top} (\Sample - \Mean)]\transp \right] \Bigg|_{_{\nsqCov=0}} \notag\\
   &=  -\frac{1}{2} \big[ \idM
-[\sqCov^{-\top} (\Sample - \Mean)] \cdot \idM
\cdot [\sqCov^{-\top} (\Sample - \Mean)]\transp \big] \notag\\
   &=  \frac{1}{2} (\nSample \nSample\transp - \idM) \label{eq:logderivM}
\end{align}
These results give us the updates in the natural coordinate system
\begin{align}
 \nabla_{\nCenter} J = & \sum_{k=1}^\popsize f(\Sample_k) \cdot \nSample_k  \label{eq:graddelta}\\
 \nabla_{\nsqCov}  J = & \sum_{k=1}^\popsize f(\Sample_k) \cdot (\nSample_k \nSample_k\transp - \idM) \label{eq:gradM}
\end{align}
which are then mapped back onto $(\Mean, \sqCov)$ using equation~\eqref{eq:update}:
\begin{align*}
  \Mean_{\text{new}} \leftarrow \Mean + \sqCov\transp \nCenter 
  & =\Mean + \eta \sqCov\transp  \nabla_{\nCenter} J \\
  & =\Mean + \eta \sqCov\transp \left(\frac{1}{\popsize} 
  \sum_{k=1}^{\popsize} \nabla_{\nCenter} \log\pi\left(\Sample_k | \theta\right)\cdot f(\Sample_k) \right)\\
  & =\Mean + \frac{\eta}{\popsize}  \sqCov\transp \left( 
  \sum_{k=1}^{\popsize}  f(\Sample_k) \cdot \nSample_k \right) \\
  \sqCov_{\text{new}} \leftarrow \sqCov \cdot\exp\left(\frac{1}{2}\nsqCov\right) 
  &= \sqCov \cdot\exp\left(\eta\frac{1}{2}\nabla_{\nsqCov} J \right)\\
  &= \sqCov \cdot\exp\left(\frac{\eta}{2} \cdot
  \frac{1}{\popsize}
\sum_{k=1}^{\popsize} 
\nabla_{\nsqCov} \log\pi\left(\Sample_k | \theta\right)
\cdot f(\Sample_k)  \right)\\
  &= \sqCov \cdot\exp\left(\frac{\eta}{4\popsize}
\sum_{k=1}^{\popsize} f(\Sample_k) \cdot
(\nSample_k \nSample\transp_k - \idM)\right)
\end{align*}

\subsection{Orthogonal Decomposition of Multinormal Parameter Space}
\label{sec:decomposition}
 
We decompose the parameter vector space $(\nCenter, \nsqCov) \in \R^d \times \Symm_d$
into the product 
\begin{align}
	\label{eq:decomposition}
	\R^d \times \Symm_d
			= \underbrace{\R^d}_{(\nCenter)}
		\times \underbrace{\Symm_d^{\parallel}}_{(\sigma)}
		\times \underbrace{\Symm_d^{\perp}}_{(\sqCovB)}
	\enspace,
\end{align}
of orthogonal subspaces. The one-dimensional space
$\Symm_d^{\parallel} = \{\lambda \cdot \idM \,|\, \lambda \in \R\}$ is
spanned by the identity matrix $\idM$, and
$\Symm_d^{\perp} = \{\nsqCov \in \Symm_d \,|\, \trace(\nsqCov) = 0\}$ denotes its
orthogonal complement in $\Symm_d$.
The different components have roles with clear interpretations:
The $(\nCenter)$-component $\nabla_{\nCenter} J$ describes
the update of the center of the search distribution,
the $(\sigma)$-component with value $\nabla_{\sigma} J \cdot \idM$ for
$\nabla_{\sigma} J = \trace(\nabla_\nsqCov J)/d$ has the role of a step size update, which
becomes clear from the identity $\det(\exp(\nsqCov)) = \exp(\trace(\nsqCov))$, and
the $(\sqCovB)$-component $\nabla_\sqCovB J = \nabla_\nsqCov J - \nabla_{\sigma} J$ 
describes the update of the transformation matrix, normalized to unit
determinant, which can thus be attributed to the shape of the search
distribution.
This decomposition is canonical in being the finest decomposition such
that updates of its components result in invariance of the search
algorithm under linear transformations of the search space.

On these subspaces we introduce independent learning rates $\eta_{\nCenter}$,
$\eta_{\sigma}$, and $\eta_\sqCovB$, respectively. For simplicity we also
split the transformation matrix $\sqCov = \sigma \cdot \sqCovB$ into the step size
$\sigma \in \R^+$ and the normalized transformation matrix $\sqCovB$ with
$\det(\sqCovB) = 1$. Then the resulting update is
\begin{align}
	\Mean_{\text{new}} & 
	=\Mean +\eta_{\nCenter} \cdot \nabla_{\nCenter} J = \Mean + \eta_{\nCenter} \cdot \sum_{k=1}^\popsize f(\Sample_k) \cdot \nSample_k \label{eq:update-mu} \\ 	
	\sigma_{\text{new}} & 
	= \sigma \cdot \exp \left( \frac{\eta_{\sigma}}{2} \cdot \nabla_{\sigma} \right) 
	= \sigma \cdot \exp \left( \frac{\eta_{\sigma}}{2} \cdot \frac{\trace(\nabla_\nsqCov J)}{d} \right) 
	\label{eq:update-sigma} \\ 
	\sqCovB_{\text{new}} & = \sqCovB \cdot \exp \left( \frac{\eta_\sqCovB}{2} \cdot \nabla_\sqCovB J \right) 
	= \sqCovB \cdot \exp \left( \frac{\eta_\sqCovB}{2} \cdot 
	\left(\nabla_\nsqCov J - \frac{\trace(\nabla_\nsqCov J)}{d} \cdot \idM \right)\right)\label{eq:update-B} 
	,
\end{align}
with $\nabla_\nsqCov J$ from equation~\eqref{eq:gradM}.
In case of $\eta_{\sigma} = \eta_\sqCovB$, in this case referred to as
$\eta_\sqCov$, the updates \eqref{eq:update-sigma} and \eqref{eq:update-B}
simplify to
\begin{align}
	\sqCov_{\text{new}} & = \sqCov \cdot \exp \left( \frac{\eta_\sqCov}{2} \cdot \nabla_\nsqCov \right) \label{eq:update-A} \\
		& = \sqCov \cdot \exp \left( \frac{\eta_\sqCov}{2} \cdot \sum_{k=1}^\popsize f(\Sample_k) \cdot (\nSample_k \nSample_k\transp - \idM) \right) \notag
.
\end{align}
The resulting algorithm is called \emph{exponential} NES (xNES), and shown in Algorithm~\ref{alg:xnes}.
We also give the pseudocode for its hill-climber variant (see also section~\ref{sec:elitism}).

Updating the search distribution in the natural coordinate system is an
alternative to the exponential parameterization (section~\ref{sec:expmap})
for making the algorithm invariant under linear transformations of the
search space, which is then achieved in a direct and constructive way.

\begin{algorithm}
\DontPrintSemicolon
\SetKwInOut{Input}{input}
\caption{Exponential Natural Evolution Strategies (xNES), for multinormal distributions}
\label{alg:xnes}
 \Input{$f$, $\Mean_{init}$, $\Cov_{init}=\sqCov\transp\sqCov$}
\vspace{0.4cm}
 initialize 
 $\begin{array}{l}
 \sigma \leftarrow \sqrt[d]{|\det(\sqCov)|} \\
 \sqCovB \leftarrow \sqCov / \sigma
\end{array}$ \\
 \Repeat{stopping criterion is met}{
 \For{$k=1\ldots\popsize$}{
  draw sample $\nSample_k \sim \Normal(0, \idM)$\\
  $\Sample_k \leftarrow \Mean + \sigma \sqCovB\transp\nSample_k$\\
  evaluate the fitness $f(\Sample_k)$\\
}
sort $\{(\nSample_k, \Sample_k)\}$ with respect to $f(\Sample_k)$ and compute utilities $u_k$ \\
\vspace{0.4cm}
		compute gradients
$\begin{array}{lll}  
    \nabla_{\nCenter} J \leftarrow \sum_{k=1}^{\popsize} u_k \cdot \nSample_k & &
		\nabla_\nsqCov J    \leftarrow \sum_{k=1}^{\popsize} u_k \cdot (\nSample_k \nSample_k\transp - \idM) \\
	  \nabla_{\sigma} J   \leftarrow \trace(\nabla_\nsqCov J) / d & &
		\nabla_\sqCovB J \leftarrow \nabla_\nsqCov J - \nabla_{\sigma} J \cdot \idM 
\end{array}$ \\
\vspace{0.4cm}
update parameters
$\begin{array}{lll}
	\Mean \leftarrow \Mean + \eta_{\nCenter} \cdot \sigma \sqCovB \cdot \nabla_{\nCenter} J \\
	\sigma \leftarrow \sigma \cdot \exp(\eta_{\sigma}/2 \cdot \nabla_{\sigma} J) \\
	\sqCovB \leftarrow \sqCovB \cdot \exp(\eta_\sqCovB/2 \cdot \nabla_\sqCovB J)
\end{array} $\\
 }
\end{algorithm}

\begin{algorithm}
\caption{(1+1)-xNES}
\label{alg:hillclimber-xnes}
\DontPrintSemicolon
\SetKwInOut{Input}{input}
 \Input{$f$, $\Mean_{init}$, $\Cov_{init}=\sqCov\transp\sqCov$}
 $f_{max} \leftarrow - \infty$\\
 \Repeat{stopping criterion is met}{
 
  draw sample $\nSample \sim \Normal(0, \idM)$\\
  evaluate the fitness $f(\Sample=\sqCov\transp\nSample+\Mean)$\\
  calculate log-derivatives $\nabla_{\nsqCov} \log \pi\left(\Sample |\theta\right)
  = \frac{1}{2}\left(\nSample \nSample\transp -\idM\right)$\\
 	$(\Sample_1, \Sample_2) \leftarrow (\Mean, \Sample)$\\
  \eIf{$f_{max} < f(\Sample)$}
 	{$f_{max} \leftarrow f(\Sample)$\\
 	$\mathbf{u} \leftarrow (-4, 1)$\\
 	update mean $\Mean \leftarrow \Sample$\\
 	}
 	{$\mathbf{u} \leftarrow (\frac{4}{5}, 0)$\\
 	} 	
\vspace{0.1cm}
$\nabla_{\nsqCov} J \leftarrow \frac{1}{2}
\sum_{k=1}^{2} 
\nabla_{\nsqCov} \log\pi\left(\Sample_k | \theta\right)
\cdot u_k = -\frac{u_1}{2} \idM+\frac{u_2}{4}\left(\nSample \nSample\transp -\idM\right) 
$\\
\vspace{0.1cm}
$ \sqCov \leftarrow \sqCov \cdot\exp\left(\frac{1}{2}\eta \nabla_{\nsqCov} J\right)$\\
 }
\end{algorithm}

\subsection{Connection to CMA-ES.}
\label{sec:cmalike}

It has been noticed independently by
\citep{Glasmachers2010} and shortly afterwards by \citep{Akimoto2010ppsn}
that the natural gradient updates of xNES and the strategy updates of
the CMA-ES algorithm \citep{hansen:2001} are closely connected. 
However, since xNES does not feature evolution paths, this connection is
restricted to the so-called rank-$\mu$-update (in the terminology of
this study, rank-$\popsize$-update) of CMA-ES.

First of all observe that xNES and CMA-ES share the same invariance
properties. But more interestingly, although derived from different
heuristics, their updates turn out to be nearly equivalent. A closer
investigation of this equivalence promises synergies and new
perspectives on the working principles of both algorithms.
In particular, this insight shows that CMA-ES can be explained as a
natural gradient algorithm, which may allow for a more thorough analysis
of its updates, and xNES can profit from CMA-ES's mature settings of
algorithms parameters, such as search space dimension-dependent
population sizes, learning rates and utility values.

Both xNES and CMA-ES parameterize the search distribution with three
functionally different parameters for mean, scale, and shape of the
distribution. xNES uses the parameters $\Mean$, $\sigma$, and $\sqCovB$,
while the covariance matrix is represented as
$\sigma^2 \cdot \mathbf{C}$ in CMA-ES, where $\mathbf{C}$ can by any
positive definite symmetric matrix. Thus, the representation of the
scale of the search distribution is shared among $\sigma$ and
$\mathbf{C}$ in CMA-ES, and the role of the additional parameter
$\sigma$ is to allow for an adaptation of the step size on a faster time
scale than the full covariance update. In contrast, the NES updates
of scale and shape parameters $\sigma$ and $\sqCovB$ are properly
decoupled.

Let us start with the updates of the center parameter $\Mean$.
The update \eqref{eq:update-mu} is very similar to the update of the
center of the search distribution in CMA-ES, see \citep{hansen:2001}. The
utility function exactly takes the role of the weights in CMA-ES, which
assumes a fixed learning rate of one.

For the covariance matrix, the situation is more complicated.
From equation~\eqref{eq:update-A} we deduce the update rule
\begin{align*}
	\Cov_{\text{new}} = & (\sqCov_{\text{new}})\transp \cdot \sqCov_{\text{new}} \\
		= & \sqCov\transp \cdot \exp \left( \eta_\Cov \cdot \sum_{k=1}^\popsize u_k \left(\nSample_k \nSample_k\transp - \idM \right) \right) \cdot \sqCov
\end{align*}
for the covariance matrix, with learning rate $\eta_\Cov = \eta_\sqCov$.
This term is closely connected to the exponential parameterization of
the natural coordinates in xNES, while CMA-ES is formulated in global
linear coordinates. The connection of these updates can be shown either
by applying the xNES update directly to the natural coordinates without
the exponential parameterization, or by approximating the exponential
map by its first order Taylor expansion. \citep{Akimoto2010ppsn}
established the same connection directly in coordinates based on the
Cholesky decomposition of $\Cov$, see \citep{sun:2009a,sun:2009b}.
The arguably simplest derivation of the equivalence relies on the
invariance of the natural gradient under coordinate transformations,
which allows us to perform the computation, w.l.o.g., in natural
coordinates. We use the first order Taylor approximation of $\exp$ to
obtain
\begin{align*}
	\exp \left( \eta_\Cov \cdot \sum_{k=1}^\popsize u_k \left( \nSample_k \nSample_k\transp - \idM \right) \right) \approx \idM + \eta_\Cov \cdot \sum_{k=1}^\popsize u_k \left( \nSample_k \nSample_k\transp - \idM \right)
	\enspace,
\end{align*}
so the first order approximate update yields
\begin{align*}
	\Cov_{new}^{\prime } = & \sqCov\transp \cdot \left(\idM + \eta_\Cov \cdot \sum_{k=1}^\popsize u_k \left(\nSample_k \nSample_k\transp - \idM \right) \right) \cdot \sqCov \\
		= & \left( 1 - U \cdot \eta_\Cov \right) \cdot \sqCov\transp \sqCov + \eta_\Cov \cdot \sum_{k=1}^\popsize u_k \left( \sqCov\transp \nSample_k \right) \left( \sqCov\transp \nSample_k \right)\transp \\
		= & \left( 1 - U \cdot \eta_\Cov \right) \cdot \Cov + \eta_\Cov \cdot \sum_{k=1}^\popsize u_k \left( \Sample_k - \Mean \right) \left( \Sample_k - \Mean \right)\transp
\end{align*}
with $U = \sum_{k=1}^\popsize u_k$, from which the connection to the
CMA-ES rank-$\mu$-update is obvious (see \citealp{hansen:2001}).

Finally, the updates of the global step size parameter $\sigma$ turn
out to be identical in xNES and CMA-ES without evolution paths.

Having highlighted the similarities, let us have a closer look at the
differences between xNES and CMA-ES, which are mostly two aspects.
CMA-ES uses the well-established technique of evolution paths to
smoothen out random effects over multiple generations. This technique
is particularly valuable when working with minimal population sizes,
which is the default for both algorithms. Thus, evolution paths are
expected to improve stability; further interpretations have been
provided by \citep{hansen:2001}. However, the presence of evolution paths
has the conceptual disadvantage that the state of the CMA-ES algorithms
is not completely described by its search distribution. The other
difference between xNES and CMA-ES is the exponential parameterization of
the updates in xNES, which results in a multiplicative update equation
for the covariance matrix, in contrast to the additive update of CMA-ES.
We argue that just like for the global step size $\sigma$, the
multiplicative update of the covariance matrix is natural.

A valuable perspective offered by the natural gradient updates in xNES
is the derivation of the updates of the center~$\Mean$, the step size~$\sigma$,
and the normalized transformation matrix~$\sqCovB$, all from the \emph{same}
principle of natural gradient ascent. In contrast, the updates applied
in CMA-ES result from different heuristics for each parameter. Hence, it is
even more surprising that the two algorithms are closely connected. This
connection provides a post-hoc justification of the various heuristics
employed by CMA-ES, and it highlights the consistency of the intuition
behind these heuristics.

\subsection{Elitism}
\label{sec:elitism}
The principle of the NES
algorithm is to follow the natural gradient of expected fitness.
This requires
sampling the fitness gradient. 
Naturally, this amounts to what, within the realm of evolution strategies, 
is generally referred to as
``comma-selection'', that is, updating the search distribution based solely
on the current batch of ``offspring'' samples, disregarding older samples
such as the ``parent'' population. This seems to exclude approaches that retain
some of the best samples, like elitism,
hill-climbing, or even steady-state selection~\citep{Goldberg}. In this section we show
that elitism (and thus a wide variety of selection schemes) is indeed
compatible with NES. We exemplify this technique by deriving a NES
algorithm with (1+1) selection, i.e., a hill-climber (see also
\citealp{Glasmachers2010}).

It is impossible to estimate any information about the fitness gradient
from a single sample, since at least two samples are required to
estimate even a finite difference. The (1+1) selection scheme indicates
that this dilemma can be resolved by considering two distinguished
samples, namely the elitist or parent $\Sample^\text{parent} = \Mean$,
and the offspring $\Sample$. Considering these two samples in the update
is in principle sufficient for estimating the fitness gradient w.r.t.\
the parameters~$\theta$.

Care needs to be taken for setting the algorithm parameters, such as
learning rates and utility values. The extremely small population size of
one indicates that learning rates should generally be small in order to
ensure stability of the algorithm. Another guideline is the well known
observation \citep{RechenbergES} that a step size resulting in a
success rate of roughly $1/5$ maximizes progress. This indicates that a
self-adaptation strategy should increase the learning rate in case of
too many successes, and decrease it when observing too few successes.

Let us consider the basic case of \emph{radial} Gaussian search distributions
\begin{align*}
	\pi(\Sample \,|\, \Mean, \sigma) = \frac{1}{\sqrt{2\pi}\sigma} \exp\left(\frac{\|\Sample - \Mean\|^2}{2\sigma^2}\right)
\end{align*}
with parameters $\Mean \in \R^d$ and $\sigma > 0$. We encode these
parameters as $\theta = (\Mean, \ell)$ with $\ell = \log(\sigma)$. Let
$\nSample \sim \Normal(0, 1)$ be a standard normally distributed
vector, then we obtain the offspring as
$\Sample = \Mean + \sigma \cdot \nSample \sim \Normal(\Mean, \sigma)$,
and the natural gradient components are
\begin{align*}
	\tilde \nabla_{\Mean} J & = u_1^{(\Mean)} \cdot \mathbf{0} + u_2^{(\Mean)} \cdot \sigma \cdot \nSample \\
	\tilde \nabla_{\ell} J & = u_1^{(\ell)} \cdot (-1) + u_2^{(\ell)} \cdot (\|\nSample\|^2 - 1) .
\end{align*}
The corresponding strategy parameter updates read
\begin{align*}
	\Mean & \leftarrow \Mean + \eta_{\Mean} \cdot \left[ u_1^{(\Mean)} \cdot \mathbf{0} + u_2^{(\Mean)} \cdot \sigma \cdot \nSample \right] \\
	\sigma & \leftarrow \sigma \cdot \exp \left( \eta_{\ell} \cdot \left[ u_1^{(\ell)} \cdot (-1) + u_2^{(\ell)} \cdot (\|\nSample\|^2 - 1) \right] \right) .
\end{align*}
The indices $1$ and $2$ of the utility values refer to the `samples'
$\Mean$ and $\Sample$, namely parent and offspring. Note that these
samples correspond to the vectors $\mathbf{0}$ and $\nSample$ in the
above update equations (see also section~\ref{sec:gauss-natcoord}). The
superscripts of the utility values indicate the different parameters.
Now elitist selection and the $1/5$ rule dictate the settings of these
utility values as follows:
\begin{itemize}
\item
	The elitist rule requires that the mean
	remains unchanged in case of no success ($u_1^{(\Mean)} = 1$ and
	$u_2^{(\Mean)} = 0$), and that the new sample replaces the mean in
	case of success ($u_1^{(\Mean)} = 0$ and $u_2^{(\Mean)} = 1$, with a
	learning rate of $\eta_{\Mean} = 1$).
\item
	Setting the utilities for $\ell$ to $u_1^{(\ell)} = 1$ and
	$u_2^{(\ell)} = 0$ in case of no success,
	effectively reduces the learning rate. Setting $u_1^{(\ell)} = -5$
	and $u_2^{(\ell)} = 0$ in case of success has the opposite effect
	and roughly implements the $1/5$-rule. The self-adaptation process
	can be stabilized with a small learning rate~$\eta_{\ell}$.
\end{itemize}
Note that we change the utility values based on success or failure of
the offspring. This seems natural, since the utility of information
encoded in the sample $\Sample$ depends on its success.
Highlighting elitism in the selection, we call these utility values
success-based. This is similar but not equivalent to rank-based
utilities for the joint population $\{\Mean, \Sample\}$. 

The NES hill-climber for radial Gaussian search distributions is
illustrated in algorithm~\ref{alg:(1+1)-nes}. This formulation offers a
more standard perspective on the hill-climber by using explicit case
distinctions for success and failure of the offspring instead of
success-based utilities.
The same procedure can be generalized to more flexible search
distributions. A conservative strategy is to update further
shape-related parameters only in case of success, which can be expressed
by means of success-based utility values in the very same way.
The corresponding algorithm for multi-variate Gaussian search
distributions is Algorithm~\ref{alg:hillclimber-xnes} (in section~\ref{sec:xnes})
and for multi-variate Cauchy it is Algorithm~\ref{alg:cauchy-xnes} (in section~\ref{sec:cauchy-nes}).

\begin{algorithm}
\SetKwInOut{Input}{input}
\caption{(1+1)-NES with radial Gaussian distribution}
\label{alg:(1+1)-nes}
 \Input{$f$, $\Mean_{init}$, $\sigma_{init}$}
 $f_{\text{best}} \leftarrow f(\Mean_{init})$\\
 \Repeat{stopping criterion is met}{
  draw sample $\nSample \sim \Normal(0,1)$ \\
  create offspring $\Sample \leftarrow \Mean + \sigma \cdot \nSample$ \\
  evaluate the fitness $f(\Sample)$\\
  \eIf{$f(\Sample) > f_{\text{best}}$}
  {
 	update mean $\Mean \leftarrow \Sample$\\
 	$\sigma \leftarrow \sigma \cdot \exp(5 \eta_{\sigma})$\\
 	$f_{\text{best}} \leftarrow f(\Sample)$\\
  }
  {
	$\sigma \leftarrow \sigma \cdot \exp(-\eta_{\sigma})$\\
  }
}
\end{algorithm}

\section{Beyond Multinormal Distributions}
\label{sec:other-dists}
In the previous section we have seen how natural gradient ascent is
applied to multi-variate normal distributions, which arguably constitute
the most important class of search distributions in modern evolution
strategies. 
In this section we expand the NES framework in breadth, 
motivating the usefulness and deriving 
a number of NES variants with different search distributions.

\subsection{Separable NES}
\label{sec:snes}
Adapting the full covariance matrix of the search
distribution can be disadvantageous, particularly in high-dimensional
search spaces, for two reasons.

For many problems it can be safely assumed that the computational costs are
governed by the number of fitness evaluations. This is particularly true
if such evaluations rely on expensive simulations. However, for
applications where fitness evaluations scale
gracefully with the search space dimension, the $\Order(d^3)$ xNES
update (due to the matrix exponential)
can dominate the computation. One such application is the evolutionary
training of recurrent neural networks (i.e., neuroevolution), where the
number of weights $d$ in the network can grow quadratically with the
number of neurons $n$, resulting in a complexity of $\Order(n^6)$ for a
single NES update.

A second reason not to adapt the full covariance matrix in high
dimensional search spaces is sample efficiency. The covariance matrix
has $d(d+1)/2 \in \Order(d^2)$ degrees of freedom, which can be  huge
 in large dimensions. Obtaining a stable estimate of this matrix
based on samples may thus require many (costly) fitness
evaluations, in turn requiring very small learning rates. As a result, the
algorithm may simply not have enough time to adapt its search
distribution to the problem with a given budget of fitness
evaluations. In this case, it may be advantageous to restrict the class
of search distributions in order to adapt at all, even
if this results in a less steep learning curve in the (then practically
irrelevant) limit of infinitely many fitness evaluations.

The only two distinguished parameter subsets of a multi-variate
distribution that do not impose the choice of a particular coordinate
system onto our search space are the `size' $\sigma$ of the distribution,
corresponding to the $(2d)$-th root of the determinant of the covariance
matrix, and its orthogonal complement, the covariance matrix normalized
to constant determinant $\sqCovB$ (see section~\ref{sec:decomposition}).
The first of these candidates results in a standard evolution strategy
without covariance adaptation at all, which may indeed be a viable
option in some applications, but is often too inflexible. The set of
normalized covariance matrices, on the other hand, is not interesting
because it is clear that the size of the distribution needs
to be adjusted in order to ensure convergence to an optimum.

Thus, it has been proposed to give up some invariance properties of the
search algorithm, and to adapt the class of search distribution with
diagonal covariance matrices \emph{in some predetermined coordinate
system}~\citep{ros:2008}. Such a choice is justified in many
applications where a certain degree of independence can be assumed
among the fitness function parameters. It has
even been shown in \citep{ros:2008} that this approach can work
surprisingly well even for highly non-separable fitness functions.

Restricting the class of search distributions to Gaussians with diagonal
covariance matrix corresponds to restricting a general class of
multi-variate search distributions to separable distributions
\begin{align*}
	p(\Sample \,|\, \theta) = \prod_{i=1}^d \tilde p(\Sample_i \,|\, \theta_i)
	\enspace,
\end{align*}
where $\tilde p$ is a family of densities on the reals, and
$\theta = (\theta_1, \dots, \theta_d)$ collects the parameters of all of
these distributions. 
In most cases
these parameters amount to $\theta_i = (\Mean_i, \covs_i)$, where
$\Mean_i \in \R$ is a position and $\covs_i \in \R^+$ is a scale
parameter (i.e., mean and standard deviation, if they exist), such that
$\Sample_i = \Mean_i + \covs_i \cdot \nSample_i \sim \tilde p(\cdot \,|\, \Mean_i, \covs_i)$
for $\nSample_i \sim \tilde p(\cdot \,|\, 0, 1)$.

\begin{algorithm}
\SetKwInOut{Input}{input}
\caption{Separable NES (SNES)}
\label{alg:snes}
 \Input{$f$, $\Mean_{init}$, $\covs_{init}$}
 \Repeat{stopping criterion is met}{
 \For{$k=1\ldots\popsize$}{
  draw sample $\nSample_k \sim \Normal(0, \idM)$\\
  $\Sample_k \leftarrow \Mean + \covs \nSample_k$\\
  evaluate the fitness $f(\Sample_k)$\\
}
sort $\{(\nSample_k, \Sample_k)\}$ with respect to $f(\Sample_k)$ and compute utilities $u_k$ \\
\vspace{0.2cm}
		compute gradients
$\begin{array}{lll}  
    \nabla_{\Mean} J \leftarrow \sum_{k=1}^{\popsize} u_k \cdot \nSample_k \\
		\nabla_{\covs} J \leftarrow \sum_{k=1}^{\popsize} u_k \cdot (\nSample_k^2 - 1) 
\end{array}$ \\
\vspace{0.2cm}
update parameters
$\begin{array}{lll}
	\Mean \leftarrow \Mean + \eta_{\Mean} \cdot \covs \cdot \nabla_{\Mean} J \\
	\covs \leftarrow \covs \cdot \exp(\eta_{\covs}/2 \cdot \nabla_{\covs} J) 
\end{array} $\\
 }
\end{algorithm}

Obviously, this allows us to sample new offspring in $\Order(d)$
time (per sample). Since the adaptation of each component's parameters is
independent, the strategy update step also
takes only $\Order(d)$ time. 

Thus, the sacrifice of invariance, amounting to the selection of a
distinguished coordinate system, allows for a linear time algorithm
(per individual), that still maintains a reasonable amount of
flexibility in the search distribution, and allows for a considerably
faster adaptation of its parameters. The resulting NES
variant, called \emph{separable}~NES (SNES), is illustrated in
algorithm~\ref{alg:snes} for Gaussian search distributions. Note that
each of the steps requires only $\Order(d)$ operations. Later in this
section we extend this algorithm to other search distributions, without
affecting its computational complexity.

\subsection{Rotationally-symmetric Distributions}
\label{sec:rotsym}

The class of \emph{radial} or rotationally-symmetric search
distributions are distributions with the property $p(x) = p(Ux)$, for
all $x \in \R^d$ and all orthogonal matrices $U \in \R^{d \times d}$.
Let $Q_{\Tail}(\Sample)$ be a family of densities of a rotationally symmetric
probability distributions in $\R^d$ with parameter $\Tail$. Thus, we can
write $Q_{\Tail}(\Sample) = q_{\Tail}(r^2)$ with $r^2 = \|\Sample\|^2$
for some family of functions $q_{\Tail} : \R^{\geq 0} \to \R^{\geq 0}$.
In the following we consider classes of search distributions with
densities
\begin{align*}
	\pi \big( \Sample \,\big|\, \Mean, \sqCov, \Tail \big) = & \frac{1}{\det(\sqCov)}
	\cdot q_{\Tail} \big( \|\left(\sqCov^{-1}\right)^\top (\Sample - \Mean)\|^2 \big)
\end{align*}
with additional transformation parameters $\Mean \in \R^d$ and
invertible $\sqCov \in \R^{d \times d}$. The function $q_{\Tail}$
is the accordingly transformed density of the variable
$\nSample = \left(\sqCov^{-1}\right)^\top (\Sample - \Mean)$.
This setting is rather general. It covers many important families of
distributions and their multi-variate forms, such as multi-variate
Gaussians. In addition, parameters of the radial distribution, most
prominently its tail (controlling whether large mutations are common or
rare) can be controlled with the parameter~$\Tail$.

We apply the procedure presented in section~\ref{sec:xnes} to this more
general case. In local exponential coordinates
\begin{align*}
	(\nCenter, \nsqCov) \mapsto (\Mean_{\text{new}}, \sqCov_{\text{new}}) = \left( \Mean + \sqCov\transp \nCenter, \sqCov \exp \left(\frac{1}{2} \nsqCov \right) \right)
\end{align*}
we obtain the three components of log-derivatives
\[
\nabla_{\nCenter,\nsqCov,\Tail}|_{_{\nCenter=0,\nsqCov=0}} \log\pi\left(\Sample \,|\, \Mean, \sqCov, \Tail, \nCenter, \nsqCov\right) = \left(g_{\nCenter}, g_\nsqCov, g_{\Tail}\right)
,
\]
\begin{align*}
	g_{\nCenter} = & -2 \cdot \frac{q_{\Tail}'(\|\nSample\|^2)}{q_{\Tail}(\|\nSample\|^2)} \cdot \nSample \\
	g_{\nsqCov} = & -\frac{1}{2} \idM - \frac{q_{\Tail}'(\|\nSample\|^2)}{q_{\Tail}(\|\nSample\|^2)} \cdot \nSample \nSample^{\top} \\
	g_{\Tail} = & \frac{1}{q_{\Tail}(\|\nSample\|^2)} \cdot \nabla_{\Tail} \, q_{\Tail}(\|\nSample\|^2)
\end{align*}
where
$q_{\Tail}' = \frac{\partial}{\partial (r^2)} q_{\Tail}$
denotes the derivative of $q_{\Tail}$ with respect to $r^2$, and
$\nabla_{\Tail} \, q_{\Tail}$ denotes the gradient w.r.t.~$\Tail$.

In the special case of Gaussian search distributions,
$q$ does not depend on a parameter $\Tail$ and we have
\begin{align*}
	q(r^2) = & \frac{1}{(2\pi)^{d/2}} \cdot \exp \left( -\frac{1}{2} r^2 \right) \\
	q'(r^2) = & -\frac{1}{2} \cdot \frac{1}{(2\pi)^{d/2}} \cdot \exp \left( -\frac{1}{2} r^2 \right)
		= -\frac{1}{2} \cdot q(r^2)
	\enspace,
\end{align*}
resulting in $g_{\nCenter} = \nSample$ and $g_{\nsqCov} = \frac{1}{2} (\nSample \nSample\transp - \idM)$, recovering equations~\eqref{eq:logderivdelta} and~\eqref{eq:logderivM}.

\subsubsection{Sampling from Radial Distributions}

In order to use this class of distributions for search we need to be
able to draw samples from it. The central idea is to first draw a sample
$\nSample$ from the `standard' density
$\pi(\nSample \,|\, \Mean=0, \sqCov=\idM, \Tail)$, which is then
transformed into the sample $\Sample = \sqCov\transp \nSample + \Mean$,
corresponding to the density $\pi(\Sample \,|\, \sqCov, \Mean, \Tail)$.
In general, sampling $\nSample$ can be decomposed into sampling the
(squared) radius component $r^2 = \|\Sample\|^2$ and a unit vector
$\mathbf{v} \in \R^d$, $\|\mathbf{v}\| = 1$. The squared radius has the
density
\begin{align*}
	\tilde q_{\Tail}(r^2) = \!\!\!\! \int\limits_{\|\Sample\|^2 = r^2} \!\!\!\! Q_{\Tail}(\Sample) \, d\Sample = \frac{2 \pi^{d/2}}{\Gamma(d/2)} \cdot (r^2)^{(d-1)/2} \cdot q_{\Tail}(r^2)
	\enspace,
\end{align*}
where $\Gamma(\cdot)$ denotes the gamma function. In the following we
assume that we have an efficient method of drawing samples from this
one-dimensional density. 
Besides the radius we draw a unit vector $\mathbf{v} \in \R^d$
uniformly at random, for example by normalizing a standard normally
distributed vector. Then
$\nSample = r \cdot \mathbf{v}$ is effectively sampled from
$\pi(\nSample \,|\, \Mean=0, \sqCov=\idM, \Tail)$, and the composition
$\Sample = r \sqCov\transp \mathbf{v} + \Mean$ follows the
density~$\pi(\Sample \,|\, \Mean,\sqCov,  \Tail)$. In many special
cases, however, there are more efficient ways of sampling
$\nSample = r \cdot \mathbf{v}$ directly.

%

\subsubsection{Computing the Fisher Information Matrix}

In section~\ref{sec:gauss-natcoord} the natural gradient coincides with the 
plain gradient, because coordinate system is constructed in such a
way that the Fisher information matrix for the parameters $(\nCenter, \nsqCov)$
is the identity. However, this is in general not possible in the
presence of parameters $\Tail$, typically controlling the radial shape,
and in particular the tail, of the search distribution.

Consider the case $\Tail \in \R^{d'}$. The dimensions of $\nCenter$ and $\nsqCov$
are $d$ and $d(d+1)/2$, respectively, making for a total number of
$m = d(d+3)/2+d'$ parameters. Thus, the Fisher information matrix is an
$(m \times m)$ matrix of the form
\begin{align*}
	\fisher = \bmat \idM & v \\ v^{\top} & c \emat
		\qquad\qquad \text{with }
	v = \frac{\partial^2 \log \pi(\Sample)}{\partial (\nCenter, \nsqCov) \partial \Tail} \in \R^{(m-d') \times d'}
		, \quad
	c = \frac{\partial^2 \log \pi(\Sample)}{\partial \Tail^2} \in \R^{d' \times d'}
		.
\end{align*}
Using the Woodbury identity, we compute the inverse
of the Fisher matrix as
\begin{align*}
	\fisher^{-1} = \bmat \idM & v \\ v^{\top} & c \emat^{-1}
		= \bmat \idM + H v v^{\top} & -H v \\
			-H v^{\top} & H \emat
\end{align*}
with $H = (c - v\transp v)^{-1}$, and exploiting $H\transp = H$.
The natural gradient becomes
\begin{align*}
	\fisher^{-1} \cdot g = \bmat (g_{\nCenter}, g_{\nsqCov}) - H v (v\transp (g_{\nCenter}, g_{\nsqCov}) - g_{\Tail}) \\
					H (v\transp (g_{\nCenter}, g_{\nsqCov}) - g_{\Tail}) \emat
	\enspace,
\end{align*}
which can be computed efficiently in only $\Order(d'^3 + m d')$
operations. 
Assuming that $d'$ does not grow with $d$, this complexity 
corresponds to $\Order(d^2)$ operations in
terms of the search space dimension, compared to $\Order(d^6)$
operations required for a na\"ive inversion of the full Fisher matrix.
In other words, the benefits of the natural coordinate system carry over,
even if we no longer have $\fisher = \idM$.

\subsection{Heavy-tailed NES}
\label{sec:cauchy-nes}

Natural gradient ascent and plain gradient ascent in natural
coordinates provide two equivalent views on the working principle of
NES. In this section we introduce yet another interpretation, with the
goal of extending NES to heavy-tailed distributions, in particular distributions
with infinite variance, like the Cauchy distribution. The problem posed
by these distributions within the NES framework is that they do not
induce a Riemannian structure on the parameter space of the distribution
via their Fisher information, which renders the information geometric
interpretation of natural coordinates and natural gradient ascent
invalid.

Many important types of search distributions have strong \emph{invariance}
properties, e.g.,
multi-variate and heavy-tailed distributions. 
In
this still very general case, the NES principle can be derived solely
based on invariance properties, without ever referring to information
geometric concepts.

The direction of the gradient $\nabla_{\theta} J(\theta)$ depends on the
inner product $\langle \cdot, \cdot \rangle$ in use, corresponding to
the choice of a coordinate system or an (orthonormal) set of basis
vectors. Thus, expressing a gradient ascent algorithm in arbitrary
coordinates results in (to some extent) arbitrary and often sub-optimal
updates. NES resolves this dilemma by relying on the natural gradient,
which corresponds to the distinguished coordinate system (of the tangent
space of the family of search distributions) corresponding to the Fisher
information metric.

The natural coordinates of a multi-variate Gaussian search distribution
turn out to be those local coordinates w.r.t.\ which the
current search distribution has zero mean and unit covariance. This
coincides with the coordinate system in which the invariance properties
of multi-variate Gaussians are most apparent. This connection turns out
to be quite general. In the following we exploit this property
systematically and apply it to distributions with infinite (undefined)
Fisher information.

\subsubsection{Groups of Invariances}

The invariances of a search distribution can be expressed by a group
$\mathcal{G}$ of (affine) linear transformations.
Typically, $\mathcal{G}$ is a sub-group of the group of orthogonal
transformations (i.e., rotations) w.r.t.\ a local coordinate system.
For the above example of a rotationally-symmetric density $Q : \R^d \to \R^+_0$
(e.g., a Gaussian), the densities
\begin{align*}
	\pi(\Sample \,|\, \Mean, \sqCov) = \frac{1}{\det(\sqCov)} \cdot Q \left( \sqCov^{-1} (\Sample - \Mean) \right)
\end{align*}
with $\Mean \in \R^d$ and
$\sqCov \in \R^{d \times d}$, $\det(\sqCov) \not= 0$
form the corresponding multi-variate distribution. Let
$\mathcal{G}_{(\Mean, \sqCov)}$ be the group of invariances of
$\pi(\Sample \,|\, \Mean, \sqCov)$, that is,
$\mathcal{G}_{(\Mean, \sqCov)} = \big\{ g \,\big|\,
	\pi(g(\Sample) \,|\, \Mean, \sqCov) = \pi(\Sample \,|\, \Mean, \sqCov)
	\, \forall \Sample \in \R^d \big\}$. We have
$\mathcal{G}_{(0, \idM)} = \mathbb{O}_{\langle \cdot, \cdot \rangle}(\R^d)
	= \big\{ g \,\big|\, \langle g(\Sample), g(\SampleTwo) \rangle =
	\langle \Sample, \SampleTwo \rangle \,\forall \Sample, \SampleTwo \in \R^d \big\}$,
where the right hand side is the group of orthogonal transformations
w.r.t.\ an inner product, defined as the (affine) linear transformations
that leave the inner product (and thus the properties induced by the
orthonormal coordinate system) invariant.
Here the inner product is the one w.r.t.\ which the density $Q$ is
rotation invariant. For general $(\Mean, \sqCov)$ we have
$\mathcal{G}_{(\Mean, \sqCov)} = h \circ \mathcal{G}_{(0, \idM)} \circ h^{-1}$,
where $h(\Sample) = \sqCov \Sample + \Mean$ is the affine linear
transformation corresponding to the current search distribution.
In general, the group of invariances is only a subgroup of an orthogonal
group, e.g., for a separable distribution~$Q$, $\mathcal{G}$ is the
finite group generated by coordinate permutations and axis flips.

We argue that it is most natural to rely on a gradient or coordinate
system which is \emph{compatible} with the invariance properties of the
search distribution in use. In other words, we should ensure the
compatibility condition
\begin{align*}
	\mathcal{G}_{(\Mean, \sqCov)} \subset \mathbb{O}_{\langle \cdot, \cdot \rangle}(\R^d)
\end{align*}
for the inner product $\langle \cdot, \cdot \rangle$ with respect to
which we compute the gradient $\nabla J$. This condition has a
straight-forward connection to the natural coordinate system introduced
in section~\ref{sec:gauss-natcoord}: It is fulfilled by performing all
updates in local coordinates, in which the current search distribution is
expressed by the density $\pi(\cdot \,|\, 0, \idM) = Q(\cdot)$. In these
coordinates, the distribution is already rotationally symmetric by construction
(or similar for separable distributions), where the rotational symmetry is
defined in terms of the `standard' inner product of the local coordinates.
Local coordinates save us from the cumbersome explicit construction of an
inner product that is left invariant by the
group~$\mathcal{G}_{(\Mean, \sqCov)}$.

Note, however, that $Q(\Sample)$ and $Q(\sigma \cdot \Sample)$ have the
same invariance properties. Thus, the invariance properties make only
the gradient components $\nabla_{\Mean} J$ and  $\nabla_\sqCovB J$
unique, but not the scale component $\nabla_{\sigma} J$. Luckily this
does not affect the (1+1) hill-climber variant of NES, which relies on a
success-based step size adaptation rule (see section~\ref{sec:elitism}). 
Also note that this derivation
of the NES updates works only for families of search distributions with
strong invariance properties, while natural gradient ascent extends to
much more general distributions, such as mixtures of Gaussians.

\subsubsection{Cauchy Distributions}
Given these results, NES, formulated in local coordinates, can
be used with heavy-tailed search distributions without modification. This
applies in particular to the (1+1) hill-climber, which is the most
attractive choice for heavy-tailed search distributions,
because when the search distribution converges to a local optimum and
a better optimum is located by a mutation, then averaging this step over
the offspring population will usually result in a sub-optimal step that
stays within the same basin of attraction. In contrast, a
hill-climber can jump straight into the better basin of attraction, and
can thus make better use the specific advantages of heavy-tailed search
distributions.

\begin{algorithm}[ht]
\SetKwInOut{Input}{input}
\caption{(1+1)-NES with multi-variate Cauchy distribution}
\label{alg:cauchy-xnes}
 \Input{$f$, $\Mean_{init}$, $\Cov_{init}=\sqCov\transp\sqCov$}
 $f_{\text{best}} \leftarrow - \infty$\\
 \Repeat{stopping criterion is met}{
  draw sample 
  $\begin{array}{ll}  
    \nSample & \sim \Normal(0, \idM)\\
    r & \sim \pi_{Cauchy}(0,1
    )\\
    \Sample &\leftarrow r
    \sqCov\transp\nSample+\Mean
\end{array}$ \\
\vspace{0.1cm}
  evaluate the fitness $f(\Sample)$\\
 	$(\nSample_1, \nSample_2) \leftarrow (0, \nSample)$\\
 	$(\Sample_1, \Sample_2) \leftarrow (\Mean, \Sample)$\\
  \vspace{0.05cm}
  \eIf{$f(\Sample) > f_{\text{best}}$}
  {
 	update mean $\Mean \leftarrow \Sample$\\
 	$f_{\text{best}} \leftarrow f(\Sample)$\\
 	$\mathbf{u} \leftarrow (-4, 1)$\\
 	}
 	{$\mathbf{u} \leftarrow (\frac{4}{5}, 0)$\\
 	} 	
  calculate log-derivatives \ \ $\displaystyle\nabla_{\nsqCov} \log \pi\left(\Sample_k |\theta\right)
  \leftarrow \frac{1}{2}\left(\frac{d+1}{r^2 + 1} \nSample_k \nSample_k\transp -\idM\right)$\\  
\vspace{0.1cm}
$\nabla_{\nsqCov} J \leftarrow \frac{1}{2}
\sum_{k=1}^{2} 
\nabla_{\nsqCov} \log\pi\left(\Sample_k | \theta\right) \cdot u_k 
$\\
\vspace{0.1cm}
$ \sqCov \leftarrow \sqCov \cdot\exp\left(\frac{1}{2}\eta_{\sqCov} \nabla_{\nsqCov} J\right)$\\
 }
\end{algorithm}

Of course, the computation of the plain gradient changes depending on
the distribution in use. Once this gradient is computed in the local
coordinate system respecting the invariances of the current search
distribution, it can be used for updating the search parameters $\Mean$
and $\sqCovB$ without further corrections like multiplying with the
(in general undefined) inverse Fisher matrix. For the multi-variate
Cauchy distribution we have
\begin{align*}
	q(\nSample) & = \frac{\Gamma((d+1)/2)}{\pi^{(d+1)/2}} \cdot (\|\nSample\|^2 + 1)^{-(d+1)/2} 
	\enspace,
\end{align*}
which results in the gradient components
\begin{align*}
	\nabla_{\nCenter} J = & \frac{d+1}{\|\nSample\|^2 + 1} \cdot \nSample \\
	\nabla_{\nsqCov} J = & \frac{d+1}{2 \cdot (\|\nSample\|^2 + 1)} \cdot \nSample \nSample\transp - \frac{1}{2} \cdot \idM
	\enspace.
\end{align*}
The full NES hill-climber with multi-variate Cauchy mutations is
provided in algorithm~\ref{alg:cauchy-xnes}.

\section{Experiments}
\label{sec:experiments}

In this section, we empirically validate the new algorithms,
with the goal of answering the following questions:
\begin{itemize}
\item
	How do NES algorithms perform compared to state-of-the-art
	evolution strategies?
\item
	Can we identify specific strengths and limitations of the different
	variants, such as SNES (designed for separable problems) and Cauchy-NES (with heavy-tailed distribution)?
\item
	Going beyond standardized benchmarks,
	should natural evolution strategies be applied to real-world problems?
\end{itemize}
We conduct a broad series of experiments on standard benchmarks, as well
as more specific experiments testing special capabilities. In total, six
different algorithm variants are tested and their behaviors compared
qualitatively as well as quantitatively, w.r.t.\ different modalities.

We start by detailing and justifying the choices of hyperparameters,
then we proceed to evaluate the performance of a number of different variants of NES 
(with and without the importance mixing and adaptation sampling techniques)
on a broad collection of benchmarks.
We further conduct experiments using the separable variant on high-dimensional problems, 
and address the question of global optimization and to what degree a heavy-tailed search distribution
(namely multivariate Cauchy) can alleviate the problem of 
getting stuck in local optima.

\subsection{Experimental Setup and Hyperparameters}

Across all NES variants, we distinguish three hyperparameters: 
the population size $\popsize$,
the learning rates $\eta$
and the utility function $u$ (because we always use fitness shaping, see section~\ref{sec:fs}).
In particular, for the multivariate Gaussian case (xNES)
we have the three learning rates $\eta_{\Mean}$, $\eta_\sigma$, and $\eta_\sqCovB$.

It is highly desirable to have good default settings that scale with the
problem dimension and lead to robust performance on a broad class of
benchmark functions. Table~\ref{tab:settings} provides such default
values as functions of the problem dimension $d$ for xNES.
We borrowed several of the settings from CMA-ES~\citep{hansen:2001}, which seems natural due
to the apparent similarity discussed in section~\ref{sec:cmalike}. Both
the population size $\popsize$ and the learning rate $\eta_{\Mean}$ are the same
as for CMA-ES, even if this learning rate never explicitly appears in
CMA-ES. For the utility function we copied the weighting scheme of
CMA-ES, but we shifted the values such that they sum to zero, which
is the simplest form of implementing a fitness baseline; 
\citet{Jastrebski2006} proposed a similar approach for CMA-ES.
The remaining parameters were determined via an empirical
investigation, aiming for robust performance. 
In addition, in the separable case (SNES) the number of
parameters in the covariance matrix is reduced from
$d(d+1)/2 \in \Order(d^2)$ to $d \in \Order(d)$, 
which allows us to increase the learning rate $\eta_{\covs}$ by a factor of
$d/3 \in \Order(d)$, a choice which has proven robust in practice~\citep{ros:2008}.

\begin{table}%
\centering
\begin{tabular}{l|c}
parameter & default value \\[1mm]
\hline
$\popsize$ & $4 + \lfloor 3\log(d) \rfloor$ \\[1mm]
$\eta_{\Mean}$ & 1 \\[1mm]
$\eta_\sigma = \eta_\sqCovB$ & $\displaystyle  \frac{(9+3\log(d))}{5 d \sqrt{d}}$ \\[1mm]
&\\[1mm]
$\eta_{\covs}$ & $\displaystyle  \frac{(3+\log(d))}{5 \sqrt{d}}$ \\[1mm]
&\\[1mm]
$u_k$ & $\displaystyle \frac{\max\left(0, \log(\frac{\popsize}{2}+1) - \log(i)\right)}{\sum_{j=1}^{\popsize} \max\left(0, \log(\frac{\popsize}{2}+1) - \log(j)\right)} - \frac{1}{\popsize}$ \\[1mm]
\end{tabular}
\caption[Default parameters]{\textbf{}\label{tab:settings}
Default parameter values for xNES and SNES (including the utility function)
as a function of problem dimension~$d$.}
\end{table}

The six algorithm variants that we will be evaluating below are
xNES (algorithm~\ref{alg:xnes}),
its hill-climber variant (1+1)-xNES (see algorithm~\ref{alg:hillclimber-xnes}),
``xNES-im-as'', that is xNES using both importance mixing (section~\ref{sec:im})
and adaptation sampling (section~\ref{sec:as}),
the separable SNES (as in algorithm~\ref{alg:snes}),
its own hill-climber variant (1+1)-SNES (pseudocode not shown),
and finally the heavy-tailed variant (1+1)-NES-Cauchy (as in algorithm~\ref{alg:cauchy-xnes}).
A Python implementation of all these (and more)
is available within the open-source machine learning library PyBrain~\citep{Schaul2010pybrain}.

\subsection{Black-box Optimization Benchmarks}
\label{sec:bbob}

For a practitioner it is important to understand how NES algorithms
compare to other methods on a wide range black-box optimization
scenarios.
Thus, we evaluate our algorithm on all the benchmark functions
of the `Black-Box Optimization Benchmarking' collection (BBOB)
from the 2010 GECCO Workshop for Real-Parameter Optimization.
The collection consists of 24 noise-free functions (12 unimodal, 12 multimodal;~\citealp{bbobnoisefree}) 
and 30 noisy functions~\citep{bbobnoisy}.
In order to make our results fully comparable, we also use the identical 
setup~\citep{bbobsetup}, which transforms the pure benchmark functions
to make the parameters non-separable (for some) and avoid trivial optima at the origin.
To facilitate the comparison 
for the reader without overcrowding the plots, 
we provide the GECCO 2010 results of 
(1,4)-CMA-ES 
alongside our own
(chosen as the most comparable representative among the 13 CMA-ES-based submissions to the workshop).

On all of these benchmarks, we compare xNES (as described in Algorithm~\ref{alg:xnes}) and xNES-im-as, 
that is, the same algorithm but augmented with both importance mixing and adaptation sampling.
On the multi-modal, as well as the noisy benchmarks, we also use the restart strategy (see section~\ref{sec:restarts}).

\begin{figure}
	\centerline{
		\includegraphics[width=0.95\columnwidth, clip=true, trim=1.5cm 1.5cm 2.0cm 1.5cm]{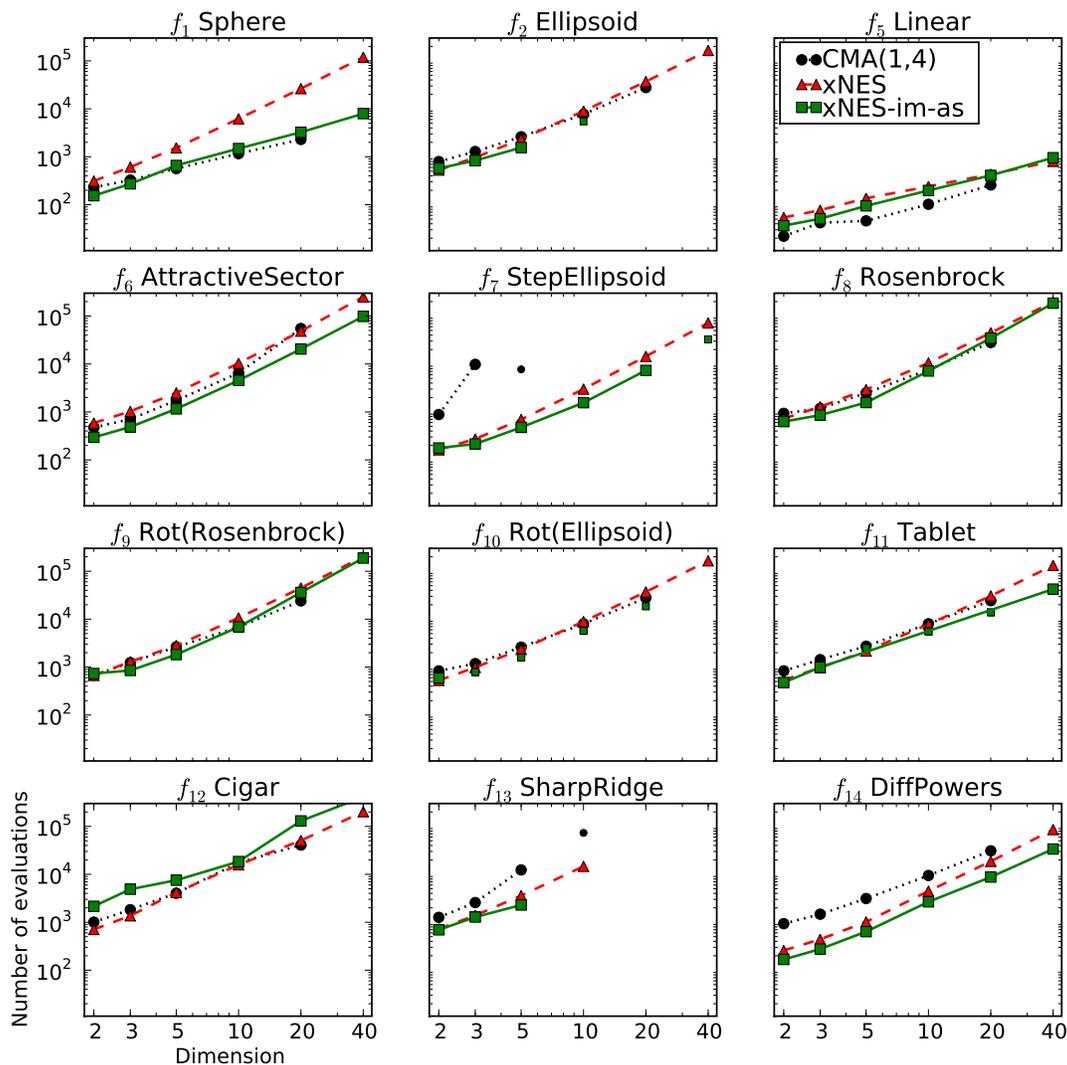}
	}
	\caption[Unimodal benchmarks]{\textbf{}Unimodal benchmarks. 
	Log-log plot of the median number of fitness evaluations
		(over 100 trials) required to reach the
		target fitness value of $-10^{-7}$  for the 12 unimodal 
		benchmark functions on dimensions 2 to 40.
		Disconnected (smaller) plot markers denote cases where the
		corresponding algorithm converged prematurely in at least 50\% of the runs 
		(cases for which 90\% or more prematurely converged are not shown at all). 
		No data was available for (1,4)-CMA-ES on dimension 40.
		Note that xNES (red triangles) consistently solves all benchmarks, with a scaling factor
		that is almost the same over all functions.
		Also, xNES appears to be more stable, especially on the SharpRidge function.
    When employing importance mixing and adaptation sampling (xNES-im-as, green squares),
    performance increases, most substantially on the Sphere function, while
    robustness decreases.
	}
	\label{fig:scale}
\end{figure}

For the unimodal benchmarks (see Figure~\ref{fig:scale}), 
we plot how the performance
of xNES scales with problem dimension (between 2 and 40). 
Shown are the median number of evaluations required to reach a fitness of 
$-10^{-7}$. We find that xNES is on par with CMA-ES for most benchmarks,
but generally more robust (e.g., on the StepEllipsoid benchmark). 
Employing importance mixing and adaptation sampling further increases performance
(most significantly on the simple benchmarks like the Sphere function),
but at the cost of robustness.

\begin{figure}
	\centering
	\includegraphics[width=1.05\columnwidth, clip=true, trim=0cm 1.3cm 0cm 0.6cm]{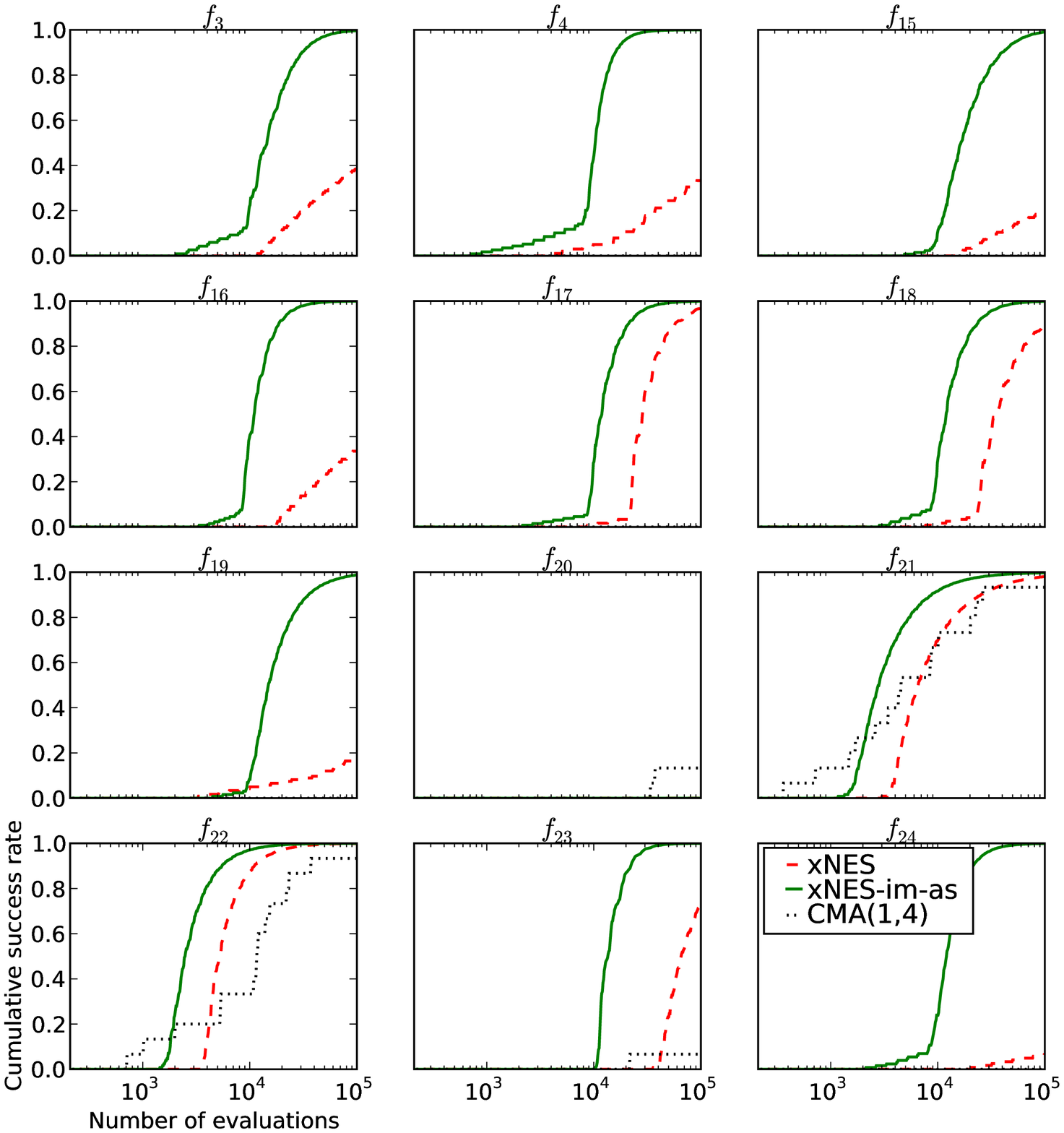}		
	\caption[Multi-modal benchmarks]{\textbf{}Multi-modal benchmarks, comparing the performance of xNES 
	with restart strategies (dashed red)
	and (1,4)-CMA-ES (dotted black).
	Shown is the empirical cumulative success rate (over 100 runs, 15 for CMA-ES). 
	Note that xNES clearly outperforms CMA-ES on all functions except $f_{21}$ 
	and $f_{22}$ (both of which are simply combinations of 101 and 21 Gaussian peaks, respectively, 
	without any global structure), 
	where the results are very similar. 
	When additionally employing importance mixing and adaptation sampling (xNES-im-as, solid green),
  performance improves even further.
	}
	\label{fig:multimodal}
\end{figure}

\begin{figure}
	\centering
	\includegraphics[width=1.05\columnwidth, clip=true, trim=0cm 1.3cm 0cm 0.6cm]{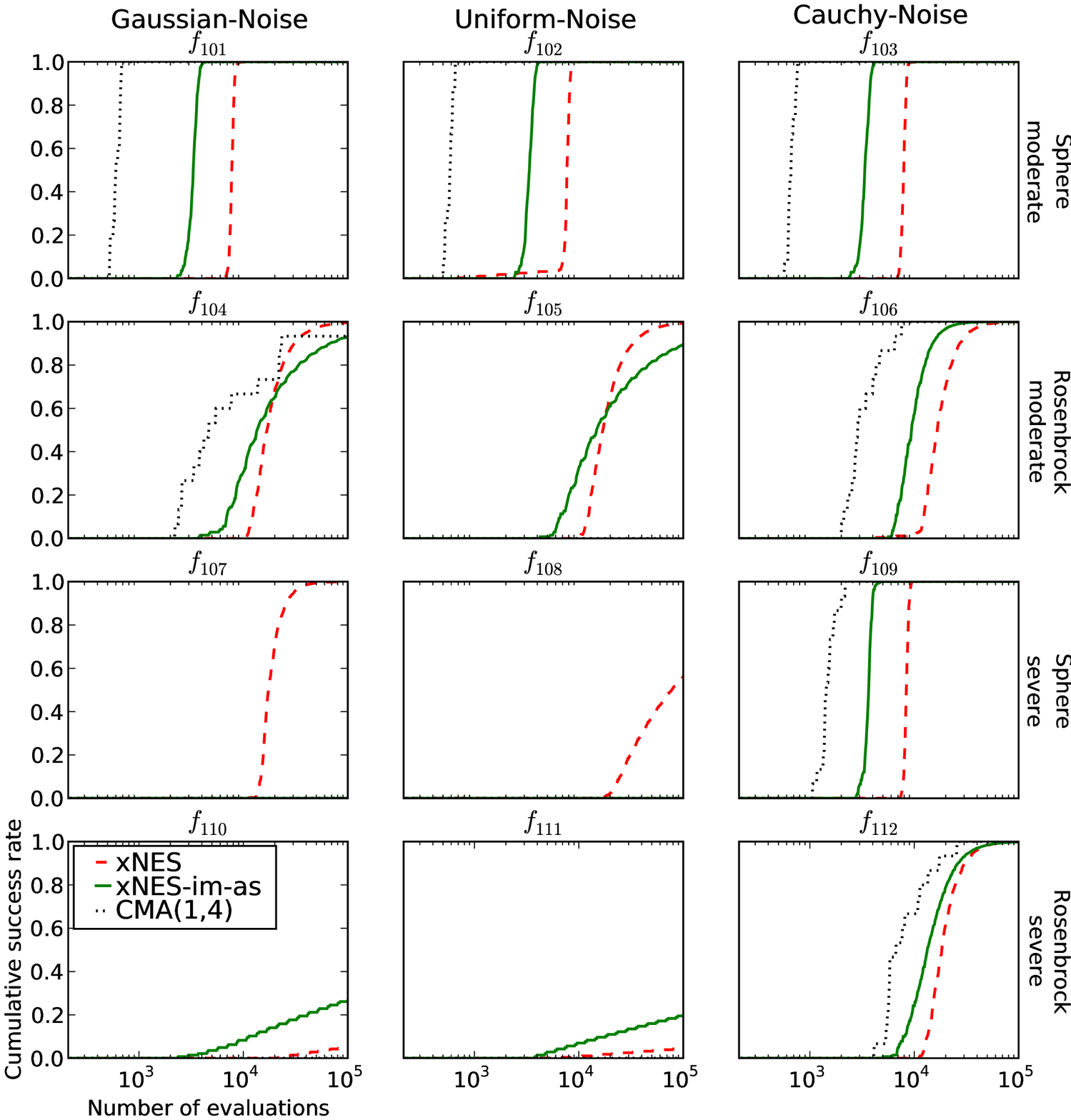}		
	\caption[Noisy benchmarks]{\textbf{}Noisy benchmarks, comparing the performance of xNES 
	with restart strategies (dashed red), 
	and (1,4)-CMA-ES (dotted black). 
	Shown is the empirical cumulative success rate (over 100 runs, 15 for CMA-ES). 
	The benchmarks are grouped by type of noise (vertical) and underlying function (horizontal).
	Generally speaking, Cauchy-noise is	the least harmful, as the rare outliers do not affect
	the ranks all that much.
	We find that xNES outperforms CMA-ES on the majority of the functions. 
	Additionally employing importance mixing and adaptation sampling (xNES-im-as, solid green),
  however, improves performance only on some of the functions.
	Continued in Figure~\ref{fig:noisy2}.
	}
	\label{fig:noisy1}
\end{figure}

\begin{figure}
	\centering
	\includegraphics[width=0.95\columnwidth, clip=true, trim=0cm 1.3cm 0cm 0.6cm]{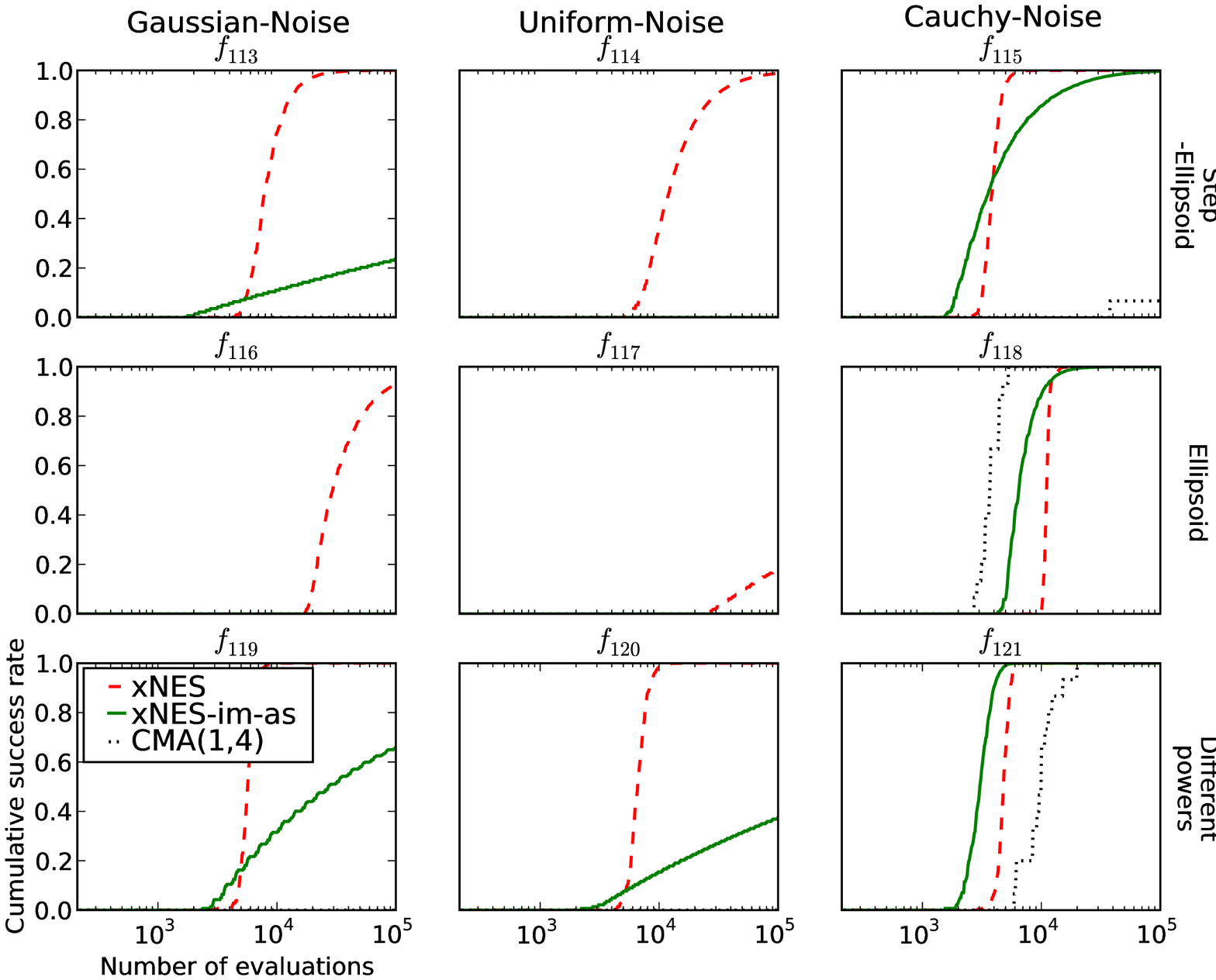}		
	\includegraphics[width=0.95\columnwidth, clip=true, trim=0cm 0.6cm 0cm 1.2cm]{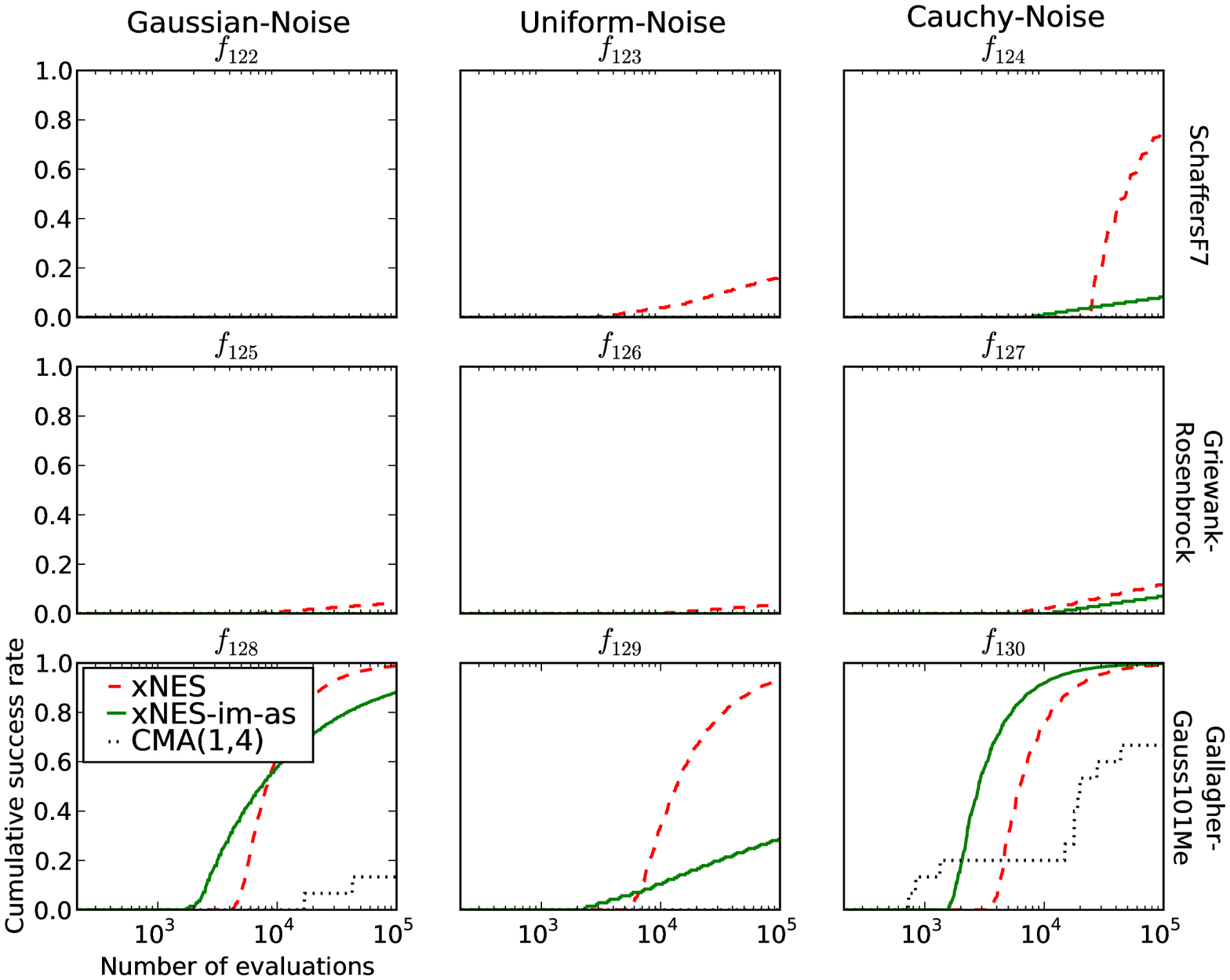}
	\caption[Noisy benchmarks (continued)]{\textbf{}Noisy benchmarks (continued).}
	\label{fig:noisy2}
\end{figure}

For the multi-modal benchmarks (see Figure~\ref{fig:multimodal}), as well as the 
noisy benchmarks (see Figures~\ref{fig:noisy1} and~\ref{fig:noisy2}),
where not all runs succeed, we instead show the cumulative success rates 
(again, for reaching a fitness of $-10^{-7}$),
for problem dimension $d=5$. 
We find that xNES together with our restart strategies 
lead to very good performance, clearly outperforming CMA-ES on 
almost all multi-modal functions, and many noisy functions.
For xNES-im-as, the results correspond to the expected trade-off
that these techniques improve performance, with a substantial boost on the 
noise-free multi-modal functions, but reduce robustness, 
as is evident from the results on many of the harder noisy benchmarks
(only attenuated in part by the restart strategy).

\subsection{Separable NES}

The SNES algorithm is expected to perform at least as well as xNES on
separable problems, while it should show considerably worse performance
 in the presence of highly dependent variables. We first present a
number of experiments on standard benchmarks with the aim of understanding
its behavior and in particular its limitations in more detail.
Furthermore, the algorithm is specifically designed to scale gracefully
to high-dimensional search problems. We thus apply SNES to two tasks with a
scalable and considerably high search space dimension, namely
neuro-evolutionary controller design, and the problem of finding low
energy states in Lennard-Jones potentials.

\subsubsection{Separable and Non-separable Benchmarks}
\begin{figure}
	\centerline{
		\includegraphics[width=1.1\textwidth, clip=true, trim=3cm 0cm 3cm 0cm]{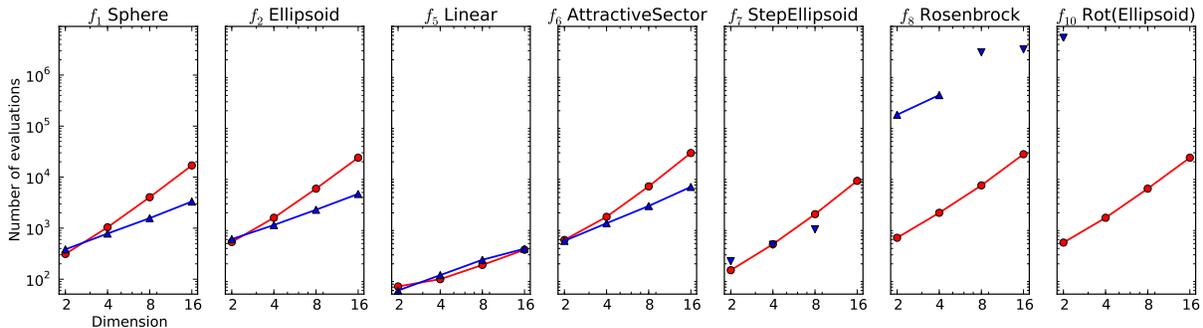}
	}
	\caption[SNES vs. xNES]{\textbf{}Comparison of the performance of xNES (red circles) 
	and SNES (blue triangles) on a subset of
	the unimodal BBOB benchmark functions. The log-log plots show the median number of evaluations 
	required to reach the target fitness $10^{-7}$, for problem dimensions
	ranging from 2 to 16 (over 20 runs). 
	The first 4 benchmark functions are separable, 
	the other three are not.	
	The inverted triangles indicate cases where SNES converged 
	to the optimum in less than 90\% of the runs.
	}
	\label{fig:bbob}
\end{figure}

First, we evaluate SNES on a subset of
the unimodal benchmark problems
from the BBOB framework~\citep{bbobnoisefree,bbobsetup}. These benchmarks
test the capability of SNES to descend quickly into local optima, a key
property of most evolution strategies. The results in
figure~\ref{fig:bbob} show how SNES dominates when the
function is separable ($f_1$ through $f_6$), and
converges much slower than xNES in non-separable benchmarks, as expected.
In particular, on the rotated ellipsoid function $f_{10}$, which is designed
to make separable methods fail, SNES requires 4 orders of magnitude
more evaluations. In dimensions $d > 2$ it fails completely because the
resolution of double precision numbers is insufficient for this task.

\subsubsection{Neuro-evolution}
\begin{figure}[ht]
	\centerline{
		\includegraphics[width=0.5\columnwidth, clip=true, trim=0.6cm 0.1cm 0cm 0cm]{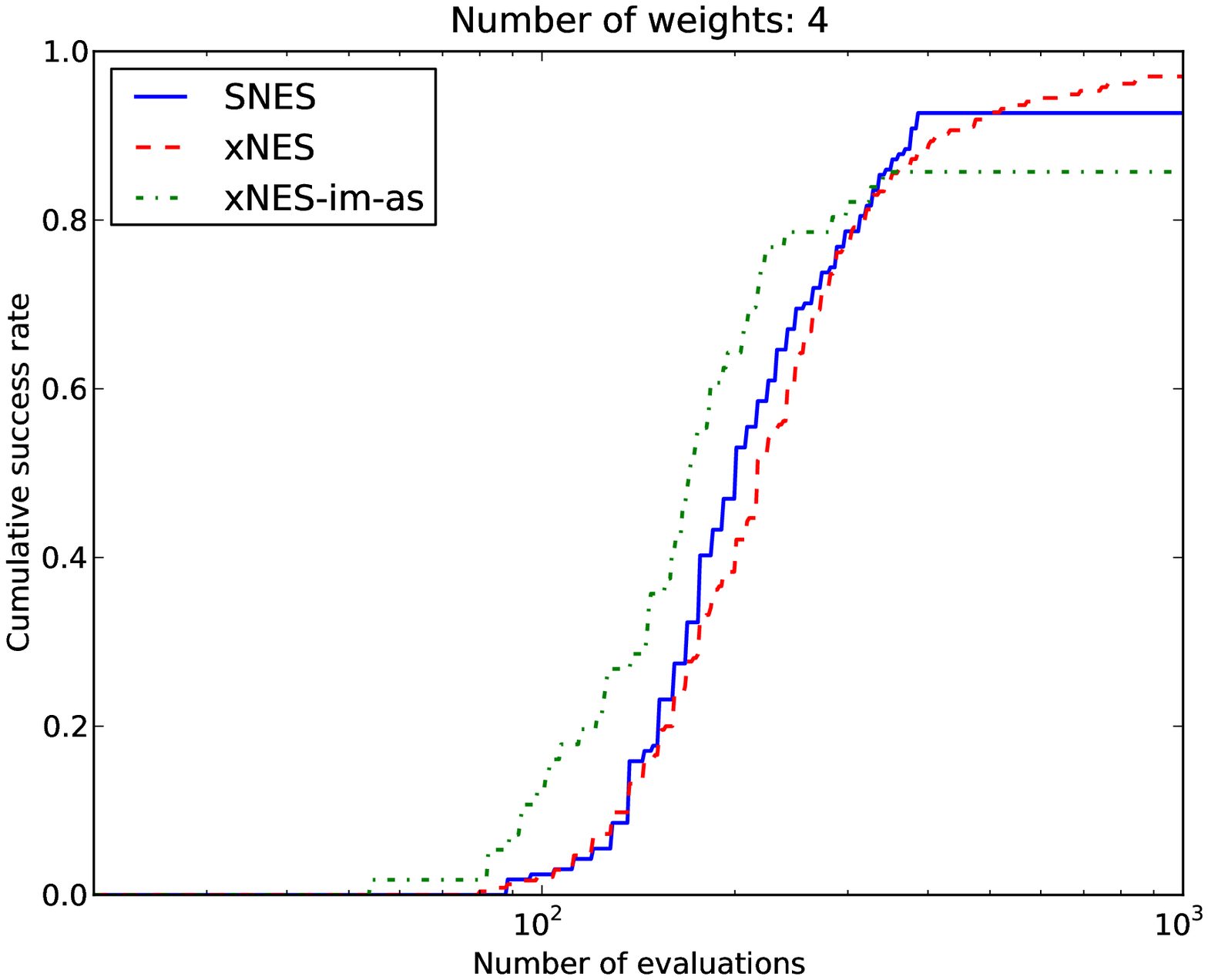}
		\includegraphics[width=0.5\columnwidth, clip=true, trim=0.6cm 0.1cm 0cm 0cm]{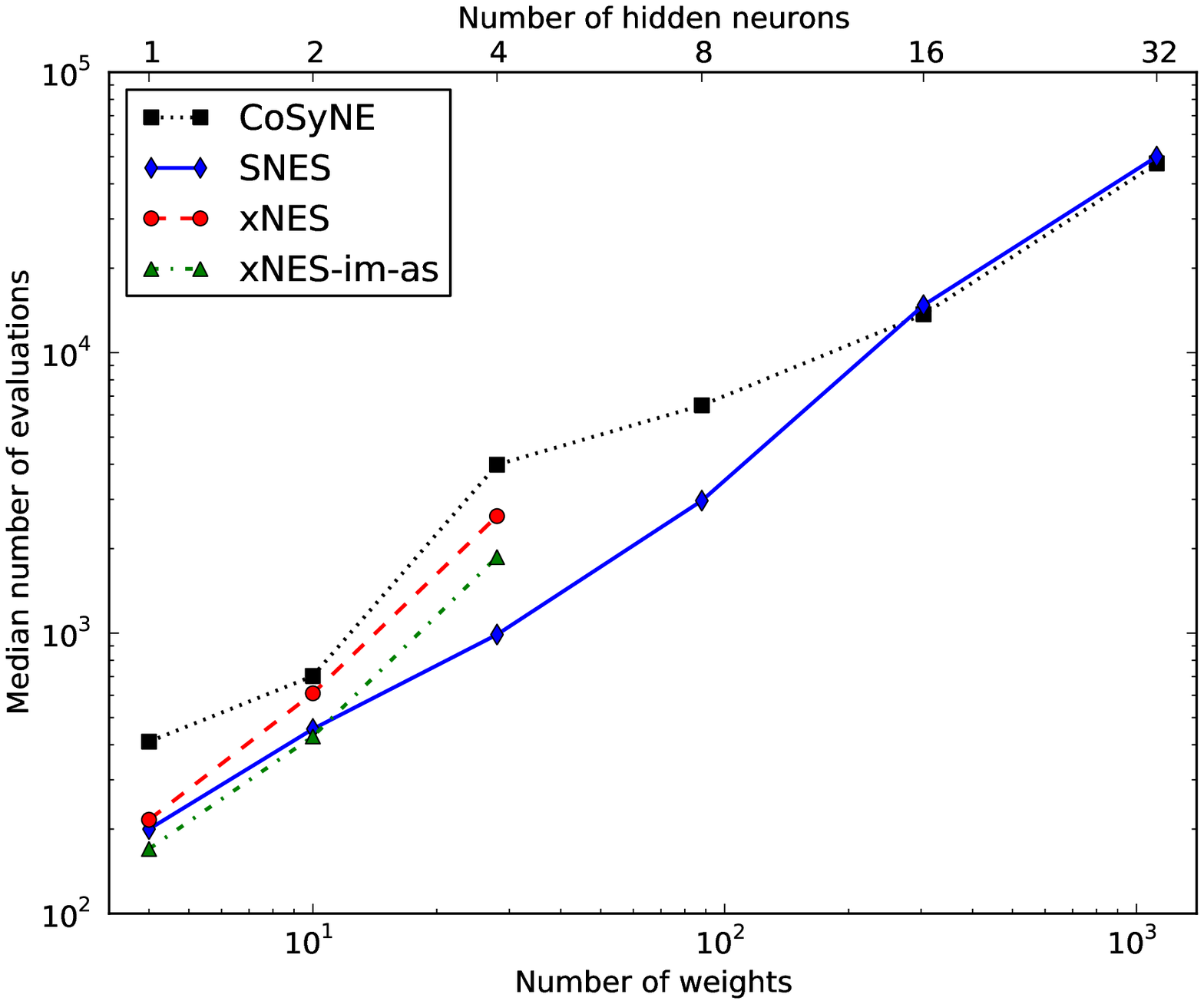}
		}
	\caption[Non-Markovian double-pole balancing]{\textbf{Left:} Plotted are the cumulative success rates on 
	the non-Markovian double-pole balancing task
	after a certain number of evaluations,
	empirically determined over 100 runs for each algorithm, using a single $\tanh$-unit ($n=1$) 
	(i.e., optimizing 4~weights).
	We find that all three algorithm variants give state-of-the-art results, with a slightly faster but less robust performance for xNES with importance mixing and adaptation sampling.
	\textbf{Right:} Median number of evaluations required to solve the same task, but with increasing number of neurons (and corresponding number of weights). We limited the runtime to one hour per run, which explains why no results are available for xNES on higher dimensions (cubic time complexity). The fact that SNES quickly outperforms xNES, also in number of function evaluations,
	indicates that the benchmark is (sufficiently close to) separable, and it is unnecessary to use the full covariance matrix.
	For reference we also plot the corresponding results of the previously best performing algorithm CoSyNE~\citep{Gomez:08jmlr}.
	}
	\label{fig:2poles}
\end{figure}

In the second experiment, we show how SNES is well-suited for
neuroevolution problems because they tend to be high-dimensional,
multi-modal, but with highly redundant global optima (there is
not a unique set of weights that defines the optimal behavior).
In particular, we run it on Non-Markovian double pole balancing,
a task which involves
balancing two differently sized poles hinged on a cart that moves on
a finite track. The single control consists of the force $F$ applied to the
cart, and observations include the cart's position and the poles' angles, but
no velocity information, which makes this task partially observable. 
It provides a perfect testbed for algorithms focusing on learning fine control
with memory in continuous state and action spaces~\citep{wieland91pole}. 
The controller is represented by a simple recurrent neural network, with three inputs,
(position $x$ and the two poles' angles $\beta _{1}$ and $\beta _{2}$),
and a variable number $n$ of $\tanh$ units in the output layer,
which are fully connected (recurrently), resulting in a total of $n(n+3)$
weights to be optimized.
The activation of the first of these recurrent neurons directly determines the force to be applied.
We use the implementation found in PyBrain \citep{Schaul2010pybrain}.

An evaluation is considered a success if the poles do not fall over
for $100,000$ time steps. 
We experimented with recurrent layers of sizes $n=1$ to $n=32$ 
(corresponding to between 4 and 1120 weights). 
It turns out that a single recurrent neuron is sufficient to solve the task (Figure~\ref{fig:2poles}, left).
In fact,  both the xNES and SNES results are state-of-the-art, 
outperforming the previously best algorithm  
(CoSyNE;~\citealp{Gomez:08jmlr}, with a median of 410 evaluations) by a factor two.

In practical scenarios however, we cannot know the best network size a priori,
and thus the prudent choice consists in overestimating the required size.
An algorithm that graciously scales with problem dimension is therefore highly desirable,
and we find (Figure~\ref{fig:2poles}, right) that SNES is exhibiting precisely that behavior.
The fact that SNES outperforms xNES with increasing dimension, 
also in number of function evaluations, 
indicates that the benchmark is separable, 
and it is unnecessary to use the full covariance matrix.
We conjecture that this a property shared with the majority
of neuroevolution problems that have enough weights to exhibit 
redundant global optima (some of which can be found without 
considering all parameter covariances).

\subsubsection{Lennard-Jones Potentials}

\begin{figure}[ht]
	\centerline{
		\includegraphics[width=0.8\columnwidth, clip=true, trim=1cm 0.8cm 0cm 0cm]{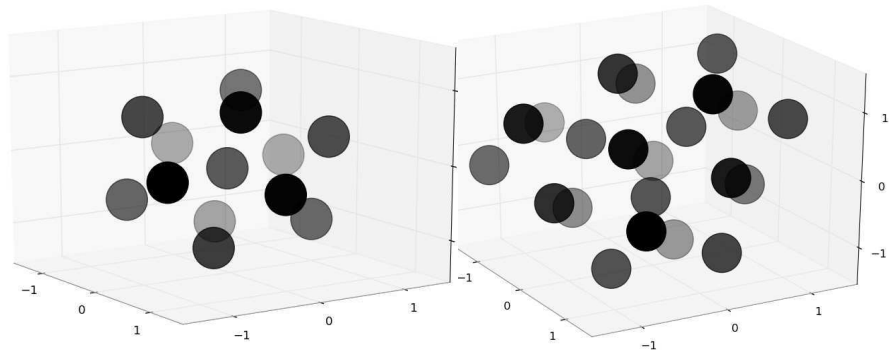}
	}
	\caption[Lennard-Jones clusters]{\textbf{}Illustration of the best configuration found for 13 atoms
	(symmetric, left), and 22 atoms (asymmetric, right).}
	\label{fig:13atoms}
\end{figure}

\begin{figure}[ht]
  \centerline{
    \includegraphics[width=1.1\textwidth, clip=true, trim=0cm 0cm 0cm 0cm]{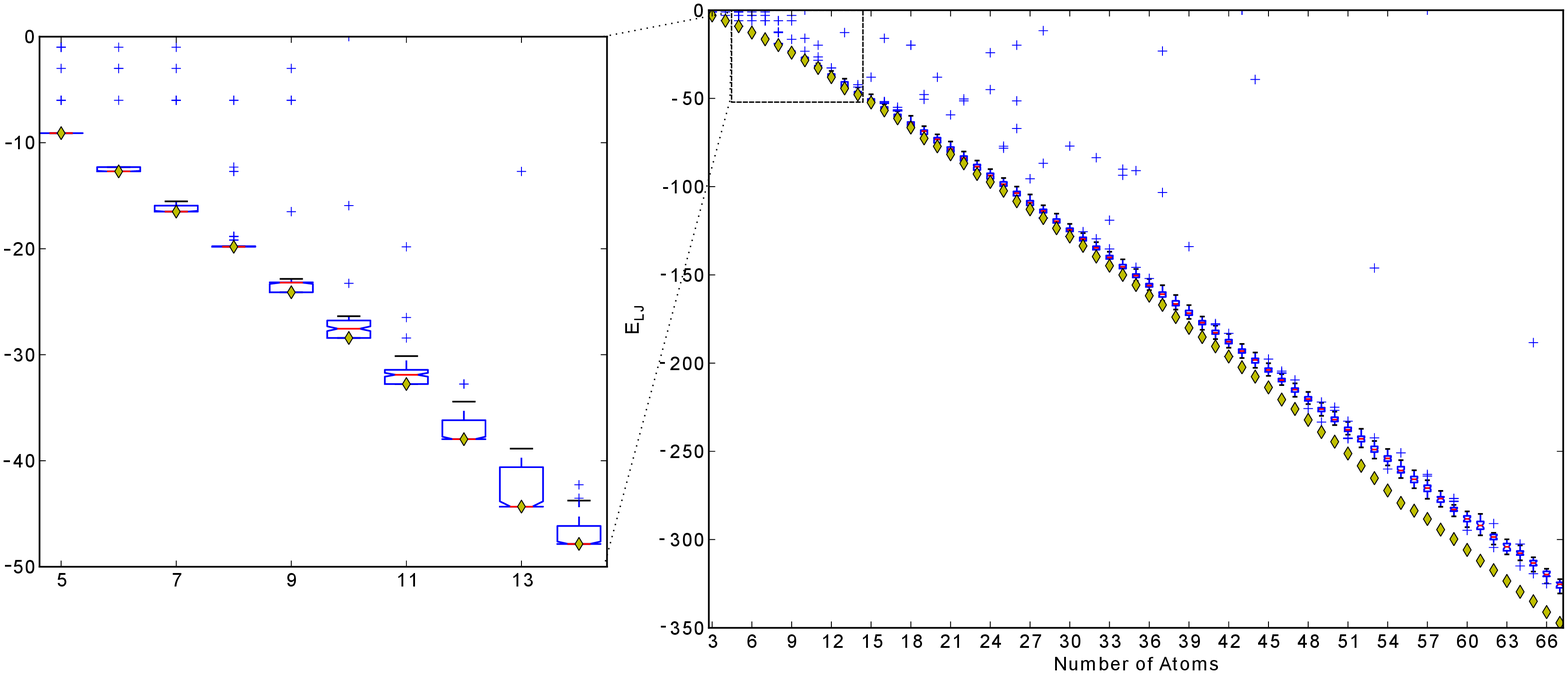}
  }
  \caption[Lennard-Jones benchmark results]{\textbf{}Performance of (1+1)-SNES 
  on the Lennard-Jones benchmark
	for atom clusters ranging from 3 to 67 atoms (corresponding to 
	problem dimensions $d$ of 9 to 201). 
	The yellow diamonds indicate the best known configurations (taken
	from~\citep{Wales1998}), and the box-plots show upper and lower quartile
	performance (the red line being the median) of SNES, over 100 runs. The
	inset is a zoom on the behavior in small dimensions, where SNES succeeds in locating the true optimum in a large fraction of the runs.
	}
	\label{fig:lj-plot}
\end{figure}

In our third benchmark, we show the performance of SNES 
on the widely studied problem of minimizing
the Lennard-Jones atom cluster potentials, which is known for being
extremely multi-modal~\citep{Wales1998}.
For that reason we employ the separable hill-climber variant (1+1)-SNES.
The objective consists in finding that configuration of $N$ atoms which minimizes
the potential energy function
\[
	E_{LJ} \propto \sum_{i,j\leq N} \left[ \left(\frac{1}{r_{ij}} \right)^{12} 
	- \left(\frac{1}{r_{ij}} \right)^6\right]
	\enspace,
\]
where $r_{ij}$ is the distance between atoms $i$ and $j$ 
(see also figure~\ref{fig:13atoms} for an illustration).
For the setup here, we initialized $\Mean$ near 0 and the step-sizes at
$\covs_i=0.01$ to avoid jumping into a local optimum in the fist generation.
The results are plotted in figure~\ref{fig:lj-plot}, showing how SNES
scales convincingly to hundreds of parameters (each run up to $500d$ function evaluations).

\subsection{Heavy Tails and Global Optimization}
Can NES algorithms with heavy-tailed search distributions enhance the capability of
escaping local optima? 
Our tests with the extremely heavy-tailed Cauchy
distribution investigate the handling of multi-modality.

\begin{figure}[ht]
	\centerline{
		\includegraphics[width=0.55\columnwidth, clip=true, trim=0cm 0.8cm 0cm 0cm]{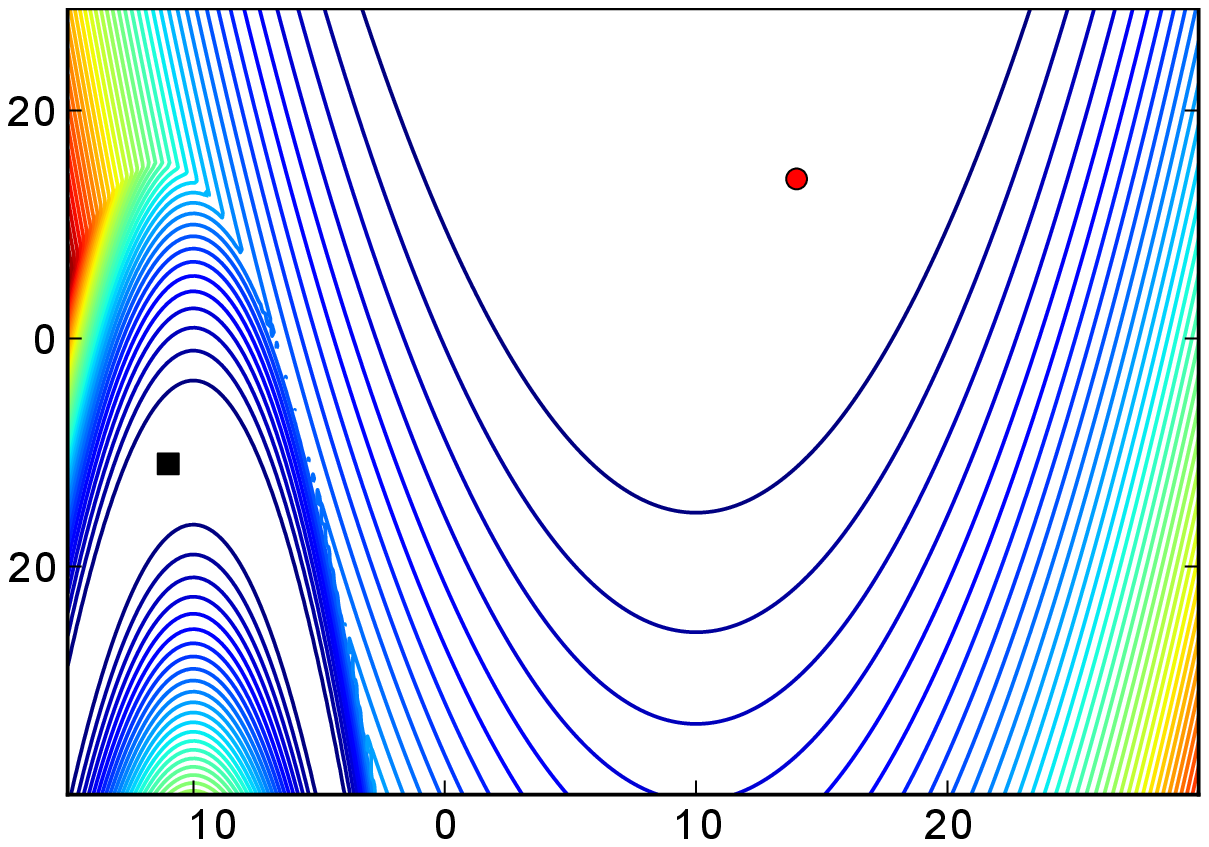}
		\includegraphics[width=0.55\columnwidth, clip=true, trim=0cm 0.1cm 0cm 0cm]{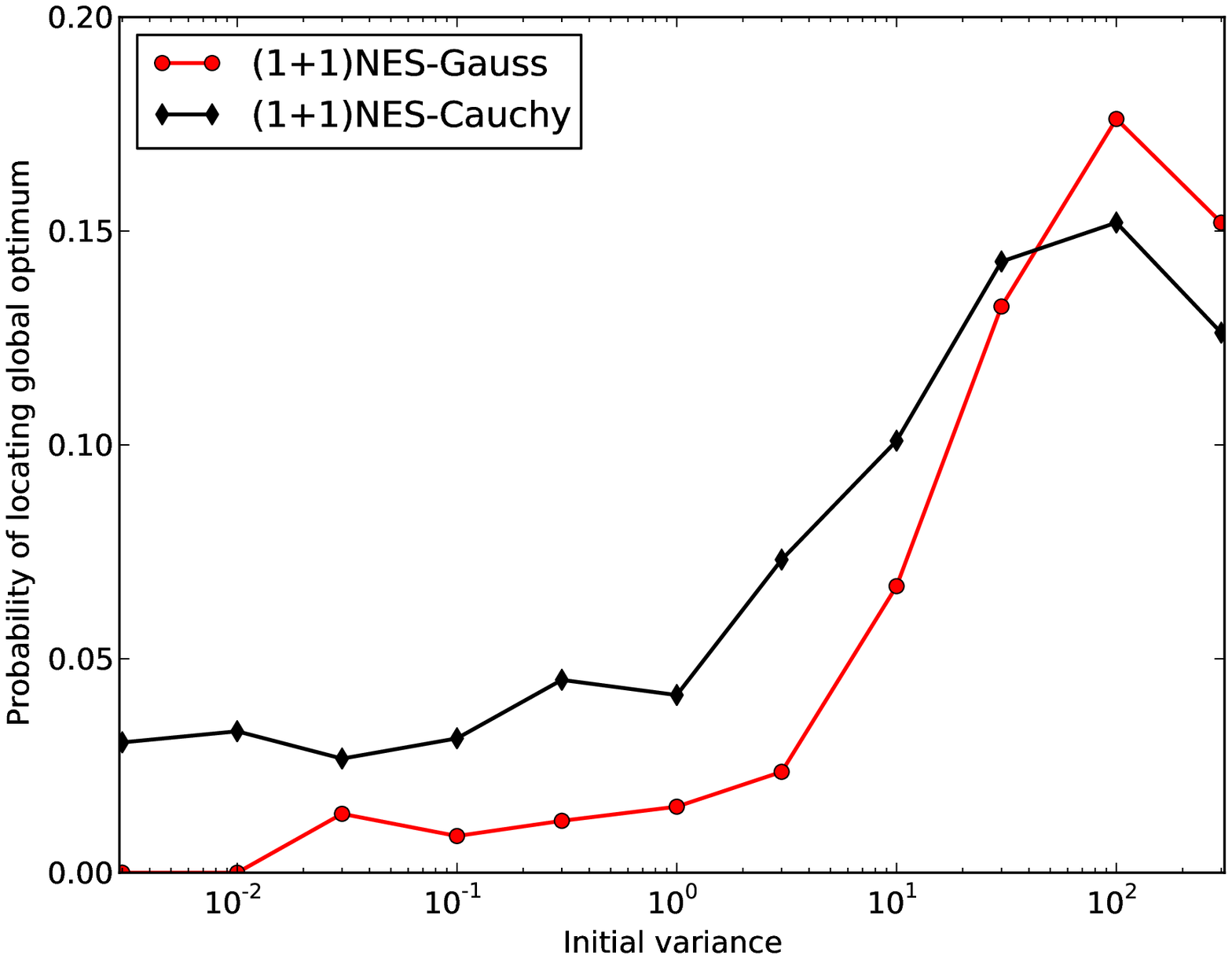}
		}
	\caption[Double-funnel benchmark]{\textbf{Left:} 
	Contour plot of the 2-dimensional Double-Rosenbrock function $f_{2Rosen}$,
	illustrating its deceptive double-funnel structure. The global structure leads the search
	to the local optimum ((14,14), red circle), whereas the true optimum ((-11,-11), black square) is located
	in the smaller valley.
  \textbf{Right:} Empirical success probabilities (of locating the global optimum),
	evaluated over 200 runs, of 1000 function evaluations each,
	on the same benchmark, 
	while varying the size of the initial search distribution.
	The results clearly show the robustness of using a heavy-tailed distribution.
	}
	\label{fig:double-funnel-plot}
\end{figure}

The first benchmark function is 
\[
f_{2Rosen}(\Sample) = \min\left\{ f_8(-\Sample-10), 5+f_8\left(\frac{\Sample-10}{4}\right) \right\}
\enspace,
\]
where
$f_8$ is the well-known Rosenbrock function~\citep{bbobnoisefree},
and the transformation $(\Sample - 10) / 4$ is component-wise.
Our variant has a deceptive \emph{double-funnel} structure, with a large valley
containing a local optimum and a smaller but deeper valley containing
the global optimum. The global structure will tend to guide the search
towards the local optimum
(see also figure~\ref{fig:double-funnel-plot}, left, for an illustration). 
For this experiment, the search distribution is
initialized at mid-distance between the two optima, and the initial
step-size $\sigma$ is varied.
Figure~\ref{fig:double-funnel-plot}(right) shows the proportion of runs
that converge to the global optimum, instead of the (easier to locate)
local one, comparing  for a multivariate Cauchy and Gaussian (1+1)-NES.

The second experiment uses the following `random-basin' benchmark function:
\begin{align*}
	f_{rb}(\Sample) = 1 - & \frac{9}{10} r\left( \left\lfloor \frac{\Sample_1}{10} \right\rfloor, \dots, \left\lfloor \frac{\Sample_d}{10} \right\rfloor \right)\\
	- & \frac{1}{10} r(\lfloor \Sample_1 \rfloor, \dots, \lfloor \Sample_d \rfloor) \cdot \prod_{i=1}^d \sin^2(\pi \Sample_i)^{\frac{1}{20d}}
	\enspace
\end{align*}
to investigate the degree to which a heavy-tail distribution
can be useful when the objective function
is highly multi-modal, but there is no global structure to exploit.
Here $r : \Z^d \to [0, 1]$ is a pseudo-random number generator,
which approximates an i.i.d.\ uniformly random distribution for each tuple of
integers, while still being deterministic, i.e., each tuple
evaluates to the same value each time.
In practice, we implement it as a Mersenne twister~\citep{Matsumoto1998}, 
seeded with the hash-value of the integers. 
Further, to avoid axis-alignment, we rotate the function
by multiplying with an orthonormal random $d\times d$ matrix.

One interesting property of this function is that each unit-sized
hypercube is an ``attractor'' of a local optimum. Thus, while sampling
points from one hypercube, an ES will contract its search distribution,
making it harder to escape from that local optimum. Furthermore, the values
of the local optima are uniformly distributed in $[0,1]$, and do
not provide a systematic global trend (in contrast to the Rastrigin
function).
If the optimization results in a value of, say, $0.11$, then we
know that only $11\%$ of the local optima are better than this.

\begin{figure}[ht]
	\centerline{
		\includegraphics[width=0.55\columnwidth]{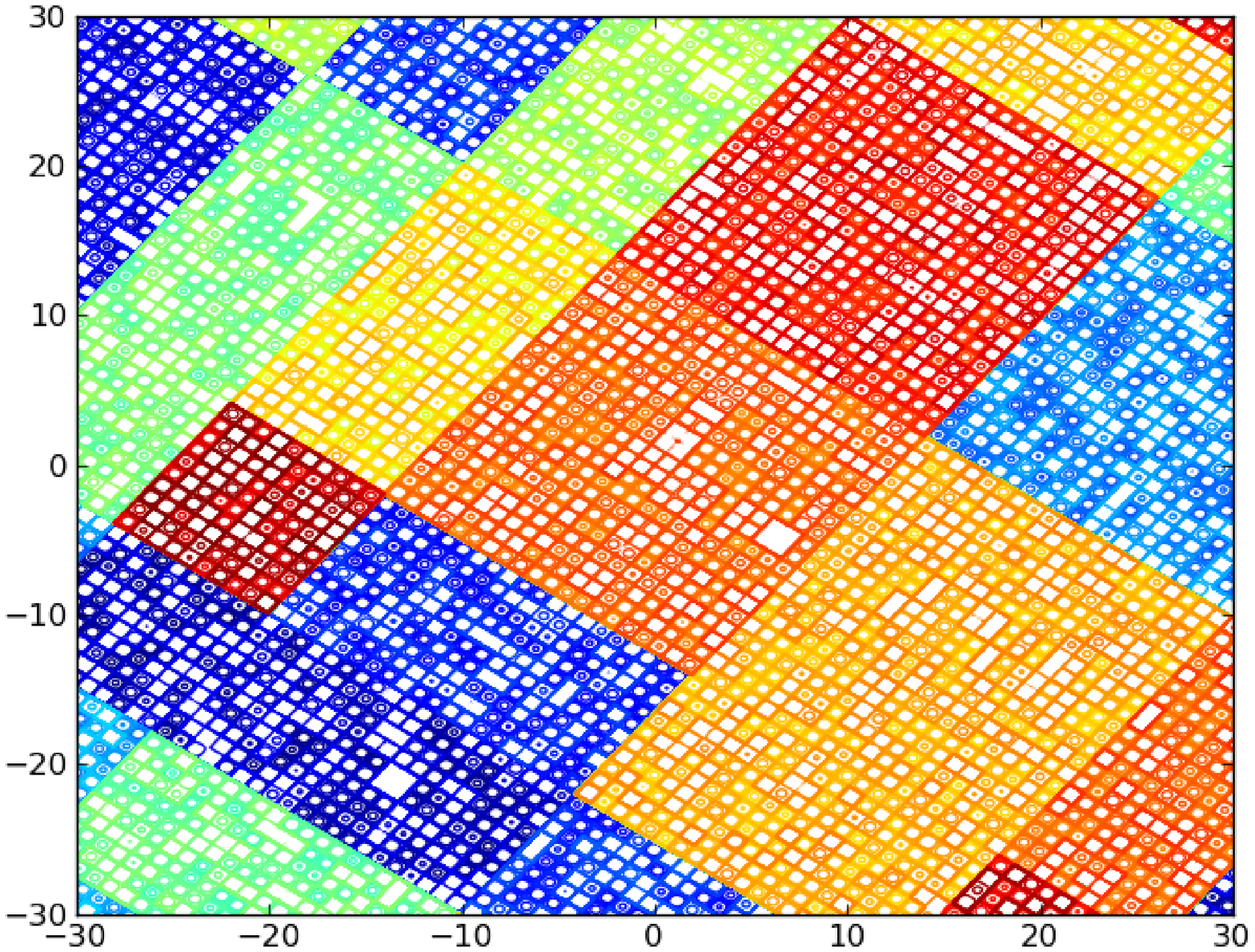}
		\includegraphics[width=0.55\columnwidth, clip=true, trim=0cm 0.1cm 0cm 0cm]{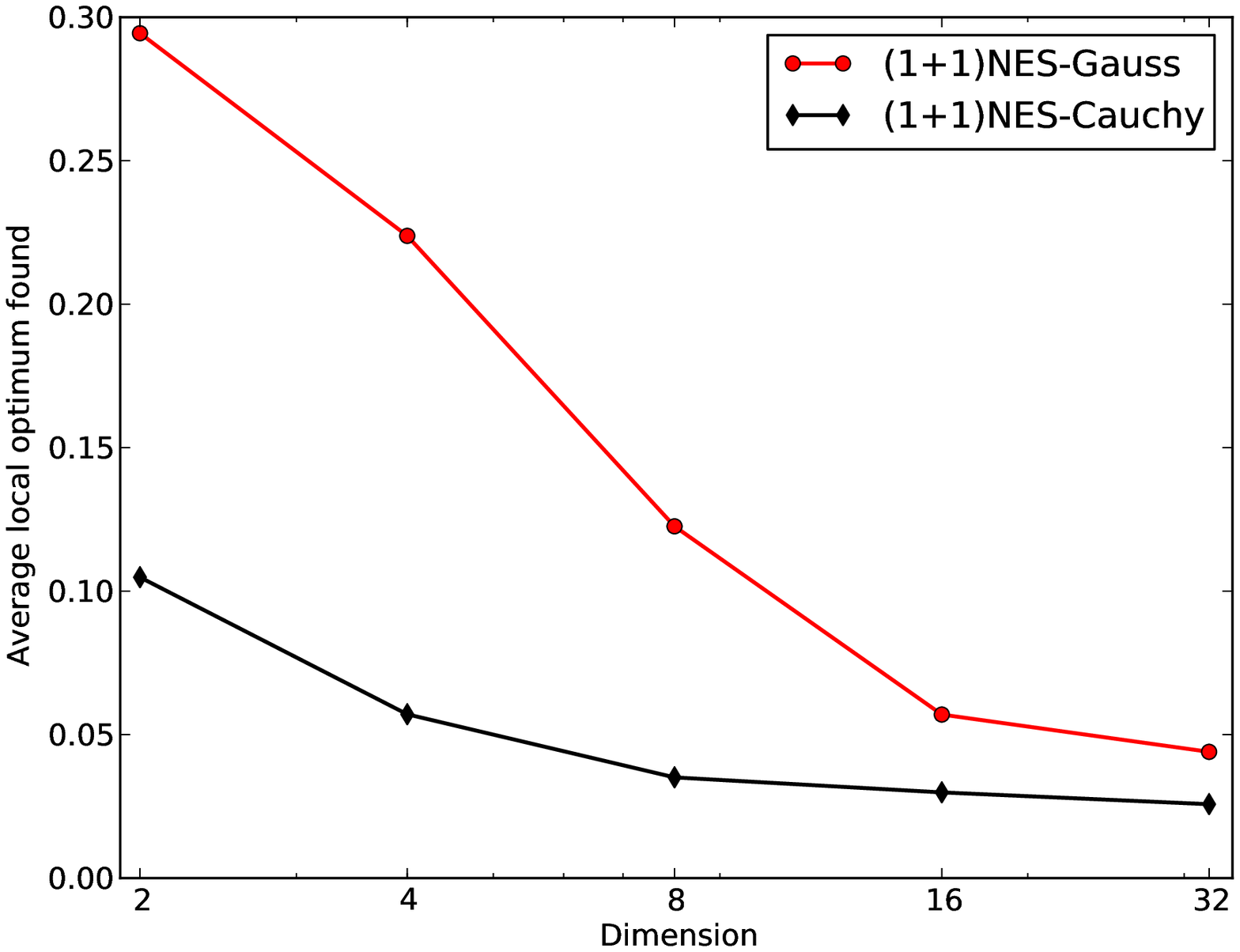}
		}
	\caption[Deceptive global optima]{\textbf{Left:} Contour plot of (one instantiation of) the deceptive global
	optimization benchmark function $f_{rb}$, in two dimensions. It is
	constructed to contain local optima in unit-cube-sized spaces, whose
	best value is uniformly random. In addition, it is superimposed on
	$10^d$-sized regional plateaus, also with uniformly random value.
	\textbf{Right:} Value of the local optimum discovered on $f_{rb}$
	(averaged over 250 runs, each with a budget of $100d$ function evaluations) 
	as a function of problem dimension.
	Since the locally optimal values are uniformly distributed in
	$[0,1]$, the results can equivalently be interpreted as the top
	percentile in which the found local optimum is located. E.g., on
	the 4-dimensional benchmark, NES with the Cauchy distribution
	tends to find one of the 6\% best local optima, whereas employing
	the Gaussian distribution only leads to one of the best 22\%.
	}
	\label{fig:non-global-plot}
\end{figure}

Figure~\ref{fig:non-global-plot} shows the results: 
not surprisingly, employing the Cauchy distribution for search
permits longer jumps, and thus enables the algorithm to find better
local optima on average. The Cauchy version outperforms the Gaussian
version by a factor of two to three, depending on the problem dimension.
Note that the improvement (for both distributions) is due 
the number of neighbor-cubes increasing exponentially with dimension, 
thus increasing the chance that a relatively small
jump will reach a better local optimum.
At the same time, the adaptation of the step size is slowed down by the
dimension-dependency of the learning rate, which leaves the algorithm
more time to explore before it eventually converges into one of the local optima.

\subsection{Results Summary}

Our results have a number of implications. First of all, the results on
the BBOB benchmarks show that NES algorithms are 
competitive with the state-of-the-art across a wide variety of black-box
optimization problems.

Beyond this very general statement, we have demonstrated advantages and
limitations of specific variants, and as such established the generality and
flexibility of the NES framework.
Experiments with heavy-tailed and separable distributions demonstrate
the viability of the approach on high-dimensional and complex,
deceptively multi-modal domains. We obtained best reported results on
the difficult task of training a neural controller for double
pole-balancing. This test, together with good results on the
Lennard-Jones problem, show the feasibility of the algorithm for
real-world search and optimization problems.

The multi-start strategy, although simple and non-adaptive in spirit,
brings considerable improvements on many difficult multi-modal
benchmarks. Its technique of interleaving multiple runs has advantages
over truly sequential restart strategies, waiting for convergence
before restarting.

In summary, our results demonstrate that it
is indeed possible for an algorithm with a clean derivation from first
principles to achieve state-of-the-art to best performance in the
heuristics-dominated field of black-box optimization.

\section{Discussion}
\label{sec:discussion}

\begin{table}[ht]
	\centering
		\begin{tabular}{|c|c|c|c|}
		\hline
		Technique & Issue addressed & Applicability & Relevant \\
		          &                 & limited to    & section \\
		\hline
		Natural gradient & Scale-invariance, many more & - & \ref{sec:ng}\\
		Fitness shaping  & Robustness & - & \ref{sec:fs}\\
		Importance mixing & Performance, parameter sensitivity & - & \ref{sec:im}\\
		Adaptation sampling &Performance, parameter sensitivity &  - & \ref{sec:as}\\
		Restart strategies & Reduced sensitivity to local optima & - & \ref{sec:restarts}\\
		Exponential parameterization & Covariance constraints & Radial & \ref{sec:expmap}\\
		Natural coordinate system &  Efficiency & Radial & \ref{sec:xnes}\\
		\hline
		\end{tabular}
	\caption[Summary of enhancing techniques]{Summary of enhancing techniques}
	\label{tab:summary}
\end{table}

Table \ref{tab:summary} summarizes the various techniques we introduced. 
The plain search gradient suffers from premature convergence and lack of scale invariance 
(see section~\ref{sec:vanillalimitations}). 
Therefore, we use the natural gradient instead, which turns NES into a viable optimization method.
To improve performance and robustness, we introduced several novel techniques.
Fitness shaping makes the NES algorithm invariant to
order-preserving transformations of the fitness function, thus increasing robustness.
Importance mixing reduces the number of necessary samples to estimate the search gradient, while
adaptation sampling adjusts NES's learning rates online. 
Empirically we found that
using both importance mixing and adaptation sampling yields highly performant results 
on the standard benchmarks.
In addition, restart strategies substantially improve success probabilities on multi-modal functions and noisy benchmarks, clearly outperforming alternative approaches.
Last, the exponential parameterization is crucial for maintaining positive-definite covariance matrices,
and the use of the natural coordinate system guarantees computational feasibility.

NES applies to general parameterizable distributions. 
In this paper, we have experimentally investigated three variants,
adjusted to the particular properties of different problem classes.
We demonstrated the power of the xNES variant using a full multinormal distribution, which is invariant under arbitrary translations and rotations, on the canonical suite of standard benchmarks.
Additionally, we showed that the restriction of the covariance matrix to a diagonal parameterization (SNES)
allows for scaling to very high dimensions, both on the difficult non-Markovian double pole balancing
task and the Lennard-Jones cluster potential optimization problem. 
Furthermore, we demonstrated that using heavy-tailed distributions (Cauchy) instead of Gaussian distributions
yields substantial benefits in global optimization scenarios with multiple or deceptive local optima.

Unlike many black-box optimization algorithms, NES boasts a clean derivation from first
principles. The relationship of NES to methods from other fields, notably evolution 
strategies~\citep{hansen:2001} and policy gradients~\citep{nac,kakade2002nips,bagnell},
should be evident to readers familiar with
both of these domains, as it marries the concept of fitness-based black-box optimization
from evolutionary methods with the concept of Monte Carlo-based gradient estimation from
the policy gradient framework.

\subsection{Future Work}
\label{sec:future}

One intriguing possibility of extending NES starts by realizing 
that NES is not limited to continuous (flat) domains.
In fact, NES can be designed to directly operate on discrete 
search spaces (e.g., graphs) 
or distributions
with deep structure, such as deep belief networks and deep Boltzmann machines~\citep{dbn,dbm}.
This would allow us to find more complex solutions by combining and recombining building blocks produced by a deep network, opening up an interesting connection to genetic algorithms and their cross-over operators.

Program space constitutes one discrete search space of particular interest, and 
it seems promising to extend the NES approach to program search (genetic programming, \citealp{koza:book92}),
building upon probabilistic representations of program trees (e.g., \citealp{Salustowicz:97ecj,Bosman2004})
for which the natural gradient can be estimated.

Another possible direction of future research could comprise the investigation of sparsely parametrized distributions
by means of which we can exploit \emph{selected} covariance structure while keeping
linear complexity, even in high dimensions. This structure can be either given in the problem domain
(e.g., grouping weights by neuron~\citep{Gomez:08jmlr}) or learned incrementally.

Lastly, the approach is not necessarily restricted to purely following the gradient on expected fitness;
for example, the objective could be formulated differently, 
as in including a trade-off between fitness and information gain~\citep{Schmidhuber2009sice}.

\subsection{Conclusion}
\label{sec:conclusion}
	In this paper, we introduced 
	Natural Evolution Strategies, a novel family of algorithms that
	constitutes a principled alternative to standard stochastic search methods such as evolutionary algorithms.
	Maintaining a parameterized distribution on the set of solution candidates,
	the natural gradient is used to update the distribution's
	parameters in the direction of higher expected fitness. 
	
	A collection of techniques have been introduced, which addresses issues
	of convergence, robustness, computational complexity,
	sensitivity to hyperparameters,
	and sampling methods for algorithm speed.
	We investigated a number of instantiations of the NES family,
	ranging from general-purpose multi-variate normal distributions
	to heavy-tailed and separable distributions specialized
	in global optimization and high dimensionality, respectively.
	The results show best published performance on various standard benchmarks,
	as well as competitive performance on others.		
	
	In conclusion, NES algorithms are high-performing, derived from first principles and easy to implement.
 	Their clean conceptual framework allows for a broad range of intriguing future developments.

\acks{This research was funded through the 7th framework program of the European Union, under grant number 231576 (STIFF project), SNF grants 200020-116674/1, 200021-111968/1 and 200020-122124/1, and SSN grant Sinergia CRSIK0-122697. We thank Faustino Gomez for helpful suggestions and his CoSyNE data, 
as well as Jan Peters, Nikolaus Hansen and Andreas Krause for insightful discussions.
}

\appendix

\section*{Appendix A: Weighted Mann-Whitney Test}
\label{app:w-u-test}
This appendix defines a \emph{Weighted Mann-Whitney} test, as used in section~\ref{sec:as}
for virtual adaptation sampling. 
Although the derivations are trivial, the authors are not aware of
any previous attempt to create a test with the same purpose.

\paragraph{Background.}
The classical Mann-Whitney test determines (with confidence $\rho$) 
whether two sets of samples $S=\{s_i\}$ and $S'=\{s'_i\}$ 
are likely to come from the same distribution. 
For that, the so-called U-statistic is computed:

\begin{equation*}
U=\sum_{
		s_{i}>s_{j}'}1
 +\sum_{
 		s_{i}=s_{j}'}\frac{1}{2}
\end{equation*}

Let 
$\mu =\frac{nn'}{2}$ 
and 
$\sigma =\sqrt{\frac{nn'(n+n'+1)}{12}}$
, where $n$ and $n'$ are the number of samples in $S$ and $S'$, respectively.
We can then determine the significance of the difference between $S$
and $S'$. They are different with confidence $\rho $ if:
\begin{itemize}
	\item $\Phi (\frac{U-\mu }{\sigma })>1-\rho $ (if $S$ has larger values), or
	\item $\Phi (\frac{U-\mu }{\sigma })<\rho $ (if $S'$ has larger values).
\end{itemize}

\paragraph{Introducing Weights.}
Now, assume that every sample in $S$ and $S'$ has a (positive)
weight ($w_{i}$ or $w_{i}'$) associated to it.
We can generalize the Mann-Whitney test by interpreting the weights as
fractional number of occurrences in the sets:
\[
U=\sum_{
		s_{i}>s_{j}'}w_{i}w_{j}'
 +\sum_{
 		s_{i}=s_{j}'}\frac{1}{2}w_{i}w_{j}'
\]

Accordingly, we also need to adjust the number of samples:
$m=\sum_{i=1}^{n}w_{i}$ 
and $m^{\prime }=\sum_{i=1}^{n^{\prime
}}w_{i}^{\prime }$, and thus 
$\mu =\frac{mm'}{2}$ 
and 
$\sigma =\sqrt{\frac{mm'(m+m'+1)}{12}}$.

We can see this weighted U-statistic as an interpolation between the cases covered classical one. 
In fact, if the weights are integers, a sample $s$ with weight $w$ can be
replaced equivalently by $w$ occurrences of the same sample $s$ (each with weight 1).

\vskip 0.2in
\bibliography{jmlrbib}

\begin{thebibliography}{61}
\providecommand{\natexlab}[1]{#1}
\providecommand{\url}[1]{\texttt{#1}}
\expandafter\ifx\csname urlstyle\endcsname\relax
  \providecommand{\doi}[1]{doi: #1}\else
  \providecommand{\doi}{doi: \begingroup \urlstyle{rm}\Url}\fi

\bibitem[Akimoto et~al.(2010)Akimoto, Nagata, Ono, and
  Kobayashi]{Akimoto2010ppsn}
Y.~Akimoto, Y.~Nagata, I.~Ono, and S.~Kobayashi.
\newblock {Bidirectional Relation between CMA Evolution Strategies and Natural
  Evolution Strategies}.
\newblock In \emph{Parallel Problem Solving from Nature (PPSN)}, 2010.

\bibitem[Amari(1998)]{amari98natural}
S.~Amari.
\newblock Natural gradient works efficiently in learning.
\newblock \emph{Neural Computation}, 10\penalty0 (2):\penalty0 251--276, 1998.

\bibitem[Amari and Douglas(1998)]{whynaturalamari}
S.~Amari and S.~C. Douglas.
\newblock Why natural gradient?
\newblock In \emph{Proceedings of the 1998 IEEE International Conference on
  Acoustics, Speech, and Signal Processing (ICASSP '98)}, volume~2, pages
  1213--1216, 1998.

\bibitem[Auger(2005)]{Auger2005proof}
A.~Auger.
\newblock {Convergence results for the (1,$\lambda$)-SA-ES using the theory of
  $\phi$-irreducible Markov chains}.
\newblock \emph{Theoretical Computer Science}, 334\penalty0 (1-3):\penalty0 35
  -- 69, 2005.

\bibitem[Bagnell and Schneider(2003)]{bagnell}
J.~A. Bagnell and J.~Schneider.
\newblock Covariant policy search.
\newblock In \emph{Proceedings of the 18th international joint conference on
  Artificial intelligence}, pages 1019--1024, San Francisco, CA, USA, 2003.
  Morgan Kaufmann Publishers Inc.

\bibitem[Beyer(2001)]{EStheory}
H.-G. Beyer.
\newblock \emph{The theory of evolution strategies}.
\newblock Springer-Verlag New York, Inc., New York, NY, USA, 2001.
\newblock ISBN 3-540-67297-4.

\bibitem[Beyer and Schwefel(2002)]{beyerESintroduction}
H.-G. Beyer and H.-P. Schwefel.
\newblock {Evolution strategies: A comprehensive introduction}.
\newblock \emph{Natural Computing}, 1:\penalty0 3--52, 2002.
\newblock ISSN 1567-7818.

\bibitem[Bosman and Jong(2004)]{Bosman2004}
P.~A. Bosman and E.~D.~D. Jong.
\newblock {Learning Probabilistic Tree Grammars for Genetic Programming}.
\newblock In \emph{Parallel Problem Solving from Nature - PPSN VIII}, volume
  3242 of \emph{Lecture Notes in Computer Science}, pages 192--201, Berlin,
  Heidelberg, 2004. Springer Berlin Heidelberg.

\bibitem[Cartan(1928)]{cartan:1928}
{\'E}.~Cartan.
\newblock Sur la repr\'esentation g\'eom\'etrique des syst\`emes mat\'erieles
  non holonomes.
\newblock In \emph{Proc Int Congr Math, Bologna}, volume~4, pages 253--261,
  1928.

\bibitem[Friedrichs and Igel(2005)]{svmIgel}
F.~Friedrichs and C.~Igel.
\newblock Evolutionary tuning of multiple svm parameters.
\newblock \emph{Neurocomputing}, 64:\penalty0 107--117, 2005.

\bibitem[Glasmachers and Igel(2005)]{glasmachers:2005}
T.~Glasmachers and C.~Igel.
\newblock {Gradient-based Adaptation of General Gaussian Kernels}.
\newblock \emph{Neural Computation}, 17\penalty0 (10):\penalty0 2099--2105,
  2005.

\bibitem[Glasmachers et~al.(2010{\natexlab{a}})Glasmachers, Schaul, and
  Schmidhuber]{Glasmachers2010}
T.~Glasmachers, T.~Schaul, and J.~Schmidhuber.
\newblock {A Natural Evolution Strategy for Multi-Objective Optimization}.
\newblock In \emph{Parallel Problem Solving from Nature (PPSN)},
  2010{\natexlab{a}}.

\bibitem[Glasmachers et~al.(2010{\natexlab{b}})Glasmachers, Schaul, Sun,
  Wierstra, and Schmidhuber]{Glasmachers2010a}
T.~Glasmachers, T.~Schaul, Y.~Sun, D.~Wierstra, and J.~Schmidhuber.
\newblock {Exponential Natural Evolution Strategies}.
\newblock In \emph{Genetic and Evolutionary Computation Conference (GECCO)},
  Portland, OR, 2010{\natexlab{b}}.

\bibitem[Goldberg(1989)]{Goldberg}
D.~E. Goldberg.
\newblock \emph{Genetic Algorithms in Search, Optimization and Machine
  Learning}.
\newblock Addison-Wesley Longman Publishing Co., Inc., Boston, MA, USA, 1st
  edition, 1989.
\newblock ISBN 0201157675.

\bibitem[Gomez et~al.(2008)Gomez, Schmidhuber, and Miikkulainen]{Gomez:08jmlr}
F.~Gomez, J.~Schmidhuber, and R.~Miikkulainen.
\newblock {Accelerated Neural Evolution through Cooperatively Coevolved
  Synapses}.
\newblock \emph{Journal of Machine Learning Research}, 2008.

\bibitem[Hansen and Auger(2010)]{bbobsetup}
N.~Hansen and A.~Auger.
\newblock {Real-parameter black-box optimization benchmarking 2010:
  Experimental setup}, 2010.

\bibitem[Hansen and Finck(2010{\natexlab{a}})]{bbobnoisefree}
N.~Hansen and S.~Finck.
\newblock {Real-parameter black-box optimization benchmarking 2010: Noiseless
  functions definitions}, 2010{\natexlab{a}}.

\bibitem[Hansen and Finck(2010{\natexlab{b}})]{bbobnoisy}
N.~Hansen and S.~Finck.
\newblock {Real-Parameter Black-Box Optimization Benchmarking 2010: Noisy
  Functions Definitions}, 2010{\natexlab{b}}.

\bibitem[Hansen and Ostermeier(2001)]{hansen:2001}
N.~Hansen and A.~Ostermeier.
\newblock Completely derandomized self-adaptation in evolution strategies.
\newblock \emph{Evolutionary Computation}, 9\penalty0 (2):\penalty0 159--195,
  2001.

\bibitem[Hansen et~al.(2009)Hansen, Niederberger, Guzzella, and
  Koumoutsakos]{control}
N.~Hansen, A.~S.~P. Niederberger, L.~Guzzella, and P.~Koumoutsakos.
\newblock A method for handling uncertainty in evolutionary optimization with
  an application to feedback control of combustion.
\newblock \emph{Trans. Evol. Comp}, 13:\penalty0 180--197, 2009.

\bibitem[Hasenj\"{a}ger et~al.(2005)Hasenj\"{a}ger, Sendhoff, Sonoda, and
  Arima]{aeronautic}
M.~Hasenj\"{a}ger, B.~Sendhoff, T.~Sonoda, and T.~Arima.
\newblock {Three dimensional evolutionary aerodynamic design optimization with
  CMA-ES}.
\newblock In \emph{Proceedings of the 2005 conference on Genetic and
  evolutionary computation}, GECCO '05, pages 2173--2180, New York, NY, USA,
  2005. ACM.

\bibitem[Hinton and Salakhutdinov(2006)]{dbn}
G.~Hinton and R.~Salakhutdinov.
\newblock Reducing the dimensionality of data with neural networks.
\newblock \emph{Science}, 313\penalty0 (5786):\penalty0 504 -- 507, 2006.

\bibitem[Holland(1992)]{Holland}
J.~H. Holland.
\newblock \emph{Adaptation in natural and artificial systems}.
\newblock MIT Press, Cambridge, MA, USA, 1992.
\newblock ISBN 0-262-58111-6.

\bibitem[Igel and Husken(2003)]{rprop}
C.~Igel and M.~Husken.
\newblock Empirical evaluation of the improved rprop learning algorithm.
\newblock \emph{Neurocomputing}, 50:\penalty0 2003, 2003.

\bibitem[Jastrebski and Arnold(2006)]{Jastrebski2006}
G.~A. Jastrebski and D.~V. Arnold.
\newblock {Improving Evolution Strategies through Active Covariance Matrix
  Adaptation}.
\newblock In \emph{IEEE Congress on Evolutionary Computation}, 2006.

\bibitem[Jebalia et~al.(2007)Jebalia, Auger, Schoenauer, James, and
  Postel]{chromatography}
M.~Jebalia, A.~Auger, M.~Schoenauer, F.~James, and M.~Postel.
\newblock Identification of the isotherm function in chromatography using
  cma-es.
\newblock In \emph{IEEE Congress on Evolutionary Computation}, pages
  4289--4296, 2007.

\bibitem[Jebalia et~al.(2010)Jebalia, Auger, and Hansen]{jah:2010a}
M.~Jebalia, A.~Auger, and N.~Hansen.
\newblock {Log-linear convergence and divergence of the scale-invariant
  (1+1)-ES in noisy environments}.
\newblock \emph{Algorithmica}, pages 1--36, 2010.
\newblock online first.

\bibitem[Kakade(2002)]{kakade2002nips}
S.~Kakade.
\newblock A natural policy gradient.
\newblock \emph{Advances in Neural Information Processing Systems 14},
  2:\penalty0 1531--1538, 2002.

\bibitem[Kennedy and Eberhart(2001)]{PSO}
J.~Kennedy and R.~Eberhart.
\newblock \emph{Swarm Intelligence}.
\newblock Morgan Kaufmann, San Francisco, CA, 2001.

\bibitem[Kirkpatrick et~al.(1983)Kirkpatrick, Gelatt, Jr, and
  Vecchi]{simulatedannealing}
S.~Kirkpatrick, C.~D. Gelatt, Jr, and M.~P. Vecchi.
\newblock {Optimization by Simulated Annealing}.
\newblock \emph{Science}, 220:\penalty0 671--680, 1983.

\bibitem[Klockgether and Schwefel(1970)]{nozzle}
J.~Klockgether and H.~P. Schwefel.
\newblock Two-phase nozzle and hollow core jet experiments.
\newblock In \emph{Proc. 11th Symp. Engineering Aspects of
  Magnetohydrodynamics}, pages 141--148, 1970.

\bibitem[Koza(1992)]{koza:book92}
J.~R. Koza.
\newblock \emph{{Genetic Programming: On the Programming of Computers by Means
  of Natural Selection}}.
\newblock Cambridge, MA, 1992.

\bibitem[Kullback and Leibler(1951)]{Kullback1951}
S.~Kullback and R.~A. Leibler.
\newblock {On information and sufficiency}.
\newblock \emph{The Annals of Mathematical Statistics}, 22\penalty0
  (1):\penalty0 79--86, 1951.
\newblock ISSN 00034851.

\bibitem[Larra{\~{n}}aga(2002)]{EDA}
P.~Larra{\~{n}}aga.
\newblock \emph{{E}stimation of {D}istribution {A}lgorithms. {A} {N}ew {T}ool
  for {E}volutionary {C}omputation}, chapter An introduction to probabilistic
  graphical models, pages 25--54.
\newblock Kluwer Academic Publishers, 2002.

\bibitem[Matsumoto and Nishimura(1998)]{Matsumoto1998}
M.~Matsumoto and T.~Nishimura.
\newblock {Mersenne twister: a 623-dimensionally equidistributed uniform
  pseudo-random number generator}.
\newblock \emph{Acm Transactions On Modeling And Computer Simulation},
  8\penalty0 (1):\penalty0 3--30, 1998.
\newblock ISSN 10493301.

\bibitem[Muller et~al.(2002)Muller, Marchetto, Airaghi, and
  Koumoutsakos]{chemotaxis}
S.~D. Muller, J.~Marchetto, S.~Airaghi, and P.~Koumoutsakos.
\newblock Optimization based on bacterial chemotaxis.
\newblock \emph{IEEE Transactions on Evolutionary Computation}, 6:\penalty0
  6--16, 2002.

\bibitem[Najfeld and Havel(1994)]{najfeld:1994}
I.~Najfeld and T.~F. Havel.
\newblock {Derivaties of the Matrix Exponential and Their Computation}.
\newblock \emph{Adv. Appl. Math}, 16:\penalty0 321--375, 1994.

\bibitem[Nelder and Mead(1965)]{neldermead}
J.~A. Nelder and R.~Mead.
\newblock {A Simplex Method for Function Minimization}.
\newblock \emph{The Computer Journal}, 7\penalty0 (4):\penalty0 308--313, 1965.

\bibitem[Peters(2007)]{petersthesis}
J.~Peters.
\newblock \emph{Machine Learning of Motor Skills for Robotics}.
\newblock PhD thesis, epartment of Computer Science, University of Southern
  California, 2007.

\bibitem[Peters and Schaal(2008)]{nac}
J.~Peters and S.~Schaal.
\newblock Natural actor-critic.
\newblock \emph{Neurocomputing}, 71\penalty0 (7-9):\penalty0 1180--1190, 2008.
\newblock ISSN 0925-2312.

\bibitem[Rechenberg and Eigen(1973)]{RechenbergES}
I.~Rechenberg and M.~Eigen.
\newblock \emph{{Evolutionsstrategie: Optimierung technischer Systeme nach
  Prinzipien der biologischen Evolution}}.
\newblock Frommann-Holzboog Stuttgart, 1973.

\bibitem[Riedmiller and Braun(1993)]{riedmiller2002direct}
M.~Riedmiller and H.~Braun.
\newblock {A direct adaptive method for faster backpropagation learning: The
  RPROP algorithm}.
\newblock In \emph{IEEE International Conference on Neural Networks}, pages
  586--591. IEEE Press, 1993.

\bibitem[Ros and Hansen(2008)]{ros:2008}
R.~Ros and N.~Hansen.
\newblock {A Simple Modification in CMA-ES Achieving Linear Time and Space
  Complexity}.
\newblock In R.~et~al., editor, \emph{Parallel Problem Solving from Nature,
  PPSN X}, pages 296--305. Springer, 2008.

\bibitem[Rubinstein and Kroese(2004)]{CEM}
R.~Y. Rubinstein and D.~P. Kroese.
\newblock \emph{{The Cross-Entropy Method: A Unified Approach to Combinatorial
  Optimization, Monte-Carlo Simulation and Machine Learning (Information
  Science and Statistics)}}.
\newblock Springer, 2004.

\bibitem[Salakhutdinov and Hinton(2009)]{dbm}
R.~Salakhutdinov and G.~Hinton.
\newblock Deep {B}oltzmann machines.
\newblock In \emph{Proceedings of the International Conference on Artificial
  Intelligence and Statistics}, volume~5, pages 448--455, 2009.

\bibitem[Salustowicz and Schmidhuber(1997)]{Salustowicz:97ecj}
R.~P. Salustowicz and J.~Schmidhuber.
\newblock {Probabilistic Incremental Program Evolution}.
\newblock \emph{Evolutionary Computation}, 5:\penalty0 123--141, 1997.

\bibitem[Schaul and Schmidhuber(2010{\natexlab{a}})]{Schaul2010metalearning}
T.~Schaul and J.~Schmidhuber.
\newblock Metalearning.
\newblock \emph{Scholarpedia}, 5\penalty0 (6):\penalty0 4650,
  2010{\natexlab{a}}.

\bibitem[Schaul and Schmidhuber(2010{\natexlab{b}})]{schaul2010puns}
T.~Schaul and J.~Schmidhuber.
\newblock {Towards Practical Universal Search}.
\newblock In \emph{Conference on Artificial General Intelligence (AGI)},
  Lugano, 2010{\natexlab{b}}.

\bibitem[Schaul et~al.(2010)Schaul, Bayer, Wierstra, Sun, Felder, Sehnke,
  R\"{u}ckstie\ss, and Schmidhuber]{Schaul2010pybrain}
T.~Schaul, J.~Bayer, D.~Wierstra, Y.~Sun, M.~Felder, F.~Sehnke,
  T.~R\"{u}ckstie\ss, and J.~Schmidhuber.
\newblock {PyBrain}.
\newblock \emph{Journal of Machine Learning Research}, 11:\penalty0 743--746,
  2010.

\bibitem[Schaul et~al.(2011)Schaul, Glasmachers, and
  Schmidhuber]{Schaul2011snes}
T.~Schaul, T.~Glasmachers, and J.~Schmidhuber.
\newblock {High Dimensions and Heavy Tails for Natural Evolution Strategies}.
\newblock In \emph{To appear in: Genetic and Evolutionary Computation
  Conference (GECCO)}, 2011.

\bibitem[Schmidhuber(2009)]{Schmidhuber2009sice}
J.~Schmidhuber.
\newblock {Simple Algorithmic Theory of Subjective Beauty, Novelty, Surprise,
  Interestingness, Attention, Curiosity, Creativity, Art, Science, Music,
  Jokes}.
\newblock \emph{Journal of SICE}, 48\penalty0 (1):\penalty0 21--32, 2009.

\bibitem[Schwefel(1977)]{schwefelES}
H.-P. Schwefel.
\newblock \emph{Numerische Optimierung von Computer-Modellen mittels der
  Evolutionsstrategie}, volume~26.
\newblock Birkhaeuser, Basel/Stuttgart, 1977.

\bibitem[Shepherd et~al.(2006)Shepherd, McDowell, and Jacob]{crystal}
J.~Shepherd, D.~McDowell, and K.~Jacob.
\newblock Modeling morphology evolution and mechanical behavior during
  thermo-mechanical processing of semi-crystalline polymers.
\newblock \emph{Journal of the Mechanics and Physics of Solids}, 54\penalty0
  (3):\penalty0 467 -- 489, 2006.

\bibitem[Shir and B\"{a}ck(2007)]{quantum}
O.~M. Shir and T.~B\"{a}ck.
\newblock The second harmonic generation case-study as a gateway for es to
  quantum control problems.
\newblock In \emph{Proceedings of the 9th annual conference on Genetic and
  evolutionary computation}, GECCO '07, pages 713--721, New York, NY, USA,
  2007. ACM.

\bibitem[Storn and Price(1997)]{differentialevolution}
R.~Storn and K.~Price.
\newblock Differential evolution -– a simple and efficient heuristic for
  global optimization over continuous spaces.
\newblock \emph{J. of Global Optimization}, 11:\penalty0 341--359, December
  1997.
\newblock ISSN 0925-5001.

\bibitem[Sun et~al.(2009{\natexlab{a}})Sun, Wierstra, Schaul, and
  Schmidhuber]{sun:2009a}
Y.~Sun, D.~Wierstra, T.~Schaul, and J.~Schmidhuber.
\newblock {Stochastic Search using the Natural Gradient}.
\newblock In \emph{International Conference on Machine Learning (ICML)},
  2009{\natexlab{a}}.

\bibitem[Sun et~al.(2009{\natexlab{b}})Sun, Wierstra, Schaul, and
  Schmidhuber]{sun:2009b}
Y.~Sun, D.~Wierstra, T.~Schaul, and J.~Schmidhuber.
\newblock {Efficient Natural Evolution Strategies}.
\newblock In \emph{Genetic and Evolutionary Computation Conference (GECCO)},
  2009{\natexlab{b}}.

\bibitem[Wales and Doye(1998)]{Wales1998}
D.~Wales and J.~Doye.
\newblock {Global Optimization by Basin-Hopping and the Lowest Energy
  Structures of Lennard-Jones Clusters Containing up to 110 Atoms}.
\newblock \emph{The Journal of Physical Chemistry A}, 101\penalty0
  (28):\penalty0 8, 1998.

\bibitem[Wieland(1991)]{wieland91pole}
A.~Wieland.
\newblock {Evolving Neural Network Controllers for Unstable Systems}.
\newblock In \emph{Proceedings of the International Joint Conference on Neural
  Networks (Seattle, WA)}, pages 667--673, 1991.

\bibitem[Wierstra et~al.(2008)Wierstra, Schaul, Peters, and
  Schmidhuber]{wierstra:2008}
D.~Wierstra, T.~Schaul, J.~Peters, and J.~Schmidhuber.
\newblock {Natural Evolution Strategies}.
\newblock In \emph{Proceedings of the Congress on Evolutionary Computation
  (CEC08), Hongkong}. IEEE Press, 2008.

\bibitem[Winter et~al.(2005)Winter, Brendel, and Igel]{health}
S.~Winter, B.~Brendel, and C.~Igel.
\newblock Registration of bone structures in 3d ultrasound and ct data:
  Comparison of different optimization strategies.
\newblock \emph{International Congress Series}, 1281:\penalty0 242 -- 247,
  2005.

\end{thebibliography}

\end{document}